%% file: 00_PredictivePlanning.tex
\documentclass{ieeetj}

\usepackage{graphicx}
\usepackage{amsmath} 
\usepackage{amssymb}  
\usepackage{algorithm}
\usepackage[noend]{algpseudocode} 
\algnewcommand\AAND{\textbf{ and }}
\algnewcommand\Or{\textbf{ or }}
\usepackage{color}

\usepackage{url}
\usepackage{breakurl}
\usepackage[breaklinks]{hyperref}
\usepackage[nolist,nohyperlinks]{acronym}
\usepackage[binary-units]{siunitx}
\input{acronyms.tex}

\def\BibTeX{{\rm B\kern-.05em{\sc i\kern-.025em b}\kern-.08em
    T\kern-.1667em\lower.7ex\hbox{E}\kern-.125emX}}
\AtBeginDocument{\definecolor{tmlcncolor}{cmyk}{0.93,0.59,0.15,0.02}\definecolor{NavyBlue}{RGB}{0,86,125}}

\DeclareMathAlphabet{\pazocal}{OMS}{zplm}{m}{n}

\newcommand{\Ys}{\pazocal{Y}}

\newcommand{\Ws}{\pazocal{W}}

\newcommand{\Ds}{\pazocal{D}}

\newcommand{\Es}{\pazocal{E}}

\newcommand{\Ls}{\pazocal{L}}
\newcommand{\Vs}{\pazocal{V}}

\newcommand{\Ss}{\pazocal{S}}
\newcommand{\Is}{\pazocal{I}}

\newcommand{\Ps}{\pazocal{P}}
\newcommand{\Hs}{\pazocal{H}}
\newcommand{\Zs}{\mathcal{Z}}

\newcommand{\vbf}{\mathbf{v}}
\newcommand{\pbf}{\mathbf{p}}
\newcommand{\dbf}{\mathbf{d}}
\newcommand{\xbf}{\mathbf{x}}
\newcommand{\Rbf}{\mathbf{R}}
\newcommand{\bbf}{\mathbf{b}}

\newcommand{\abf}{\mathbf{a}}

\newcommand{\Sbb}{\mathbb{S}}
\newcommand{\Mbb}{\mathbb{M}}
\newcommand{\Gbb}{\mathbb{G}}

\newcommand{\Ibb}{\mathbb{I}}

\newcommand{\Ifk}{\mathfrak{I}}

\newcommand{\addition}[1]{\textcolor{black}{#1}}

\def \*#1 {mathbf{#1}}
\def \@#1 {\mathbb{#1}}

\newtheorem{definition}{Definition}
\newtheorem{problem}{Problem}

\DeclareMathAlphabet{\mathpzc}{OT1}{pzc}{m}{it}

\usepackage{array}
\newcolumntype{C}[1]{>{\centering\arraybackslash}p{#1}}
\newcolumntype{M}[1]{>{\raggedright\arraybackslash}p{#1}}

\usepackage{array} 
\newcolumntype{L}[1]{>{\raggedright\let\newline\\\arraybackslash\hspace{0pt}}m{#1}}	
\newcolumntype{S}[1]{>{\centering\let\newline\\\arraybackslash\hspace{0pt}}m{#1}}
\newcolumntype{R}[1]{>{\raggedleft\let\newline\\\arraybackslash\hspace{0pt}}m{#1}}

\makeatletter
\renewcommand*{\@opargbegintheorem}[3]{\trivlist
  \item[\hskip \labelsep{\itshape #1\ #2}] \textit{(#3)}\ }
\makeatother

\def\OJlogo{\vspace{-4pt}$Semantics-aware ~Predictive ~Inspection ~Path ~Planning$}
\def\seclogo{\vspace{10pt}$Semantics-aware ~Predictive ~Inspection ~Path ~Planning$}

\def\authorrefmark#1{\ensuremath{^{\textbf{#1}}}}

\begin{document}
\reviseddate{20 April, 2025}
\accepteddate{2 June, 2025}

\title{
Semantics-aware Predictive Inspection Path Planning
}


\author{Mihir Dharmadhikari\authorrefmark{1} and Kostas Alexis\authorrefmark{1}}
\affil{Norwegian University of Science and Technology (NTNU), O. S. Bragstads Plass 2D, 7034, Trondheim, Norway}
\authornote{This material was supported by the Research Council of Norway under project SENTIENT (NO-321435) and the European Commission through the project AUTOASSESS under the Horizon Europe Grant (101120732).}
\corresp{Corresponding author: Mihir Dharmadhikari (email: {\tt\small mihir.dharmadhikari@ntnu.no})}


\begin{abstract}
This paper presents a novel semantics-aware inspection path planning paradigm called ``Semantics-aware Predictive Planning'' (SPP). Industrial environments that require the inspection of specific objects or structures (called ``semantics''), such as ballast water tanks inside ships, often present structured and repetitive spatial arrangements of the semantics of interest. Motivated by this, we first contribute an algorithm that identifies spatially repeating patterns of semantics - exact or inexact - in a semantic scene graph representation and makes predictions about the evolution of the graph in the unseen parts of the environment using these patterns. Furthermore, two inspection path planning strategies, tailored to ballast water tank inspection, that exploit these predictions are proposed. To assess the performance of the novel predictive planning paradigm, both simulation and experimental evaluations are performed. First, we conduct a simulation study comparing the method against relevant state-of-the-art techniques and further present tests showing its ability to handle imperfect patterns. Second, we deploy our method onboard a collision-tolerant aerial robot operating inside the ballast tanks of two real ships. The results, both in simulation and field experiments, demonstrate significant improvement over the state-of-the-art in terms of inspection time while maintaining equal or better semantic surface coverage. A set of videos describing the different parts of the method and the field deployments are available at \url{https://tinyurl.com/spp-videos}. \addition{The code for this work is made available at \url{https://github.com/ntnu-arl/predictive_planning_ros}.}
 
\end{abstract}

\maketitle

\section{INTRODUCTION}\label{sec:intro}
\input{01_Introduction}

\section{RELATED WORK}\label{sec:related}
\input{02_RelatedWork}

\section{PROBLEM FORMULATION}\label{sec:probstat}
\input{03_ProblemFormulation}

\section{SCENE GRAPH PATTERN PREDICTION}\label{sec:pat_pred}
\input{04_PatternPrediction}

\section{PREDICTIVE PLANNING}\label{sec:app}
\input{05_PredictivePlanning}

\section{SIMULATION STUDIES}\label{sec:simulation}
\input{06_SimulationStudies}

\section{FIELD EXPERIMENTS}\label{sec:field_experiments}
\input{07_FieldExperiments}

\section{CONCLUSIONS}\label{sec:concl}
\input{08_Conclusion} 

\bibliographystyle{IEEEtran}
\bibliography{./00_PredictivePlanning}

\end{document}

%% file: acronyms.tex
\acrodef{ssg}[SSG]{Semantic Scene Graph}
\acrodef{mdl}[MDL]{Minimum Description Length}
\acrodef{dl}[DL]{Description Length}
\acrodef{fov}[FoV]{Field of View}
\acrodef{ec}[EC]{Entry Class}
\acrodef{ecs}[ECs]{Entry Classes}
\acrodef{ev}[EV]{Entry Vertex}
\acrodef{evs}[EVs]{Entry Vertices}
\acrodef{cev}[CEV]{Complementary Entry Vertex}
\acrodef{cevs}[CEVs]{Complementary Entry Vertices}
\acrodef{lev}[LEV]{Loose Entry Vertex}
\acrodef{levs}[LEVs]{Loose Entry Vertices}
\acrodef{lcev}[LCEV]{Loose Complementary Entry Vertex}
\acrodef{lcevs}[LCEVs]{Loose Complementary Entry Vertices}
\acrodef{ve}[VE]{Volumetric Exploration}
\acrodef{si}[SI]{Semantic Inspection}
\acrodef{pp}[PP]{Predictive Planning}
\acrodef{ae}[PP-AE]{Assisted Exploration}
\acrodef{oi}[PP-OI]{Opportunistic Inspection}
\acrodef{tsdf}[TSDF]{Truncated Signed Distance Field}
\acrodef{tsp}[TSP]{Travelling Salesman Problem}
\acrodef{nbv}[NBV]{Next-Best-View}
\acrodef{uav}[UAV]{Unmanned Aerial Vehicle}
\acrodef{tsdf}[TSDF]{Truncated Signed Distance Field}

%% file: 01_Introduction.tex
In recent years, a spectrum of contributions has been made in the domain of autonomous robotic exploration, mapping, and inspection path planning~\cite{GBPLANNER_JFR_2020,agha2021nebula,BABOOMS_ICRA_15,cao2021tare}. Accelerated by this research, robotic systems have been successfully deployed in a variety of environments including structured industrial settings such as those found in the oil \& gas industry~\cite{shukla2016application,sa2014vertical,gehring2019anymal,caprari2012highly,chan2015towards,BABOOMS_ICRA_15}, or unstructured natural scenes~\cite{bartolomei2023fast} such as subterranean settings~\cite{CERBERUS_SCIENCE_2022,CERBERUS_WINS_FR2022submission,GBPLANNER2COHORT_ICRA_2022,GBPLANNER_JFR_2020,hudson2021heterogeneous,agha2021nebula,rouvcek2019darpa,explorer_phase_i_ii}.
As more robots get deployed in human-made environments, more complex tasks, such as object search and inspection of specific structures, emerge. We refer to the objects and structures relevant to a mission as ``semantics''. These new tasks require the robot to have an object-level ``semantic'' understanding of its surroundings and to be able to take action accordingly.
As a result, the robotics community has seen an increased interest in metric-semantic representations of the environments~\cite{hughes2022hydra,hughes2024foundations,Wu_2021_CVPR,schmid2022panoptic} and semantics-aware path planning~\cite{Wang2021semantic_info_planning,Papatheodorou_ICRA2023,fredriksson2024topometric,ginting2024semanticbeliefbehaviorgraph,roth2023viplanner}. 

\begin{figure*}[h!]
\centering
    \includegraphics[width=0.99\textwidth]{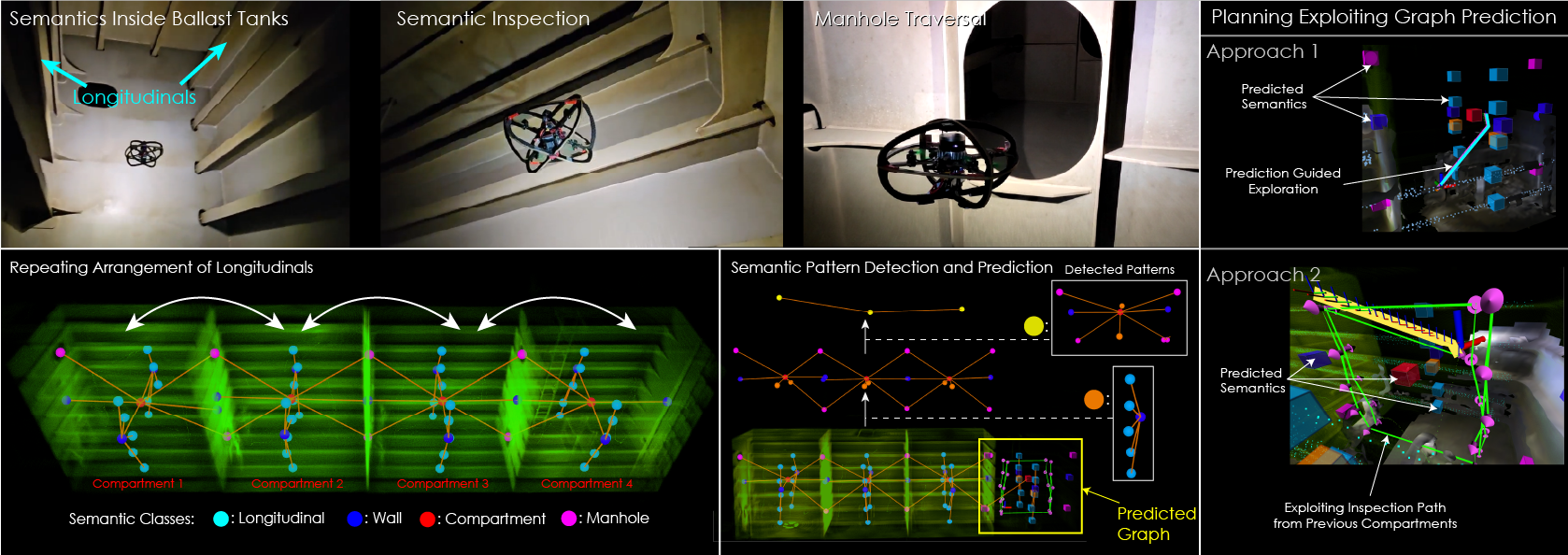}
\vspace{1ex} 
\centering
\caption{Instances of our collision-tolerant aerial robot inspecting and navigating in ballast tank environments using the proposed Semantics-aware Predictive Planner. Additionally, the semantics of interest inside these environments are shown highlighting their spatially repeating arrangement. The bottom center sub-figure shows an instance of pattern detection and graph prediction in the Semantic Scene Graph (SSG). It is on this prediction of semantics that predictive planning, shown in the right two subfigures, takes place.}
\label{fig:intro}
\end{figure*}

Despite the increased interest and the progress in semantic scene reasoning~\cite{rosinol2020kimera,maggio2024clio,bavle2022situational}, the majority of the current research in semantics-aware path planning either treats the semantics individually - without accounting for the relations between them - or only exploits the relations between the semantics seen so far~\cite{Papatheodorou_ICRA2023,2023swap,ginting2024seek}. However, such approaches do not account for the fact that especially in industrial environments, semantics of essential importance for inspection are not distributed arbitrarily but instead present highly structured relationships among them. In particular, facilities like ship ballast water tanks present spatially repeating patterns of the objects of interest (e.g., Figure~\ref{fig:intro}). This opens up the avenue to exploit the repeatable nature of semantics to make predictions about the unseen parts of the environment during an inspection mission and thus greatly improve the efficiency and systematicity of such tasks.

Motivated by the above, this paper presents a novel semantic inspection paradigm called, ``Semantics-aware Predictive Planning'' (SPP), for inspection of the semantics of interest in environments where repeatable object patterns exist.  The proposed method uses a dual environment representation consisting of a) a volumetric map capturing the geometry~\cite{voxblox} and b) a \ac{ssg}~\cite{armeni_iccv19,kim2020ssg,rosinol2020dssg}. 
Using this representation, first, an algorithm to identify repeating patterns of semantics in the \ac{ssg} is presented.
Second, a strategy to predict the evolution of the \ac{ssg} using the detected patterns is described. 
Finally, two inspection path planning strategies, tailored to the inspection of semantics inside ballast water tanks, are proposed that exploit these predictions towards superior mission efficiency. \addition{The source code of this paper is made available at: \url{https://github.com/ntnu-arl/predictive_planning_ros}.}

In the simulation studies, the proposed planner is tested and compared against state-of-the-art exploration and inspection path planning methods as benchmarks in a model of a large ballast tank \addition{and a factory}.
The study demonstrates a significant improvement (ranging from $25\%$ to $60\%$ depending on the method compared against) in the inspection time with the use of the proposed strategies. 
Additionally, tests are conducted in a model of a ballast tank containing imperfect patterns to demonstrate the ability of the proposed method to handle the case of missing semantics successfully.

Finally, the paper presents results from field deployments in the ballast tanks of two oil tanker ships, with the method integrated onboard a collision-tolerant aerial robot~\cite{rmfowl}. In both ships, the two proposed planning approaches are tested and compared. The robot is successfully able to navigate in the ballast tanks in all the missions, build the \ac{ssg}, and complete the inspection of the semantics important for inspection. The proposed planning approaches show an improvement of up to $23\%$ in the inspection time compared to a semantics-aware ``Baseline'' that is not exploiting prediction thus demonstrating the real-world applicability of the method. Significantly improving the efficiency of such inspection missions is essential, especially considering the scale of such environments (often involving multiple ballast compartments across levels) and the limited flight time of the aerial robots that can be used in such deployments given the narrow spaces that they have to navigate through. 

The remainder of this paper is organized as follows: Section~\ref{sec:related} outlines related work, followed by the problem formulation in Section~\ref{sec:probstat}. The proposed approach for \ac{ssg} pattern prediction and predictive planning is detailed in Sections~\ref{sec:pat_pred} and~\ref{sec:app} respectively. Simulation studies are shown in Section~\ref{sec:simulation}, while results from field tests are presented in Section~\ref{sec:field_experiments}. Finally, conclusions are drawn in Section~\ref{sec:concl}. 

%% file: 02_RelatedWork.tex
In recent years, there has been a boost in the research on semantics perception. Several works have investigated the problem of semantic segmentation~\cite{2022domain_adapt_seg,yin2023dformer,jia2024geminifusion,zhang2023cmx} and their representation for planning in the form of an \ac{ssg} or metric-semantic maps~\cite{hughes2022hydra,hughes2024foundations,Wu_2021_CVPR,schmid2022panoptic}.
\addition{The authors of~\cite{alama2025rayfrontsopensetsemanticray} propose a framework for fine-grained semantic mapping that supports open-set object queries both within and beyond the depth-sensing range, by storing semantic information within range using a voxel representation and representing information beyond range as rays. \cite{hughes2022hydra,hughes2024foundations} present a hierarchical semantic scene graph representation that not only captures the objects and their relations in the scene graph but also creates higher-level abstractions by grouping them into predefined structures such as places, rooms, and buildings. The authors of \cite{Wu_2021_CVPR} propose an incremental $3$D scene graph generation method using a sequence of RGBD sensor frames. \cite{schmid2022panoptic} presents a multi-resolution metric-semantic \ac{tsdf} representation that allocates high-resolution submaps for semantic objects and can achieve object-level consistency over long time horizons. \cite{viswanathan2024actionablehierarchicalscenerepresentation} presents an actionable hierarchical scene representation where the levels are guided by the proposed inspection and exploration planner. Despite the progress, these methods do not provide the possibility to detect patterns in the scene graphs or make predictions about the unseen parts of the environment. Furthermore, hierarchical representations like~\cite{hughes2022hydra} require predefined levels in the hierarchy.}
Motivated by the above, an increased research output can be seen in the use of semantic understanding in the domain of path planning.

Among the early works in informative path planning,~\cite{Wang2021semantic_info_planning} tackles the problem of object search by maintaining object-object and object-scene co-occurrence probabilities in an information map, which is then used for informative path planning. The authors in~\cite{fredriksson2024topometric} present an exploration path planning strategy utilizing a semantic topometric map built using structural semantics such as intersections, pathways, etc. As semantics-aware informative path planning requires a combined metric-semantic map. The work in~\cite{zaenker2020hypermap} presents a framework to handle multi-layered metric-semantic maps along with its application in semantics-aware exploration path planning.

In recent years, several works have proposed heuristic functions to combine the exploration and semantic mapping objectives. An information gain formulation based on entropy change in the map, where each voxel's contribution to the entropy is weighted by it belonging to an object of interest, is proposed in~\cite{kay2021semanticnbvreconstruction} for \ac{nbv} generation for the task of $3$D reconstruction. \addition{The authors of \cite{simons2025seguesemanticguidedexploration} propose an entropy-based function to score the information gain of a pose based on visible semantic features and use it in an \ac{nbv} exploration path planning formulation.}
The work in~\cite{Papatheodorou_ICRA2023} proposes an objective function for viewpoint evaluation combining the objectives of maximizing unknown volume mapping, viewing all surfaces from a given distance for object detection, and viewing all semantics at a different maximum viewing distance. The authors of~\cite{rs12050891} present an occupancy map encoding the probabilities of the voxels belonging to a semantic class and exploit it in an \ac{nbv} exploration planning approach to explore and increase the detection confidence of the semantics. Similarly, the effort in~\cite{yu2024semanticawarenextbestviewmultidofsmobile} tackles the problem of semantic-object search and mapping in unknown environment using the \ac{nbv} approach with a heuristic balancing exploration and semantic mapping quality.
Along these lines,~\cite{milas2023asep} uses a similar heuristic with a frontier-based exploration strategy. \addition{The authors in~\cite{lu2024semRecedingHorizon} present a multi-layered object-centric volumetric map which is then used to extract semantics-aware frontiers for object centric mapping.}
Departing from heuristics combining multiple objectives in an additive fashion, our previous work in~\cite{2023swap} presents a combined exploration and semantic inspection strategy that utilizes distinct planning behaviors for volumetric exploration, semantic surface reconstruction, and semantic inspection.
The contribution in~\cite{ginting2024semanticbeliefbehaviorgraph} presents a data structure called the semantic belief behavior graph that encodes behavior nodes for various semantics-aware policies. 
The work in~\cite{STACHE2023104288} presents a path planning algorithm for \ac{uav} to map large areas (e.g., agricultural fields) that adapts the paths to obtain high-resolution semantic segmentations of areas of interest. The authors of~\cite{rukin2023activelearning} detail a planning framework for active learning in \ac{uav}-based semantic mapping. 
In addition to the above single robot path planning works, the efforts in~\cite{liu2023multiaerialsemanticmapping,miller2024spomp,cladera2024challengesopportunitieslargescaleexploration} present collaborative path planning methods for multi-robot semantics mapping.

Semantics-aware planning has also found its utility in navigation tasks. The authors of~\cite{bartolomei2021activesemanticperception} propose a path planning strategy for a robot using a vision-based localization solution that aims to avoid areas leading to high localization drift using semantically segmented images. \addition{The work in~\cite{HU2025105949_indoorSemanticNav} presents a RandLA-Net and KNN-based algorithm for metric-semantic mapping and proposes a semantics-aware A* algorithm that accounts for risks associated with the types of obstacles.} The local planner presented in~\cite{roth2023viplanner} uses semantic information to learn traversability for legged robots. Similarly,~\cite{chen2023rspmp} proposes a local planner for unmanned ground vehicles using a metric-semantic traversability map. 
Furthermore, several works have focused on object goal navigation. This task involves searching for an object belonging to a given semantic class in an unknown environment. 
The effort in~\cite{ginting2024seek} proposes a probabilistic planning framework that utilizes a Relational Semantic Network trained to estimate the probability of finding the target object within the spatial elements of an \ac{ssg}.
Several learning-based approaches, such as~\cite{sun2024rsmpnet,du2020learningrelationgraph,liang2021sscnav,liang2021sscnav,Yang2019_113270,qiu2020learning}, train models to learn relationships between different semantic classes using knowledge graphs or existing \ac{ssg} datasets, to take actions based on semantics seen in the vicinity of the robot. \addition{The work in~\cite{ravichandran2025spineonlinesemanticplanning} utilizes \ac{ssg} along with a Large Language Model to perform online planning for tasks described in natural language.}

Beyond the above, this work also relates to the niche community investigating the prediction of the unknown areas of the map. The works in~\cite{shrestha2019learnedmapprediction} and~\cite{tao2023seer} propose strategies to complete partial occupancy maps using learning-based approaches. On the other hand, the contribution in~\cite{strom2015predictfromprev} presents work on map prediction beyond frontiers using previously mapped environments. The above approaches demonstrate the utility of the predicted maps for exploration path planning. However, they do not exploit semantic information. Beyond pure geometric predictions, a plethora of works has happened in semantic scene completion which aims to fill in missing information in a semantic map such as an \ac{ssg}, occupancy map, or semantic mesh~\cite{scenecompletionsurvey}. However, these methods do not make predictions beyond filling in missing information and do not present their use case in path planning tasks.

Compared to the current literature, this paper presents a new paradigm called Semantics-aware Predictive Planning (SPP) and contributes three main methodological contributions:
\begin{enumerate}
    \item An algorithm is presented to find repeating patterns of semantics in an \ac{ssg}. Specifically, this work exploits and extends the SUBDUE algorithm~\cite{Holder2002}, to find subgraphs that have multiple instances - exact or inexact - in an \ac{ssg} indicating a repeating pattern of semantics. The original work tackles the problem of finding structural and relational patterns in data represented as a graph. Key extensions to the SUBDUE algorithm are made to improve the inexact pattern detection by introducing a more accurate search strategy, and to exploit the pose of the vertices in the \ac{ssg}.
    \item Using the detected patterns a strategy to find areas in the \ac{ssg} that can potentially extend to the detected patterns is proposed.
    \item The paper presents two predictive path planning algorithms that exploit the developed \ac{ssg} prediction strategy for significantly improving the efficiency of robot inspections. Without loss of generality regarding the applicability of the pattern detection and graph prediction process, the predictive inspection planning methods are applied towards the task of visual inspection of ship ballast water tanks. The overall contribution is then extensively verified regarding its efficiency both in simulation and field experiments.
\end{enumerate}

\addition{An illustration of the components of this work is shown in Figure~\ref{fig:framework}. A video describing the proposed framework is available here: \url{https://youtu.be/uc1j6VgthM8}.}

%% file: 03_ProblemFormulation.tex
The overall problem considered in this work is that of the inspection of the surfaces of a set $\Theta$ of semantics of interest, distributed in an initially unknown volume $V$, using a camera sensor $\Ys_C$ --with \ac{fov} $[F^H_C, F^V_C]$-- at a maximum viewing distance $r_C$. The environment is represented as a) a discrete occupancy map $\Mbb$, built using the measurements from a depth sensor $\Ys_D$ (\ac{fov} $[F^H_D, F^V_D]$, max range $r_D$), consisting of cubical voxels $m \in \Mbb$ having edge length $\rho$, and b) a Semantic Scene Graph $\Gbb$, where each vertex $\nu$ represents a semantic in the environment and an edge $e$ represents the relationship between the vertices it connects. The part of $\Mbb$ consisting of occupied voxels seen by $\Ys_C$ is referred to as the \textit{seen map} and denoted by $\Mbb_C$. Similarly, the part consisting of occupied voxels seen by $\Ys_D$ and belonging to the semantics in $\Theta$ is referred to as the \textit{semantic map} and denoted by $\Mbb_S$.
Given the above, the problem at hand can be cast globally as that of seeing all possible voxels $m \in \Mbb_S$ in $V$ using $\Ys_C$ within the required viewing distance $r_C$.

\begin{definition}[Residual Semantic Surface]
    Let $\Xi$ be the set of all collision-free robot configurations, where a robot configuration $\xi \in \Xi$ consists of the robot's position $[x,y,z]$ in $3$D space and its yaw $\psi$. Let $\Xi_m \subset \Xi$ be the set of configurations from which the semantic voxel $m \in \Mbb_S$ can be seen by $\Ys_C$. The residual semantic surface is then defined as $\Mbb_S^{res} = \bigcup_{m\in \mathbb{M_S}} ( m \vert\ \Xi_m = \emptyset )$.
\end{definition}

\begin{problem} [Semantic Inspection]
    Given a volume $V$ and an initial configuration $\xi_0 = [x_0,y_0,z_0,\psi_0] \in \Xi$ find a collision-free path $\sigma$ that when traversed by the robot enables inspection of each voxel $m \in \Mbb_S \setminus \Mbb_S^{res}$ using the camera sensor $\Ys_C$.
\end{problem}

%% file: 04_PatternPrediction.tex
\begin{figure*}[h!]
\centering
    \includegraphics[width=0.99\textwidth]{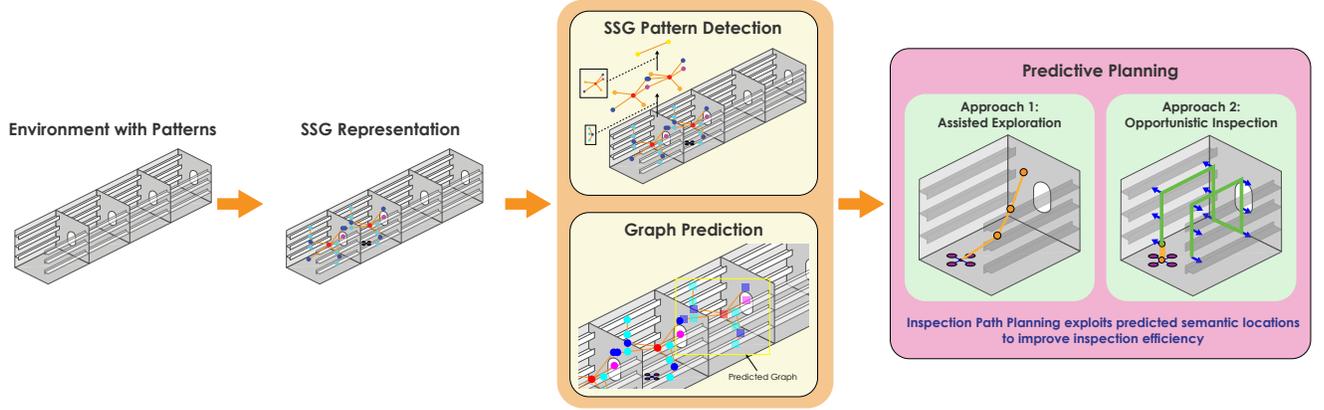}
\centering
\caption{\addition{This figure shows the various components of the proposed method and their interaction. The semantics in the environment are represented as a Semantic Scene Graph (SSG). The SSG pattern detection module aims to identify patterns in the SSG, which are then used by the graph prediction module to extend the scene graph into unknown space. The predicted semantic locations are used by the two proposed predictive planning algorithms to improve inspection efficiency. A video describing the proposed framework is available here: \url{https://youtu.be/uc1j6VgthM8}.}}
\label{fig:framework}
\end{figure*}

\addition{In the context of this paper, we tackle environments presenting spatially repeating patterns of the semantics of interest. Importantly, the method does not assume perfect pattern repetition and can tackle loosely matching patterns. A prime example of such a setting is the ballast water tanks as shown in Figure~\ref{fig:intro}. 
This work aims to utilize these patterns to predict the unknown parts of the environment, which can then be used to improve the inspection path planning efficiency. }

\addition{To enable the above, an efficient pattern detection methodology is needed that is robust to imperfect patterns. 
To this end, the semantics relevant to the mission are represented as an \ac{ssg} and a graph pattern detection algorithm, built on top of a graph-based knowledge discovery algorithm called SUBDUE~\cite{Holder2002}, is proposed. A pattern in an \ac{ssg} is defined as a subgraph that has multiple instances in the given \ac{ssg}. We present key modifications to SUBDUE for a) improving the inexact pattern detection by introducing a more accurate graph search technique and b) exploiting the pose of the vertices of the \ac{ssg}.}

\addition{Utilizing the detected patterns, a strategy to extend the \ac{ssg} in unknown parts of the environment is presented. 
We define a set of classes, such as doors, windows, and manholes inside ballast tanks, as \textbf{\ac{ecs}}. The vertices in the \ac{ssg} belonging to \ac{ecs}, called \textbf{\ac{evs}}, are used as anchor points for graph extension. 
The method aims to find \ac{evs} that do not belong to a pattern and can extend to one of the detected patterns.}

\subsection{Semantic Scene Graphs (SSG)}
In this work, an \ac{ssg} is defined as a directed graph $\Gbb$, with its vertex and edge sets $\Vs$, $\Es$ respectively, whose vertices represent semantics present in the environment and the edges represent relations between them. Each vertex $\nu \in \Vs$ has a label $\iota(\nu)$ indicating the class of the corresponding semantic. Similarly, each edge $e \in \Es$ has a label $\iota(e)$ stating the relation between the two vertices it connects. As a vertex represents an object in $3$D space, it has a pose associated with it given by $\xbf(\nu) = [\pbf_{\nu}, \Rbf_{\nu}]$, where $\pbf_{\nu} = [x,y,z]$ is the $3$D position of the geometric center of the object and $\Rbf_{\nu}$ is the rotation matrix representing the orientation of the object in an inertial frame w. Additionally, each vertex has a cuboidal bounding box $\bbf(\nu)$ encompassing the object represented by $\nu$.

\subsection{SSG Pattern Detection - Overview}
A pattern in an \ac{ssg} $\mathbb{G}$ is defined as a subgraph $\Sbb$ (also referred to as a substructure) that has multiple instances in $\mathbb{G}$. To find such substructures, we use and extend the SUBDUE algorithm~\cite{Holder2002} designed for discovering structural patterns in graph representations of data. SUBDUE uses the \ac{mdl} principle to identify a substructure $\Sbb$ that maximizes the compression of the original graph~$\mathbb{G}$. The compression is achieved by replacing each instance of the substructure with a single vertex. This process of substructure search and graph compression can be iteratively repeated to form a hierarchical representation $\Hs$ (as shown in Figure~\ref{fig:intro}) where each level $l^i$ includes the compressed graph $\Gbb^{l^i | \Sbb^i}$ and the identified substructure $\Sbb^i$. Throughout the paper, when we refer to the compressed version of a graph $\Gbb$ created using a substructure $\Sbb$ without referring to its level in the hierarchy, it is denoted as $\Gbb^{1|\Sbb}$.
To tackle minor discrepancies between instances of substructures, SUBDUE uses inexact graph-matching techniques to search for substructures whose instances are inexact isomorphs. Contrary to the exact graph-matching algorithms that return a Boolean value stating whether the two graphs under consideration are isomorphs, inexact graph-matching algorithms return a dissimilarity value (hereafter referred to as graph-matching cost) between the two graphs. This work uses the inexact graph-matching algorithm described in~\cite{Holder2002}.

First, we briefly summarize the SUBDUE algorithm from the original paper~\cite{Holder2002}, followed by detailed descriptions of the modifications done in this work. Algorithm~\ref{alg:subdue} details the steps. We provide a detailed video description of the \ac{ssg} pattern detection algorithm, including our modifications at \textbf{\url{https://youtu.be/WrqQu67gz1I}}. Let $\mathcal{V}, ~\mathcal{E}$ be the vertex and edge sets of $\mathbb{G}$, respectively. 
The algorithm begins by creating a priority queue $q$ having the vertices $\nu \in \mathcal{V}$ with unique label as the initial set of substructures. The priority of a substructure $\Sbb$ in $q$ is the compression value $\Gamma(\mathbb{G}, \Sbb)$, and the elements in the queue are sorted in ascending order. Therefore the first element $q[1]$ of $q$ has the lowest $\Gamma$ and represents the substructure that provides the most amount of graph compression.
The compression value $\Gamma(\Gbb, \Sbb)$ of a substructure $\Sbb$ for compressing the graph $\Gbb$ resulting in the compressed graph $\Gbb^{1|\Sbb}$ is defined as:


\small
\vspace{-1ex}
\begin{eqnarray}
 \Gamma(\Gbb,\Sbb) = \frac{DL(\Sbb) + DL(\Gbb^{1|\Sbb})}{DL(\Gbb)},
\end{eqnarray} \label{eqn:compression}
\normalsize
where $DL(\Sbb), DL(\Gbb^{1|\Sbb}), ~\textrm{and}~ DL(\Gbb)$ are the Description Lengths of the substructure, the compressed graph, and the original graph respectively. In this work, the Description Length of a graph is defined as the sum of the number of vertices and edges in the graph.

In each iteration of the algorithm, every instance $\mathbb{I}_{\Sbb}^i \in \Is_{\Sbb}$ ($\Is_{\Sbb}$ is the set of instances of $\Sbb$) of each substructure $\Sbb \in q$ is expanded in all possible ways by a single edge and the corresponding vertex, and the extended instances are regrouped into sets of inexact isomorphs to form the new set of substructures $\Ss_{ext}$ (line 7: \textbf{ExtendSubstruct}).
The compression value $\Gamma$ is calculated for each substructure (using the function \textbf{Evaluate}) and the substructure is added to another priority queue $q_{new}$. 
Only the first $\gamma_b > 0$ number of substructures in $q_{new}$ are retained using the $\textbf{mod}$ operator for computational tractability.
If $ \Gamma(\mathbb{G}, q_{new}[1]) < \Gamma(\mathbb{G}, \Sbb_{best})$, 
where $\Sbb_{best}$ is the substructure providing the lowest compression value found so far and $q_{new}[1]$ is the first element of $q_{new}$, then $\Sbb_{best}$ is updated to be $q_{new}[1]$. The old priority queue $q$ is replaced by the truncated $q_{new}$ and the process is repeated until either $q$ is empty or a maximum user-defined number of iterations $\gamma_l$ are carried out.

\begin{algorithm}
\caption{SUBDUE Substructure Discovery}
\label{alg:subdue}
\begin{algorithmic}[1]
\Function{SUBDUE}{$\mathbb{G}$, $\gamma_b$, $\gamma_l$} 
\State $q \gets \{ \nu \in \mathcal{V} | \nu ~\textrm{has unique label} \}$
\State $\Sbb_{best} \gets q[1]$
\While {$ q \neq \emptyset \AAND \gamma_l > 0$}
    \State $q_{new} \gets \{\}$
    \ForAll{ $\Sbb \in q$}
        \State $\Ss_{ext} \gets \textbf{ExtendSubstruct}(\Sbb, \mathbb{G})$
        \State $\textbf{Evaluate}(\Ss_{ext})$
        \State $q_{new} \gets (q_{new} \bigcup \Ss_{ext}) ~\textbf{mod}~ \gamma_b $
        \State $\gamma_l \gets \gamma_l - 1$
    \EndFor
    \If {$ \Gamma(\mathbb{G}, q_{new}[1]) < \Gamma(\mathbb{G}, \Sbb_{best})$}
        \State $\Sbb_{best} \gets q_{new}[1]$
    \EndIf
    \State $q \gets q_{new}$
\EndWhile
\State return $\Sbb_{best}$
\EndFunction
\end{algorithmic}
\end{algorithm}

\addition{To use SUBDUE effectively in the presented work, key feature additions and improvements are necessary. First, despite the inexact graph-matching formulation, the vanilla SUBDUE algorithm had difficulties identifying inexact patterns in sparse graphs presenting a high degree of imperfections (up to $31.25\%$ of instances having imperfection) in our study, as highlighted in Section~\ref{subsec:ablation}. This is due to the greedy graph search used to find inexact patterns. Furthermore, while calculating the graph-matching cost, SUBDUE treats all vertices equally. However, a mismatch at a vertex having a higher degree in the graph generally implies a higher dissimilarity between the two graphs.
Second, SUBDUE does not provide a way to utilize the vertex pose $\xbf(\nu)$ in the graph-matching procedure. This is particularly important for \ac{ssg}s as the vertices represent objects in $3$D space.}

\addition{To tackle these shortcomings, this work presents the following key modifications to the SUBDUE algorithm for substructure discovery:}

\begin{enumerate}
    \item Improvements in the \textbf{ExtendSubstruct} method: The extended substructure instance regrouping strategy inside the \textbf{ExtendSubstruct} function is improved by using a more accurate search as opposed to the original greedy search (more details in Section~\ref{subsec:mods_to_subdue}).
    \item Modifications to Inexact-Graph Matching: 
    \begin{enumerate}
        \item Since the proposed methods work with \ac{ssg} representing objects in $3$D space, an application not considered in the original work on SUBDUE, the pose $\xbf(\nu)$ of vertex $\nu$ is used in the inexact graph-matching algorithm.
        \item Given two graphs, the inexact graph-matching algorithm returns a cost relating to the transformations required to make them isomorphic. We scale the costs of each transformation by the degree of the vertices involved in the transformation to penalize discrepancies at vertices having a higher degree.
    \end{enumerate}
\end{enumerate}

\subsection{Modifications to SUBDUE} \label{subsec:mods_to_subdue}

\subsubsection{\textbf{Improvements in ExtendSubstruct method}}\label{subsub:extend_substruct}
The $\textbf{ExtendSubstruct}$ is an important function in the SUBDUE algorithm as it performs the key operations of extending and regrouping the substructures to form the new set of substructures. 
We first describe the modified version of the $\textbf{ExtendSubstruct}$ method which is detailed in Algorithm~\ref{alg:extendsubstruct}.
The method extends each instance $\Ibb_{\Sbb}^i \in \Is_{\Sbb}$ by a single edge $e$ (and the end vertex if not already in $\Ibb_{\Sbb}^i$) in all possible ways to form the set of extended subgraphs $\Ifk_{\Sbb}^i$ (lines $4-9$).
Let $\pazocal{L}_{\Sbb}$ be the set of all $\mathfrak{I}_{\Sbb}^i$ corresponding to $\Sbb$. 
The algorithm iteratively searches through the subgraphs in the sets $\Ifk_{\Sbb}^i \in \Ls_{\Sbb}$ to find the inexact isomorphs and group them together.
In each iteration, the first subgraph, denoted by $\Ibb_{+}$, in the first non-empty set $\Ifk^i_{\Sbb} \in \Ls_{\Sbb}$ is selected and added to the new substructure $\Ss_{+}$ (line $14$). 
Next, the graph-matching cost $t$ between $\Ibb_{+}$ and each element $\Ibb_{+}^\prime$ of the remaining sets $\Ifk^j_{\Sbb}, j \neq i$ is calculated. 
The inexact graph-matching algorithm (with modifications mentioned above) described in~\cite{Holder2002} is used in this work. 
The element $\Ibb_{best}^j$ with the lowest cost $t_{best}^j$ from each set $\Ifk^j_{\Sbb} \in \Ls_{\Sbb}, j \neq i$ is selected, removed from $\Ifk^j_{\Sbb}$, and added to $\Ss_{+}$ if $t_{best}^j \leq t_{thr}$ (where $t_{thr}$ is a user defined maximum allowed graph-matching cost).
This process is repeated until $\Ls_{\Sbb}$ is empty.

\begin{figure*}[h!]
\centering
    \includegraphics[width=0.99\textwidth]{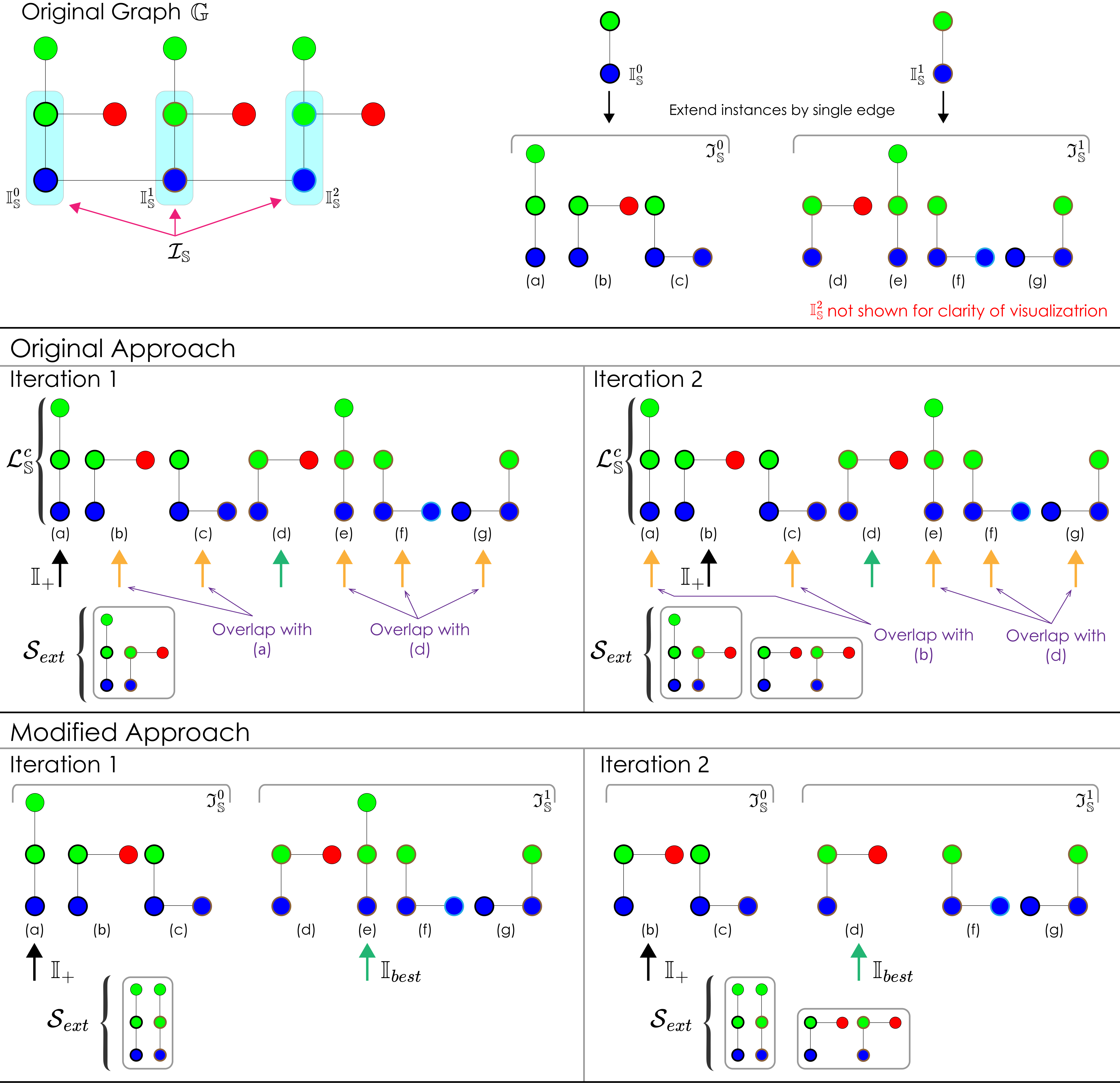}
\centering
\caption{This figure shows the improvement made to the \textbf{ExtendSubstruct} method through an example. Vertices shown in the same color belong to the same semantic class. During the extended instance regrouping step, the modified approach calculates the graph-matching cost between the selected instance and all other instances that can be its isomorphs and selects the best among them. On the other hand, the original algorithm selects the first non-overlapping instance, whose graph-matching cost is within the threshold, and adds it to the substructure group. This can result in an incorrect grouping as shown. However, due to the more accurate grouping approach, the modified algorithm can find the correct groupings.}
\label{fig:extend_substruct}
\vspace{-1ex} 
\end{figure*}

The original implementation of SUBDUE\footnote{https://ailab.wsu.edu/subdue/software/subdue-5.2.2.zip} performs the subgraph grouping differently. Figure~\ref{fig:extend_substruct} compares the original and modified approach at one instance of the algorithm. Note that the extended subgraphs of all instances of the substructure under consideration are not shown in the figure for clarity of visualization. We will briefly explain the original implementation with the help of the example in Figure~\ref{fig:extend_substruct}. Note that vertices shown in the same color belong to the same semantic class. The original approach does not maintain the subgraph sets $\Ifk_{\Sbb}^i$ according to the instances of $\Sbb$. Instead, all extended subgraphs are added to the same set $\Ls^c_{\Sbb}$. The method starts with the first element $\Ibb_{+} \in \Ls^c_{\Sbb}$ and adds it to $\Ss_{+}$ (subgraph `(a)' in Figure~\ref{fig:extend_substruct}).
The graph-matching cost $t$ is calculated between $\Ibb_{+}$ and the first element in $\Ls^c_{\Sbb}$ that does not overlap with any element of $\Ss_{+}$, which in this example is subgraph `(d)'. If $t \leq t_{thr}$, subgraph `(d)' is added to $\Ss_{+}$. The algorithm continues to find the next element not overlapping with any element of $\Ss_{+}$. Note that in this example, the subgraph`(e)' is overlapping with `(d)' and hence will not be evaluated even though it is an exact isomorph of `(a)'. This is contrary to the result in the modified approach (Figure~\ref{fig:extend_substruct}: Modified Approach) which correctly groups `(a)' and `(e)' together. Once the entire $\Ls^c_{\Sbb}$ is parsed, $\Ss_{+}$ is added to $\Ss_{ext}$ and the process is repeated with the next element in $\Ls^c_{\Sbb}$ (in this example `(b)' in Iteration 2).

\begin{algorithm}
\caption{Extend Substructures}\label{alg:extendsubstruct}
\begin{algorithmic}[1]
\Function{ExtendSubstruct}{$\Gbb$, $\Sbb$} 
\State $\Ss_{ext} \gets \emptyset$
\State $\Ls_{\Sbb} \gets \emptyset$
\ForAll{$\Ibb_{\Sbb}^i \in \Is_{\Sbb}$}
    \State $\Ifk_{\Sbb}^i \gets \emptyset$
    \ForAll{$e \in \Es ~|~ e$ is neighboring $\Ibb^i_{\Sbb}$}
        \State $\Ibb_{+} \gets \textbf{ExtendByEdge}(e, \Ibb_{\Sbb}^i, \Gbb)$
        \State $\Ifk_{\Sbb}^i \gets \Ifk_{\Sbb}^i \bigcup \Ibb_{+}$
    \EndFor
    \State $\Ls_{\Sbb} \gets \Ls_{\Sbb} \bigcup \Ifk_{\Sbb}^i$
\EndFor
\While{$\Ls_{\Sbb} \neq \emptyset$}
    \State $\Ifk_{\Sbb}^i \gets \textbf{PopFront}(\Ls_{\Sbb})$
    \While{$ \Ifk_{\Sbb}^i \neq \emptyset $}
        \State $\Ibb_{+} \gets \textbf{PopFront}(\Ifk_{\Sbb}^i) $
        \State $\Ss_{+} \gets \{ \Ibb_{+} \} $
        \ForAll{$ \Ifk_{\Sbb}^j \in \Ls_{\Sbb}, j \neq i $}
            \State $t_{best}^j \gets t_{thr} $
            \State $ \Ibb_{best}^j \gets \emptyset $
            \ForAll{$ \Ibb_{+}^\prime \in \Ifk_{\Sbb}^j $}
                \State $t \gets \textbf{\addition{GraphMatchingCost}}(\Ibb_{+}^\prime, \Ibb_{+})$
                \If{$t < t^j_{best}$}
                    \State $t^j_{best} \gets t$
                    \State $\Ibb_{best}^j \gets \Ibb_{+}^\prime$
                \EndIf
            \EndFor
            \State $\Ss_{+} \gets \Ss_{+} \bigcup \Ibb_{best}^j$
            \State $\Ifk_{\Sbb}^j \gets \Ifk_{\Sbb}^j \setminus \Ibb_{best}^j$
        \EndFor
        \State $\Ss_{ext} \gets \Ss_{ext} \bigcup \Ss_{+}$
    \EndWhile
\EndWhile
\State return $\Ss_{ext}$
\EndFunction
\end{algorithmic}
\end{algorithm}

\subsubsection{\textbf{Modifications to Inexact Graph Matching}}\label{subsub:inexact_graph_match}

For two graphs $\Gbb_1$ and $\Gbb_2$, the problem of inexact graph matching is that of finding a mapping $f : \Vs_1 \mapsto \Vs_2 \bigcup \{ \lambda \}$ --where $\Vs_1, \Vs_2$ are the vertex sets of graphs $\Gbb_1, \Gbb_2$ respectively-- that minimizes a cost function $C_{tot}$ relating to the transformations required to make $\Gbb_1$ and $\Gbb_2$ isomorphs. 
If $n(\Vs_1) \neq n(\Vs_2)$, where $n(A)$ denotes the number of elements in a set $A$, $\Gbb_1$ and $\Gbb_2$ are selected such that $n(\Vs_1) > n(\Vs_2)$ and the unmapped vertices from $\Vs_1$ are mapped to a virtual vertex $\lambda$ (called the null vertex).
A total of $7$ transformations are considered in this work, namely, the addition/deletion of a vertex, the addition/deletion of an edge, changing the label of a vertex or an edge, and changing the relative pose of a vertex.
It is noted that the original inexact graph-matching algorithm in~\cite{Holder2002} does not consider the vertex pose $\xbf(\nu)$. However, since this work operates on \ac{ssg} with vertices having pose information, this transformation has been added. The mapping $f$ is obtained through a tree search procedure as detailed in~\cite{Holder2002}. Since our contributions relate to the cost calculation, we refer the readers to the original paper~\cite{Holder2002} for further details on the algorithm to obtain the mapping $f$. 

The cost calculation is split into two parts. First, the cost related to the change in the relative pose of the vertices, referred to as the pose cost $C_P$, and second, the cost related to the remaining transformations, denoted by $C_R$.
The total cost $C_{tot}$ is then defined as:

\small
\vspace{-2ex}
\begin{eqnarray}
 C_{tot} = C_P + C_R
\end{eqnarray} \label{eqn:c_tot}
\vspace{-2ex}
\normalsize

\textit{\textbf{Modification a) Calculation of Pose Cost $C_P$}}: To calculate $C_P$, two point sets $\Ps_1$ and $\Ps_2$ corresponding to the positions of the vertices in $\Vs_1$ and $\Vs_2$ respectively, excluding those mapped to $\lambda$, are created. A rigid $3$D transform $T_{2,1}$ is calculated to align $\Ps_2$ to $\Ps_1$, with the mappings in $f$ as the correspondences, by solving the point set registration problem. Next, $\Ps_2$ is transformed using $T_{2,1}$ to get the point set $\Ps_{2,1}$ that is aligned with $\Ps_1$. 
Then the pose cost $C_P$ of $f$ is calculated as:

\small
\vspace{-1ex}
\begin{eqnarray}
    C_P &=& \gamma_p \sum_{\nu_i \in \Vs_1 | f(\nu_i) \neq \lambda} \frac{ d(\nu_i, f) }{d_{\max}} \\ \nonumber 
    d(\nu_i, f) &=&
    \begin{cases}
        0,&\textrm{if}~| \pbf_{\nu_i} - \pbf_{f(\nu_i)} | \leq d_{\min}~\\ 
        | \pbf_{\nu_i} - \pbf_{f(\nu_i)} |,&\textrm{if}~ d_{\min} \leq | \pbf_{\nu_i} - \pbf_{f(\nu_i)} | \leq d_{\max}~\\
        d_{\max} ,&\textrm{if}~ d_{\max} \leq | \pbf_{\nu_i} - \pbf_{f(\nu_i)} |
    \end{cases}
\end{eqnarray}
\normalsize

where $\pbf_{\nu_i}$ is the point in $\Ps_1$ corresponding to vertex $\nu_i \in \Vs_1$ and $\pbf_{f(\nu_i)}$ is the point in set $\Ps_{2,1}$ corresponding to the vertex $f(\nu_i) \in \Vs_2 \bigcup \{ \lambda \}$ which $\nu_i$ is mapped to. $\gamma_p > 0$ is a tunable weight for the pose cost. 
As real \ac{ssg} may contain noisy data causing the relative positions of the vertices across instances of the substructure to not be identical, a cost is added only if $| \pbf_{\nu_i} - \pbf_{f(\nu_i)} | \geq d_{\min}$ where $d_{\min}$ is a user-defined threshold. Similarly, a constant value $d_{\max}$ is assigned to $d(\nu_i,f)$ for all mappings satisfying $| \pbf_{\nu_i} - \pbf_{f(\nu_i)} | \geq d_{\max}$.

\textit{\textbf{Modification b) Calculation of $C_R$ and Cost Scaling based on Vertex Degree}}: Each transformation type $\tau^k, k=1...6$, except changing the relative pose of a vertex, is assigned a user-defined fixed cost called the transformation cost $c_{\tau^k}$.
To make the graphs $\Gbb_1$ and $\Gbb_2$ isomorphs, the vertices, and the edges connecting them, in each pair $\nu_i, f(\nu_i)$ in the mapping $f$ might need to undergo one of the above transformations.
This results in a cost $C_T(\nu_i)$ associated with the mapping of each vertex $\nu_i$. 

Let $C_{\tau^k}(\nu_i, f(\nu_i))$ be a function that returns a boolean stating if the mapping of vertex $\nu_i \in \Vs_1$ to the corresponding vertex $f(\nu_i) \in \Vs_2 \bigcup \{\lambda\}$, requires the transformation $\tau^k$ for the $\Gbb_1, \Gbb_2$ to be isomorphs.
%
The cost $C_T(\nu_i)$ of the mapping of vertex $\nu_i$ is then defined as:

\small
\vspace{-1ex}
\begin{eqnarray}
 C_T(\nu_i) = \sum_{k=1}^{6} c_{\tau^k} C_{\tau^k}(\nu_i, f(\nu_i))
\end{eqnarray} \label{eqn:C_T}
\normalsize

In the original SUBDUE algorithm, the total cost of a mapping $f$ resulting from the above transformations is the sum of the costs $C_T(\nu_i) \forall \nu_i \in \Vs_1$. However, an incorrect mapping at a vertex having higher degree in the graph implies a higher discrepancy. Hence, each cost $C_T(\nu_i)$ is scaled with a parameter, $\gamma_{d,i}$, related to the degrees of the vertices $\nu_i, f(\nu_i)$.
This parameter is defined as $\gamma_{d,i} = 1+\frac{\eta}{\eta_{\max}}$, where $\eta$ is the highest degree among $\nu_i, f(\nu_i)$ and $\eta_{\max}$ is the highest degree of any vertex in $\Gbb_1$ and $\Gbb_2$. 
Therefore, the cost $C_R$ is calculated as

\small
\vspace{-1ex}
\begin{eqnarray}
 C_R = \sum_{\nu_i \in \Vs_1} \gamma_{d,i} C_T(\nu_i)
\end{eqnarray} \label{eqn:C_R}
\normalsize

\begin{figure}[h!]
\centering
    \includegraphics[width=0.9\columnwidth]{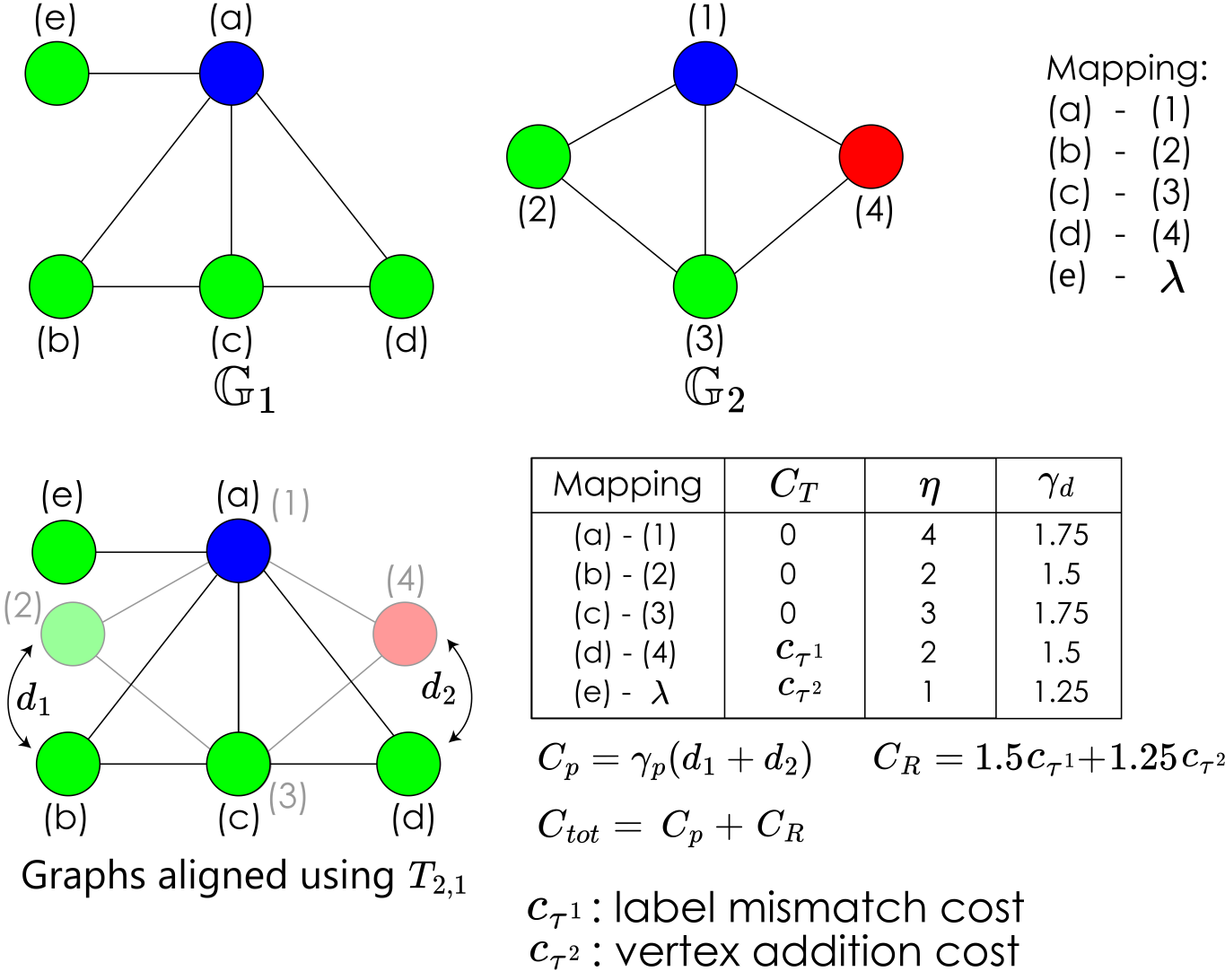}
\centering
\caption{This figure shows an example of the graph-matching cost calculation between two graphs for a given mapping $f$. Vertices shown in the same color belong to the same semantic class. The mapping $f$ is shown in the top right corner. The transformations involved in this mapping are: label mismatch ((d) - (4)) and vertex addition ((e) - $\lambda$). Each cost $C_T(\nu_i)$ for vertex $\nu_i \in \Vs_1$, $\Vs_1$ is the set of vertices of $\Gbb_1$, is scaled by the factor $\gamma_{d,i}$ related to the degree of the relevant vertices.}
\label{fig:match_cost}
\vspace{-2ex} 
\end{figure}

Figure~\ref{fig:match_cost} shows an example of the cost calculation. 



\subsection{Scene Graph Prediction using Patterns}\label{subsec:graph_pred}

The aim of graph prediction is to identify areas in the \ac{ssg} that can extend to one of the identified patterns. To this end, the proposed approach utilizes vertices whose labels belong to specific classes, hereafter referred to as `\textbf{\ac{ecs}}', as anchor points for graph extension. These classes correspond to objects such as manholes in ballast tanks, doors or windows in a building, etc. The vertices having their label as an \ac{ec} are referred to as \textbf{Entry Vertices} $\vartheta^e$. The algorithm operates on each level of the hierarchy whose corresponding substructure contains at least one $\vartheta^e$. The algorithm attempts to find $\vartheta^e$ where the substructure can fit with no overlap between vertices of different classes and maximum overlap between vertices of the same classes.

The graph prediction takes place in two steps. First, in the graphs in each level of the hierarchical structure, extra vertices called Complementary Entry Vertices $\Bar{\vartheta}^e$ are added corresponding to each $\vartheta^e$. $\Bar{\vartheta}^e$ have the same label and pose as the corresponding $\vartheta^e$. The $\Bar{\vartheta}^e$ addition process is illustrated in Figure~\ref{fig:modify_graph}.
Only the levels that have substructures containing $\vartheta^e$ are considered for graph prediction. 
Second, the algorithm attempts to fit the substructure at one of the $\vartheta^e$ or $\Bar{\vartheta}^e$. This process is illustrated in Figure~\ref{fig:graph_evoltion}.
A detailed video description of the graph prediction is available at \textbf{\url{https://youtu.be/b6-3C_rKdiY}}.

\subsubsection{\textbf{Complementary Entry Vertex Addition}} 
\begin{figure*}[h!]
\centering
    \includegraphics[width=0.99\textwidth]{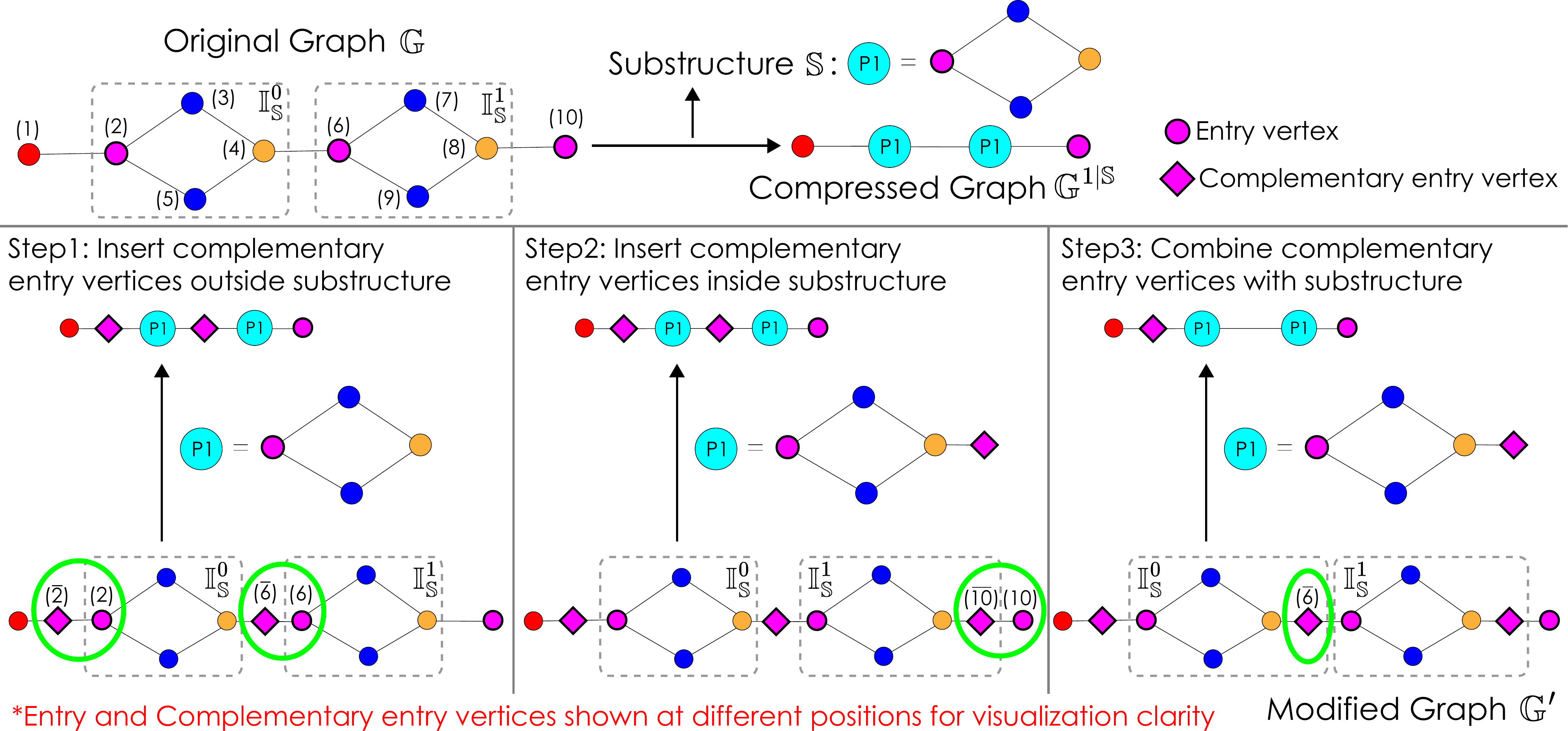}
\centering
\caption{This figure shows the steps involved in the modification of the \ac{ssg} for graph prediction. Vertices having the same color have the same semantic label. The process inserts additional vertices called Complementary Entry Vertices $\Bar{\vartheta}^e$ for each Entry Vertex $\vartheta^e$. The three steps involved in this process are illustrated along with the final modified graph $\Gbb^\prime$.}
\label{fig:modify_graph}
\end{figure*}

The subsequent section details the process of $\Bar{\vartheta}^e$ addition for one level using the example in Figure~\ref{fig:modify_graph}. In `Step 1', if an instance $\Ibb_{\Sbb}$ of substructure $\Sbb$ contains a $\vartheta^e$, a $\Bar{\vartheta}^e$ is added at the same pose but not included in $\Ibb_{\Sbb}$ (e.g., `($2$)-($\Bar{2}$)', `($6$)-($\Bar{6}$)' in Figure~\ref{fig:modify_graph}). The edges connected to that $\vartheta^e$ that are part of $\Ibb_{\Sbb}$ remain connected to $\vartheta^e$. The remaining edges are connected to $\Bar{\vartheta}^e$. An edge is added between $\vartheta^e$ and $\Bar{\vartheta}^e$. This process is repeated for each $\vartheta^e$ in all $\Ibb_{\Sbb} \in \Sbb$ in that level. It is highlighted that the position of the $\vartheta^e$ and $\Bar{\vartheta}^e$ are different in Figure~\ref{fig:modify_graph} only for visualization clarity. In `Step 2', for any instance $\Ibb_{\Sbb}$ that connects to a $\vartheta^e$ that is not part of it such that the neighbor of that $\vartheta^e$ inside $\Ibb_{\Sbb}$ is not a $\Bar{\vartheta}^e$ (e.g., vertex `($10$)'), a $\Bar{\vartheta}^e$ is added inside $\Ibb_{\Sbb}$ in the same way as described above. In `Step 3', for any instance $\Ibb_{\Sbb}$ that connects to a $\Bar{\vartheta}^e$ that is not part of it such that its neighbor inside $\Ibb_{\Sbb}$ is not a $\vartheta^e$ (e.g., vertex `($4$)' connecting to `($\Bar{6}$)'), this $\Bar{\vartheta}^e$ is included into $\Ibb_{\Sbb}$ (e.g., `($\Bar{6}$)' is included in $\Ibb^0_{\Sbb}$). 
The graph resulting from the above modification is referred to as the \textbf{Modified Graph} ${\Gbb}^\prime$.

\subsubsection{\textbf{Graph Prediction}} Once the graph modification is done, $\vartheta^e$ and $\Bar{\vartheta}^e$ in $\Gbb^\prime$ not part of any instance of a substructure are selected as potential vertices that can extend to one of the substructures. These vertices are referred to as \textbf{Loose Entry Vertices} $^L\vartheta^e$ or \textbf{Loose Complementary Entry Vertices} $^L\Bar{\vartheta}^e$. 
We now explain the procedure to select the best $^L\vartheta^e$ in the graph for extension. The procedure is described only for $^L\vartheta^e$ as the procedure for $^L\Bar{\vartheta}^e$ is identical. For a level $l$ - with the graph $\Gbb$ and the corresponding substructure $\Sbb$ - of the hierarchical representation $\Hs$, let $^L\Vs^e$ be the set of all $^L\vartheta^e$ in $\Gbb$ and $\Vs^e_{\Sbb}$ be the set of all $\vartheta^e$ in $\Sbb$. Each vertex $^L\vartheta^e_j \in ~^L\Vs^e$ is checked if it can be extended to $\Sbb$ by calculating an overlap score. 
For each $\vartheta^e_{i,\Sbb} \in \Vs^e_{\Sbb}$, the $\vartheta^e_{i,\Sbb}$ and $^L\vartheta^e_j$ are aligned by setting $\xbf(\vartheta^e_{i,\Sbb}) = \xbf(^L\vartheta^e_j)$ and the remaining vertices in $\Sbb$ are transformed accordingly.
The overlap between the bounding boxes of the transformed vertices of $\Sbb$ with those of the vertices in $\Gbb$ is computed. For each bounding box $\bbf(\nu_{k,\Sbb}), \nu_{k,\Sbb} \in \Vs_{\Sbb}$, where $\Vs_{\Sbb}$ is the set of vertices of $\Sbb$, overlapping with a vertex in $\Gbb$ having the same label as $\nu_{k,\Sbb}$, the overlap score for $\vartheta^e_{i,\Sbb}$ is increased by $1$. However, if it overlaps with a vertex that is part of an instance of a substructure or has a different label, this candidate is rejected and the process is continued with the next $\vartheta^e_{i,\Sbb}$. The $\vartheta^e_{i,\Sbb}$ with the highest overlap score is the prediction candidate for $^L\vartheta^e_j$. This process is repeated and the best prediction candidate for each $^L\vartheta^e$ is calculated. The candidates having the top $\varphi$ percentage of overlap scores are kept and returned as the graph prediction. 

\begin{figure*}[h!]
\centering
    \includegraphics[width=0.99\textwidth]{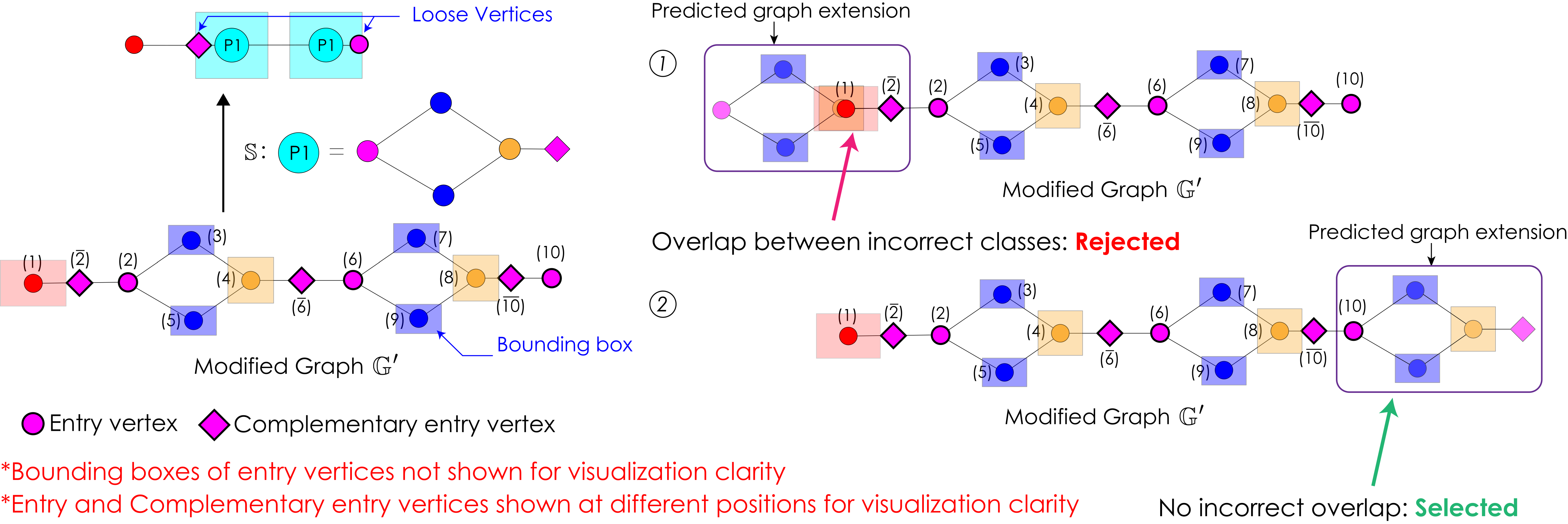}
\centering
\caption{This figure shows an example of the graph prediction process using the detected patterns. The method uses the $^L\vartheta^e$ and $^L\Bar{\vartheta}^e$ in the modified graph $\Gbb^\prime$ to find the vertex which can extend to the substructure with no overlap between vertices of different classes and maximum overlap between vertices of the same classes. The overlap check at the two candidate vertices is shown in the figure.}
\label{fig:graph_evoltion}
\end{figure*}

Figure~\ref{fig:graph_evoltion} illustrates an example of the process of overlap checking for $^L\vartheta^e$. The modified graph in Figure~\ref{fig:modify_graph} is used in this example. 
Two loose vertices, `($\Bar{2}$)' (a $^L\Bar{\vartheta}^e$) and `($10$)' (a $^L\vartheta^e$), exist in the graph. The substructure `P1' has both an $\vartheta^e$ and a $\Bar{\vartheta}^e$. Hence both of these locations are possible candidates that can extend to `P1'. In subfigure 1 (top right), the algorithm tries to extend the graph through `($\bar{2}$)' but the bounding box of one of the vertices of the predicted subgraph graph overlaps with vertex `($1$)' which does not have the same class. Hence, this prediction is rejected. The algorithm then tries at vertex `($10$)'. Here we see no incorrect overlap and this prediction is selected.

%% file: 05_PredictivePlanning.tex
Without loss of generality regarding the concept of \ac{ssg}, the detection of patterns and their exploitation in planning, this work focuses on ballast tanks inside ships. Ballast tanks represent industrial environments of great importance and present clear repeating patterns in the semantics of interest such as their longitudinal structures as shown in Figure~\ref{fig:intro}. Functionality-wise, ballast tanks are filled with water to adjust the buoyancy of the ship. Due to constant exposure to salt water, the tanks need to be inspected regularly for corrosion, cracks, and deformation. Specific structures, such as longitudinals, inside the ballast tanks are more important for inspection. The ballast tanks are divided into multiple levels with each level consisting of a number of compartments often connected by very narrow openings (as narrow as $\SI{0.6}{\meter} \times \SI{0.4}{\meter}$) called manholes. In this work, we tackle the problem of autonomous inspection the longitudinals. As Figure~\ref{fig:intro} shows, the longitudinals form a spatially repeating pattern across the ballast tank compartments. 

Motivated by the above, we first present an \ac{ssg} generation method to represent the longitudinals and other semantics in the ballast tank as a scene graph.
Second, a volumetric exploration and inspection strategy is presented for inspecting the longitudinals in the ballast tank. Finally, we describe two approaches that exploit the predicted extensions of the scene graph to make the inspection planning more efficient.

\subsection{Ballast Tank Scene Graph Generation} \label{subsec:ssg_gen}
\begin{figure}[h!]
\centering
    \includegraphics[width=0.9\columnwidth]{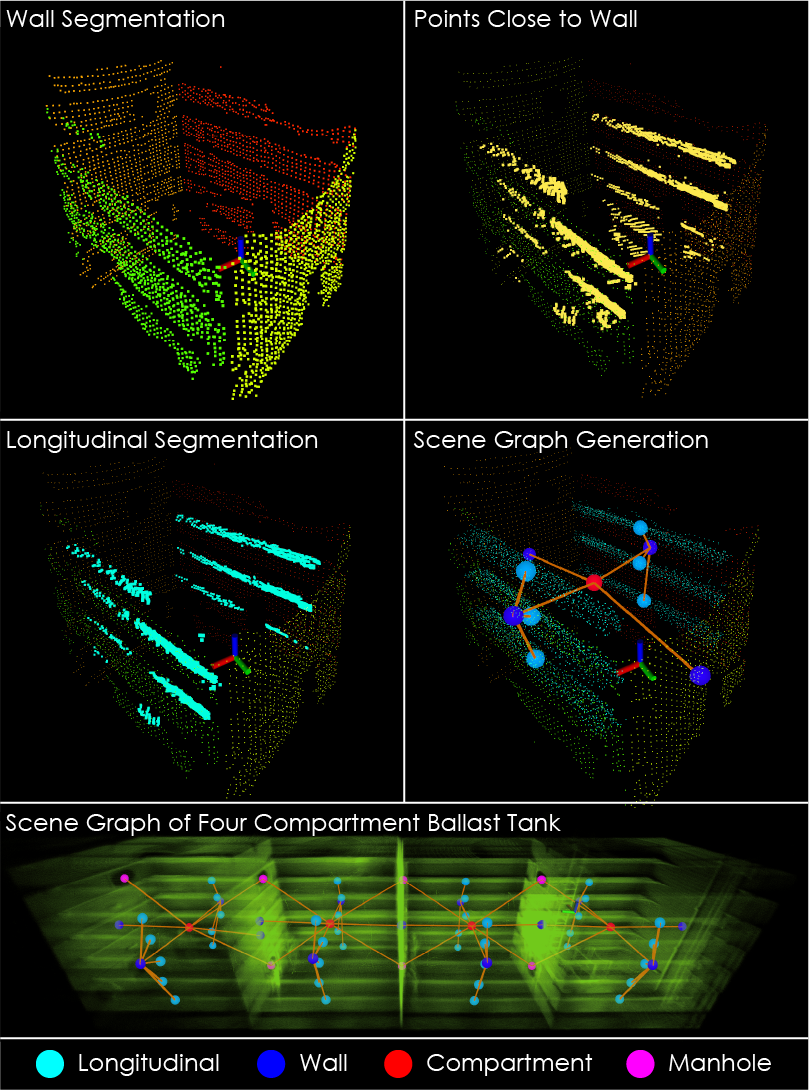}
\centering
\caption{Figure showing different steps in the \ac{ssg} generation in real data. The top four subfigures show the data from one LiDAR scan showing the extracted wall planes, part of the point cloud close to the wall planes, lines extracted within this point cloud, and the current state of the \ac{ssg}. The subfigure on the bottom shows the final \ac{ssg} and map at the end of the mission.}
\label{fig:ssg_gen}
\end{figure}

In this work, we include the following semantics in the scene graphs: 1) longitudinals, 2) walls, 3) compartments, and 4) manholes. The segmentation happens on the point cloud $\Ps$ from a 3D LiDAR sensor. In this work, the manhole is considered as the `Entry Class'. Figure~\ref{fig:ssg_gen} shows the different steps in the \ac{ssg} generation procedure.
\subsubsection{\textbf{Wall Segmentation}}
For each point cloud $\Ps$, the four largest (based on the number of points) vertical planes, hereafter referred to as wall planes, are segmented using the RANSAC algorithm. The normal of the plane is set to point towards the current robot location. 
Each newly detected wall plane $\textrm{P}_i, i=1...4$ is checked against all elements of the set $\Ws$ of previously detected walls to find a wall $\textit{W}_i \in \Ws$ that satisfies the following criteria: a) is coplanar to $\textrm{P}_i$, b) has its normal pointing in the same direction, and c) has its centroid on the same side of all walls $\textit{W}_j \in \Ws, j \neq i$ as the centroid of the point cloud belonging to $\textrm{P}_i$.
If such a wall is found, the point clouds of this wall and $\textrm{P}_i$ are combined, sub-sampled to a desired resolution, the new centroid is calculated, and the detection count for that wall is incremented. 
Otherwise, a new wall entry is added to $\Ws$. When the detection count of a wall goes above a threshold $n_{thr}^W$, a new vertex is added to the \ac{ssg} at the location of the centroid of the wall with its orientation along the normal of the plane of the wall.
\subsubsection{\textbf{Compartment Segmentation}}
A compartment is defined as the cuboidal volume between the four wall planes detected during the wall segmentation step. The position of the compartment is set to be the centroid of the centroids of these four planes.
Similar to the wall segmentation, the detected compartment locations are tracked and stored in a set $\mathcal{Q}$. The algorithm attempts to find a compartment whose position is on the same side of all walls in $\Ws$ as the newly detected compartment $\textit{Q}$. If such a compartment is found, its position is updated to be the mean of its current position and the position of $\textit{Q}$, and its detection count is incremented. Else, a new compartment entry is added to $\mathcal{Q}$. When the detection count of a compartment goes above a threshold $n_{thr}^Q$, a new vertex is added to the \ac{ssg}. 
Once a compartment vertex is added, edges are added between it and the wall vertices whose centroids and the compartment's position lie on the same side of all other walls in $\Ws$.

\subsubsection{\textbf{Longitudinal Segmentation}}
It can be seen from Figure~\ref{fig:intro} that the longitudinals are always attached to walls and are parallel to each other. Exploiting these facts, longitudinal segmentation takes place only in the part of $\Ps$ close to the detected walls. First, the points $\Ps_i^L$ from the input $\Ps$ that lie within a distance $d_w$ from the planes $\textrm{P}_i, i=1...4$ are extracted. 
A set $\mathbb{L}_i$ of parallel lines is segmented out from each $\Ps_i^L$ as candidate longitudinals.
Each longitudinal is characterized by the equation of the line corresponding to it and the centroid of the points belonging to the line. The equations of the newly detected lines $\mathbb{L}_i$ are compared against that of the longitudinals $\textit{L}_j$ in the set $\mathfrak{L}$ of previously detected longitudinals. For a line $l_i^k \in \mathbb{L}_i$, the algorithm attempts to find a longitudinal with a matching line equation and its centroid lying on the same side of all walls in $\Ws$ as the centroid of $l_i^k$. If such a longitudinal is found, its centroid is updated to be the mean of its current centroid and the centroid of $l_i^k$, and its detection count is incremented.
Else, a new longitudinal entry is created. When the detection count of a longitudinal goes above a threshold $n_{thr}^L$ (a tunable parameter), a new vertex is added to the \ac{ssg}, and an edge is added between it and the corresponding wall.

\subsubsection{\textbf{Manhole Segmentation}}
We utilize the work from~\cite{manhole2023} for manhole segmentation. Once a manhole is detected, a new vertex is added to the graph and an edge is added between it and the closest two compartment vertices.

\subsection{Predictive Planner Overview}\label{subsec:planner_overview}
As the ballast tank is divided into multiple compartments, the planner tackles the problem of exploration and longitudinal inspection one compartment at a time. 
The planner starts with no information about the ballast tank except the number of compartments to inspect. It operates in three planning modes namely \ac{ve}, \ac{si}, and \ac{pp}. At the beginning of the mission, the planner starts in the \ac{ve} mode, which aims to map the entire compartment the robot is currently in with $\Ys_D$. Upon completion, the \ac{si} mode calculates viewpoints and paths to view the part of $\Mbb_S$ within that compartment with $\Ys_C$ at a distance $r_C$. Note that the semantic segmentation and scene graph generation is taking place in real-time during the entire mission.

\begin{figure}[h!]
\centering
    \includegraphics[width=0.9\columnwidth]{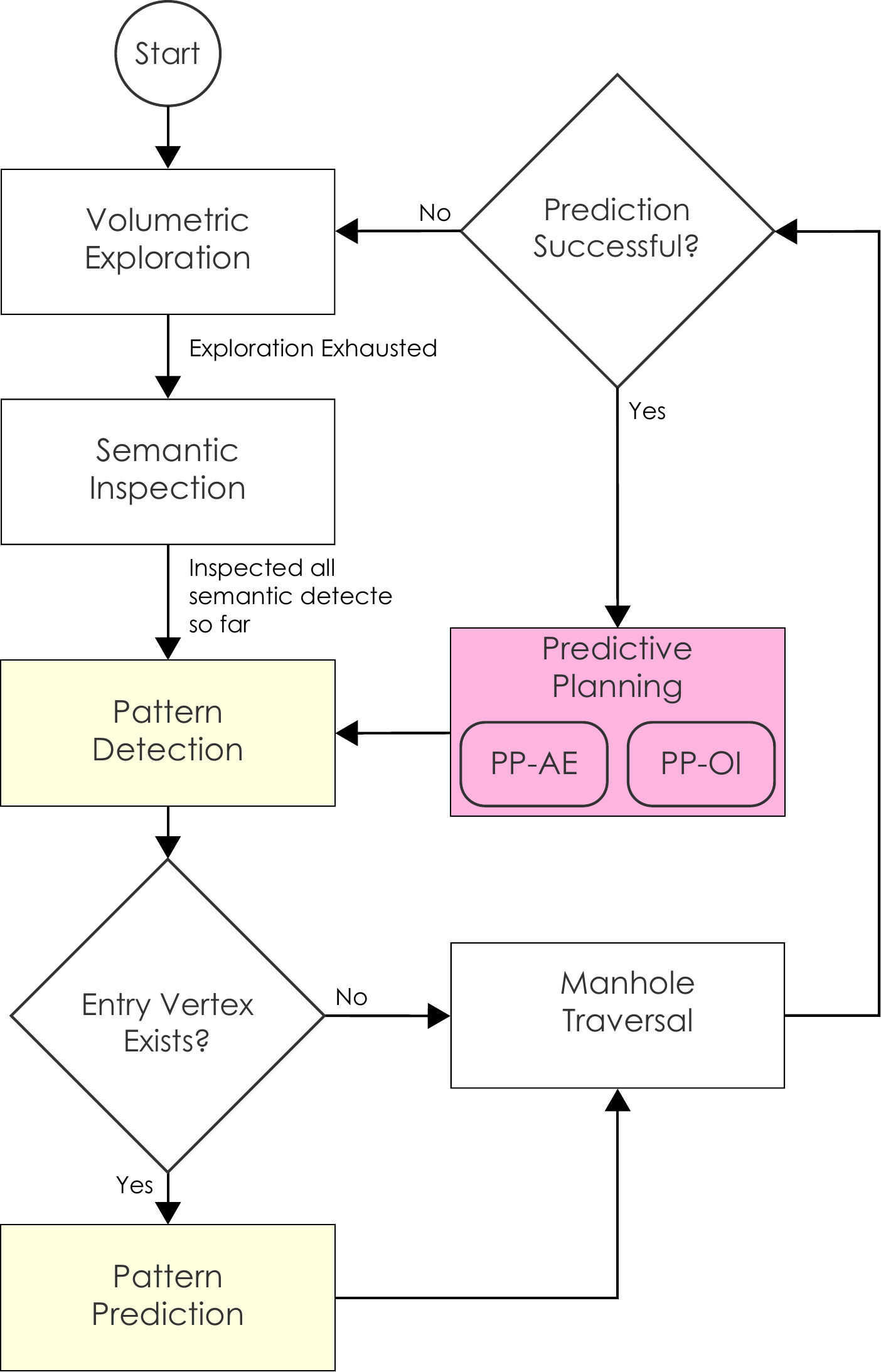}
\centering
\caption{\addition{Figure showing the flow diagram of the predictive planner.}}
\label{fig:planner_flow}
\end{figure}

After all detected longitudinals in the current compartment are inspected, the pattern detection and graph prediction steps (described in Section~\ref{sec:pat_pred}) are triggered. 
If a successful prediction is made and a feasible path - calculated using the method described in~\cite{manhole2023} - for traversing the manhole (the entry vertex) used for prediction exists, the robot is commanded to go through this manhole. Otherwise, the robot is guided to traverse through the closest untraversed manhole.
In the first case, the \ac{pp} mode is triggered which utilizes the predicted longitudinal locations to improve the exploration and inspection efficiency. We present two approaches (hereafter referred to as submodes) for this mode. The first approach called \ac{ae}, guides the robot towards predicted longitudinal locations as the robot goes through the manhole used for prediction. The second approach called \ac{oi}, utilizes the viewpoint sequence (hereafter referred to as predicted path) from the \ac{si} mode in previous compartments and follows the same sequence (with necessary collision-free path planning) as long as the longitudinal corresponding to the viewpoints in the sequence are detected. \addition{The flow diagram of the planner is presented in Figure~\ref{fig:planner_flow} and a} detailed video description is available at \textbf{\url{https://youtu.be/XDTPodpD4SE}}.
The subsequent subsections detail each planning mode.

\subsection{Volumetric Exploration (VE)} \label{subsec:ve}
In this paper, we use our previous open-sourced work on Graph-based Exploration Path Planning~\cite{GBPLANNER_JFR_2020,GBPLANNER2COHORT_ICRA_2022}, called GBPlanner, for the \ac{ve} mode. 
GBPlanner operates in a bifurcated local-global planning architecture, where the local planner is responsible for providing efficient, collision-free paths for mapping unknown space within a local volume around the robot. On the other had, the global planner is used to reposition the robot to frontiers of unexplored space when the local planner is unable to provide informative paths, as well as provide return-to-home functionality when the robot's battery reaches its endurance limit.

The \ac{ve} mode utilizes the local planner of GBPlanner. In each exploration iteration, the planner builds a $3$D collision-free graph $\Gbb_{VE}$ within a local volume around the robot. Next, an information gain $\Upsilon_{VE}(\nu_j)$ called Volume Gain is calculated for each vertex $\nu_j, j=1...\beta$ ($\beta$ is the number of vertices in $\Gbb_{VE}$) in $\Gbb_{VE}$ that relates to the amount of unknown volume the robot would see if it was at $\nu_j$. The algorithm then calculates the shortest paths $\sigma_j$ from the vertex $\nu_1$ corresponding to the current robot location to each vertex $\nu_j$ in $\Gbb_{VE}$ using Dijkstra's algorithm and computes the accumulated Exploration Gain $\Lambda_{VE}(\sigma_j)$ for each path $\sigma_j$ as:

\small
\begin{eqnarray}\label{eqn:lambda_ve}
 \Lambda_{VE}(\sigma_j) = e^{-\zeta \mathcal{Z}(\sigma_j,\sigma_{e})} \sum_{i=1}^{\beta_{j}}{\Upsilon_{VE}(\nu_{i}) e^{-\mu \mathcal{D}(\nu_{1},\nu_{i})}}.
\end{eqnarray}
\normalsize

Here, $\Ds(\nu_{1},\nu_{i})$ is length of the shortest path from $\nu_1$ to $\nu_i$ along $\Gbb_{VE}$. The function $\Zs(\sigma_j,\sigma_{e})$ calculates a similarity metric between the path $\sigma_j$ and a straight line path $\sigma_e$ along the current exploration direction having the same length penalizing a change in the exploration direction. The constants $\zeta>0$ and $\mu>0$ are tunable parameters, and $\beta_j$ is the number of vertices in $\sigma_j$.
The path $\sigma_{best}$ with the highest Exploration Gain is selected and executed by the robot. Upon completion, the process is repeated until $\Upsilon_{VE}$ of all the vertices in $\Gbb_{VE}$ is below a user defined threshold $\Upsilon_{VE,\min}$ at which point the volumetric exploration is complete.
Instead of switching to the global planning mode, the planner then switches to the \ac{si} mode for inspection of the semantics.

\begin{figure}[h!]
\centering
    \includegraphics[width=0.9\columnwidth]{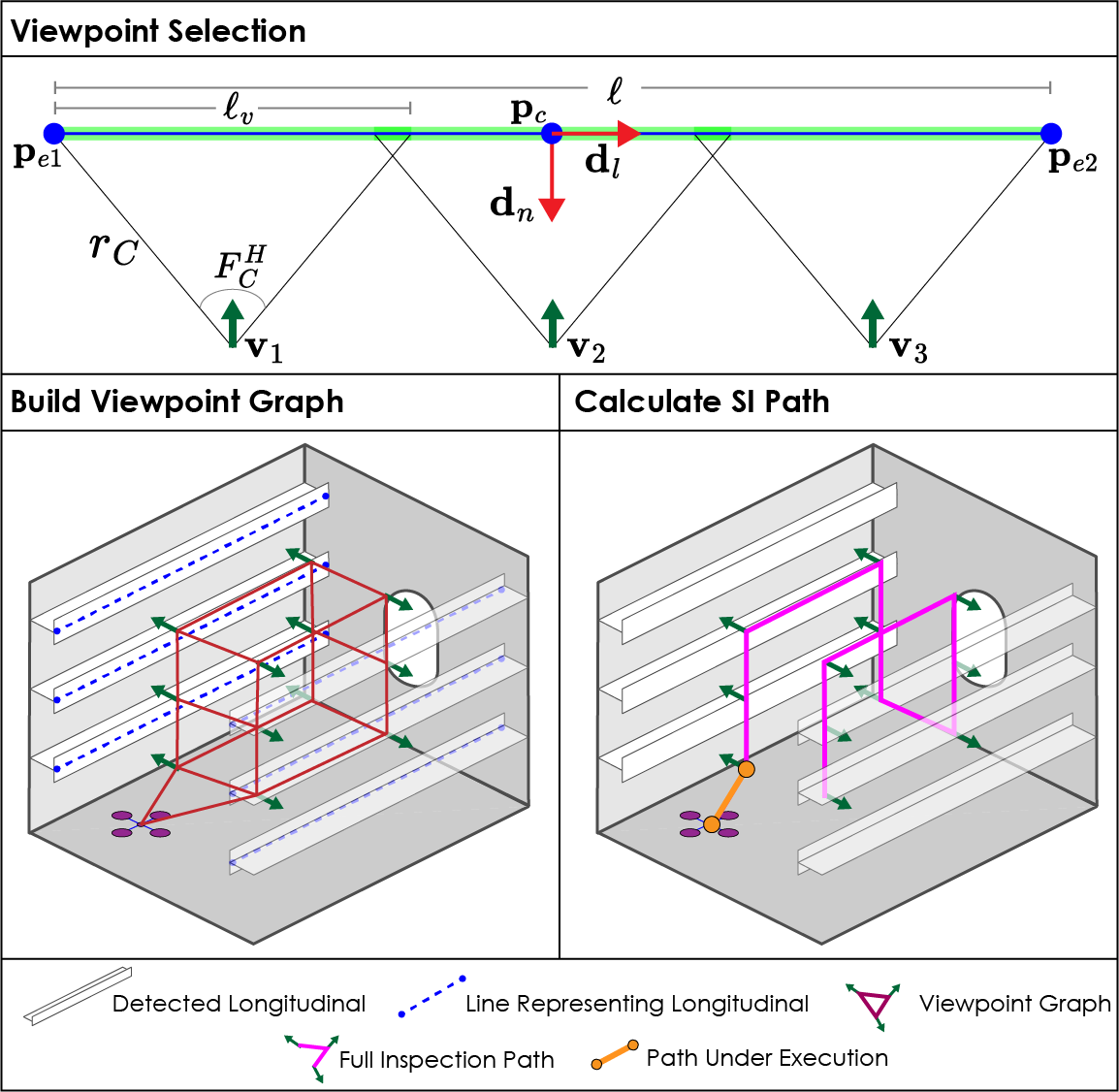}
\centering
\caption{This figure illustrates the steps involved in the Semantic Inspection mode. The planner calculates uniformly distributed viewpoints along each longitudinal respecting the \ac{fov} and max range constraints of $\Ys_C$. A collision-free graph is built connecting the viewpoints and a tour passing through all viewpoints is calculated. Both of these steps are shown in the bottom two subfigures.}
\label{fig:si}
\end{figure}

\subsection{Semantic Inspection (SI)}\label{subsec:si}
The \ac{si} mode aims to find viewpoints and paths to view the voxels in $\Mbb_S$ with the camera sensor $\Ys_C$. Given the maximum viewing distance $r_C$ and the \ac{fov} of $\Ys_C$ as $[F^H_C,F^V_C]$ radians the viewpoint calculation procedure is as follows. For a longitudinal $\textit{L}$, let $\abf = [d_l^x, d_l^y, d_l^z, p_c^x, p_c^y, p_c^z]$ be the coefficients representing the equation of the line $l_{\textit{L}}$ corresponding $\textit{L}$. The unit vector $\dbf_l = [d_l^x, d_l^y, d_l^z]$ represents the direction of the line and the point $\pbf_c = [p_c^x, p_c^y, p_c^z]$ is a point on the line in the inertial frame of reference. In this work, $\pbf_c$ is selected to be the centroid of the point cloud corresponding to the longitudinal. Let $\dbf_n$ be the unit vector orthogonal to $\dbf_l$ pointing along the normal of the wall to which $\textit{L}$ is attached to. Let $\pbf_{e1}$ and $\pbf_{e2}$ be the two endpoints of the longitudinal - both lying on $l_{\textit{L}}$ - such that $\dbf_l$ points from $\pbf_{e1}$ to $\pbf_{e2}$. The planner selects equally spaced viewpoints along $l_{\textit{L}}$ at a distance $r_C \cos(\frac{F_C^H}{2})$ from it as shown in Figure~\ref{fig:si}. The number of viewpoints $n$ is given by $n = \lceil \frac{\ell}{\ell_v} \rceil$, where $\ell = | \pbf_{e2} - \pbf_{e1} |$ is the length of the longitudinal and $\ell_v = 2 r_C \sin(\frac{F^H_C}{2})$ is the length of the section of the longitudinal that a viewpoint will see such that the farthest point viewed is at a distance $r_C$.
Then the position of the viewpoint $\vbf_i, i=1...n$ is calculated as 

\small
\vspace{-2ex}
\begin{eqnarray}
    \vbf_i = \pbf_{e1} + \frac{(2i-1)\ell_v}{2} \dbf_l + r_C \cos(\frac{F^H_C}{2}) \dbf_n
\end{eqnarray}
\normalsize

The heading of the viewpoints is set to be along $-\dbf_n$. 
The viewpoints that don't view any unseen parts of $\Mbb_S$ or have overlap with other viewpoints are removed.
Once the viewpoints for all longitudinals are calculated, the planner attempts to connect them and the current robot location using collision-free straight line edges to form a graph $\Gbb_{SI}$. If any viewpoints are not connected to the graph, extra points are uniformly sampled in the free space to connect the remaining viewpoints to $\Gbb_{SI}$. 
Next, the tour $\sigma_{tsp}$ visiting all viewpoints is calculated by solving the \ac{tsp} using the Lin-Kernighan-Helsgaun (LKH)~\cite{helsgaun2000effective} heuristic. The travel cost between two viewpoints is given by the length of the shortest path along $\Gbb_{SI}$ between them.
The path, along $\Gbb_{SI}$ up to the first viewpoint in the tour is executed by the robot and the viewpoints are recalculated. If any viewpoints are added/deleted or their location has changed more than a threshold $\varrho_{thr}$, a new graph $\Gbb_{SI}$ is built and a new tour is calculated. Otherwise, the path up to the next viewpoint in the tour is commanded to the robot. Figure~\ref{fig:si} shows an illustration of the steps involved in the \ac{si} mode.

\subsection{Predictive Planning - Assisted Exploration (PP-AE)}
The \ac{ae}, the first submode of \ac{pp}, builds upon the \ac{ve} mode and modifies the information gain to guide the planner to focus on the predicted semantic locations. 
Using the transform used to align the $\vartheta^e$ or $\Bar{\vartheta}^e$ of the substructure with the selected $^L\vartheta^e$ or $^L\Bar{\vartheta}^e$, the bounding boxes of the longitudinals in the predicted graph are transformed to be at the predicted longitudinal locations. The part of the map $\Mbb$ lying in these bounding boxes is referred to as the predicted semantic map and is denoted by $\Mbb_S^P$.
The \ac{ae} submode operates in an iterative fashion similar to \ac{ve} and, in each iteration, builds a collision-free graph $\Gbb_{AE}$ around the current robot location. It then calculates a new information gain $\Upsilon_{AE}(\nu_j)$ for each vertex $\nu_j$ in $\Gbb_{AE}$. The remaining procedure to select the best path remains the same as \ac{ve} with $\Upsilon_{VE}$ being replaced by $\Upsilon_{AE}$ in Equation~\ref{eqn:lambda_ve}. The new information gain $\Upsilon_{AE}(\nu_j)$, for a vertex $\nu_j$, is formulated as:

\small
\begin{eqnarray}
    \Upsilon_{AE}(\nu_j) = (\alpha + \delta) \Upsilon_{S}(\nu_j) + (1 - \alpha) \Upsilon_{VE}(\nu_j).
\end{eqnarray}
\normalsize

Here, $\Upsilon_{VE}(\nu_j)$ is the Volume Gain as explained in Section~\ref{subsec:ve}. $\Upsilon_{S}(\nu_j)$, called the Semantic Gain, is 
defined as the number of unknown voxels in $\Mbb_S^P$ seen by the depth sensor $\Ys_D$ if the robot was at vertex $\nu_j$. 
$\alpha$, called the overlap ratio, is the ratio of the number of detected longitudinals overlapping with the predicted longitudinals to the number of predicted longitudinals. 
To calculate the overlap, each predicted longitudinal is checked to find if its bounding box overlaps with any of the detected longitudinals. 
Two longitudinals are considered to be overlapping if their bounding boxes have more than a user-defined $\kappa$ percent overlap.
The larger the overlap ratio, the higher the contribution of Semantic Gain, leading to the planner focusing more on exploring areas with predicted semantics.
The tunable parameter $\delta$ ($0 \leq \delta \leq 1$) enables the planner to have a small bias for exploring predicted semantic areas in the initial few \ac{ae} planning iterations in each compartment when the overlap ratio is very small or zero. 

When the overlap ratio $\alpha$ goes above a user-defined threshold $\alpha_{thr}$, the \ac{ae} submode exits and the planner switches to the \ac{si} mode to inspect the detected longitudinals. This allows the robot to stop the exploration early on without mapping the entire compartment but after finding the predicted semantics of interest. 

\begin{figure}[h!]
\centering
    \includegraphics[width=0.9\columnwidth]{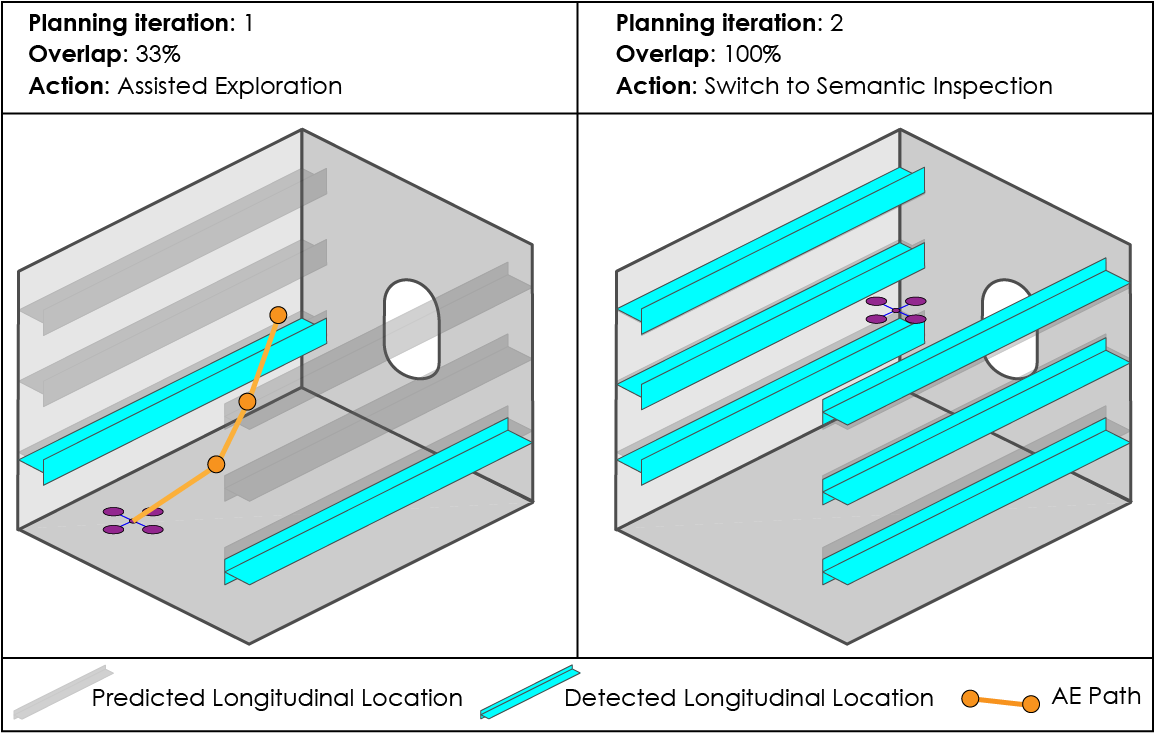}
\centering
\caption{This figure shows illustrations of two key planning steps in the Assisted Exploration (PP-AE) submode. In the planning iteration on the left, the overlap between the detected and predicted semantics is $33\%$ and the planner takes a \ac{ae} step. In the iteration on the right (not necessarily the immediate next iteration) a $100\%$ overlap between the predicted and detected semantics is seen. Hence the planner switches to the \ac{si} mode.}
\label{fig:ae}
\end{figure}

\subsection{Predictive Planning - Opportunistic Inspection (PP-OI)}
\begin{figure}[h!]
\centering
    \includegraphics[width=0.9\columnwidth]{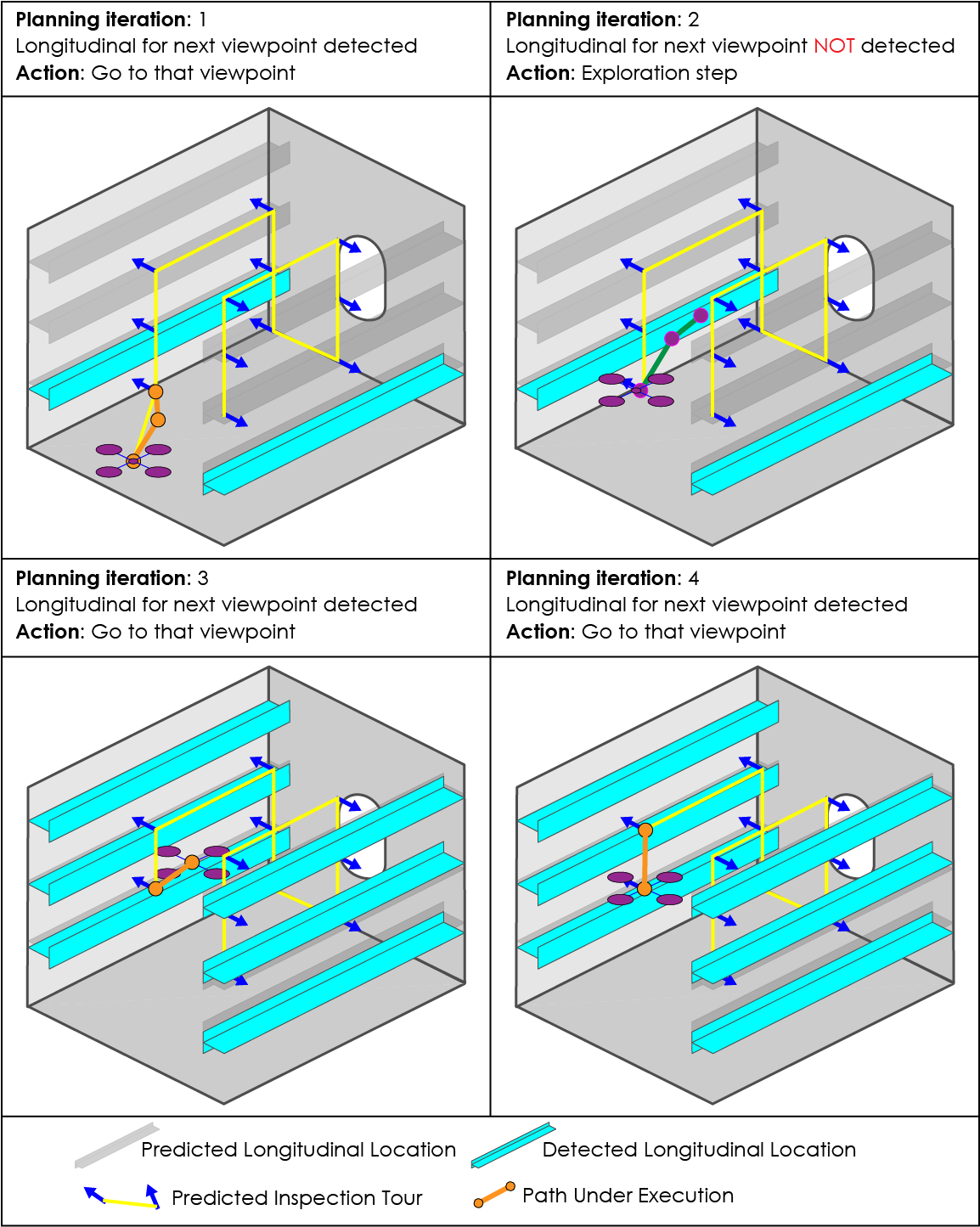}
\centering
\caption{This figure shows four consecutive planning iterations describing the Opportunistic Inspection (PP-OI) submode. As the robot enters through a manhole used for graph prediction, the \ac{oi} submode is triggered. The robot attempts to follow the predicted path as long as the longitudinals viewed by the viewpoints in the path are detected. If not, then the planner takes exploration steps until that longitudinal is detected.}
\label{fig:oi}
\vspace{-3ex} 
\end{figure}

The \ac{oi} is the second submode for the \ac{pp} mode. 
This approach assumes more exact repeatability in the patterns in the \ac{ssg} than \ac{ae}.
If the robot traverses through a manhole used for prediction, the \ac{oi} is triggered. First, the viewpoints $\vbf_i, i=1...n$ ($n$ as defined in Section~\ref{subsec:si}) in the \ac{tsp} tour $\sigma_{tsp}$ from the previous compartment are transformed to match the predicted longitudinals. The transform used to align the $\vartheta^e$ or $\Bar{\vartheta}^e$ of the substructure with the selected $^L\vartheta^e$ or $^L\Bar{\vartheta}^e$ is applied to the viewpoints in $\sigma_{tsp}$.
Next, for the first viewpoint $\vbf_i$ in $\sigma_{tsp}$, if a longitudinal overlapping with the predicted longitudinal $\textit{L}$ that $\vbf_i$ will be inspecting is detected, then a collision-free path from the current robot location to $\vbf_i$ is calculated using using the RRT*~\cite{karaman2010rrtstar} algorithm. The robot traverses this path and the procedure is repeated for the next viewpoint in $\sigma_{tsp}$.
If such a longitudinal is not detected, an exploration planning step guiding the robot towards $\vbf_i$ is taken using a modification of the algorithm used for \ac{ve} mode. The core algorithm remains identical with the only modification being in the function $\Zs$ in Equation~\ref{eqn:lambda_ve}, with each path compared against the straight line path from the current robot location to $\vbf_i$ instead of $\sigma_e$. 
These exploration steps are repeated until a longitudinal overlapping with $\textit{L}$ is detected, at which point a collision-free path to $\vbf_i$ is calculated and the procedure is repeated for the next viewpoint in $\sigma_{tsp}$. If the longitudinal is not found within a predefined number of exploration steps, the planner skips $\vbf_i$ and the procedure is continued for the next viewpoint in $\sigma_{tsp}$. 
Figure~\ref{fig:oi} illustrates indicative planning steps of the algorithm.

%% file: 06_SimulationStudies.tex
\begin{figure*}[h!]
\centering
    \includegraphics[width=0.99\textwidth]{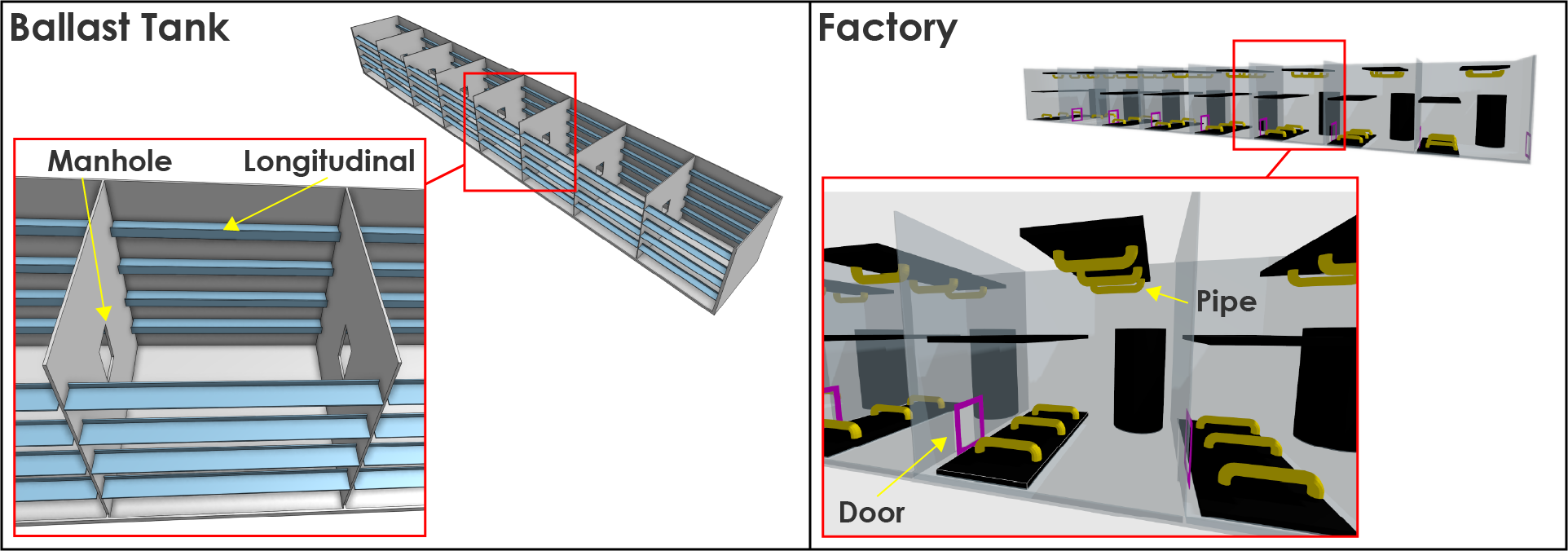}
\centering
\caption{\addition{This figure shows both simulation environments and the semantics within them. The first environment is a ballast tank divided into $8$ compartments connected by manholes of dimensions height $\times$ width $=\SI{2.0}{\meter} \times \SI{1.5}{\meter}$ as the \ac{evs}. Each compartment contains $8$ longitudinals which are the semantics to inspect. The second environment, shown on the right, is a factory containing several pipes that serve as semantics for inspection. The factory consists of $8$ rooms connected by doors of dimensions height $\times$ width $=\SI{2.0}{\meter} \times \SI{1.5}{\meter}$ as the \ac{evs}.}}
\label{fig:sim_envs}
\end{figure*}

\begin{figure}[h!]
\centering
    \includegraphics[width=0.99\columnwidth]{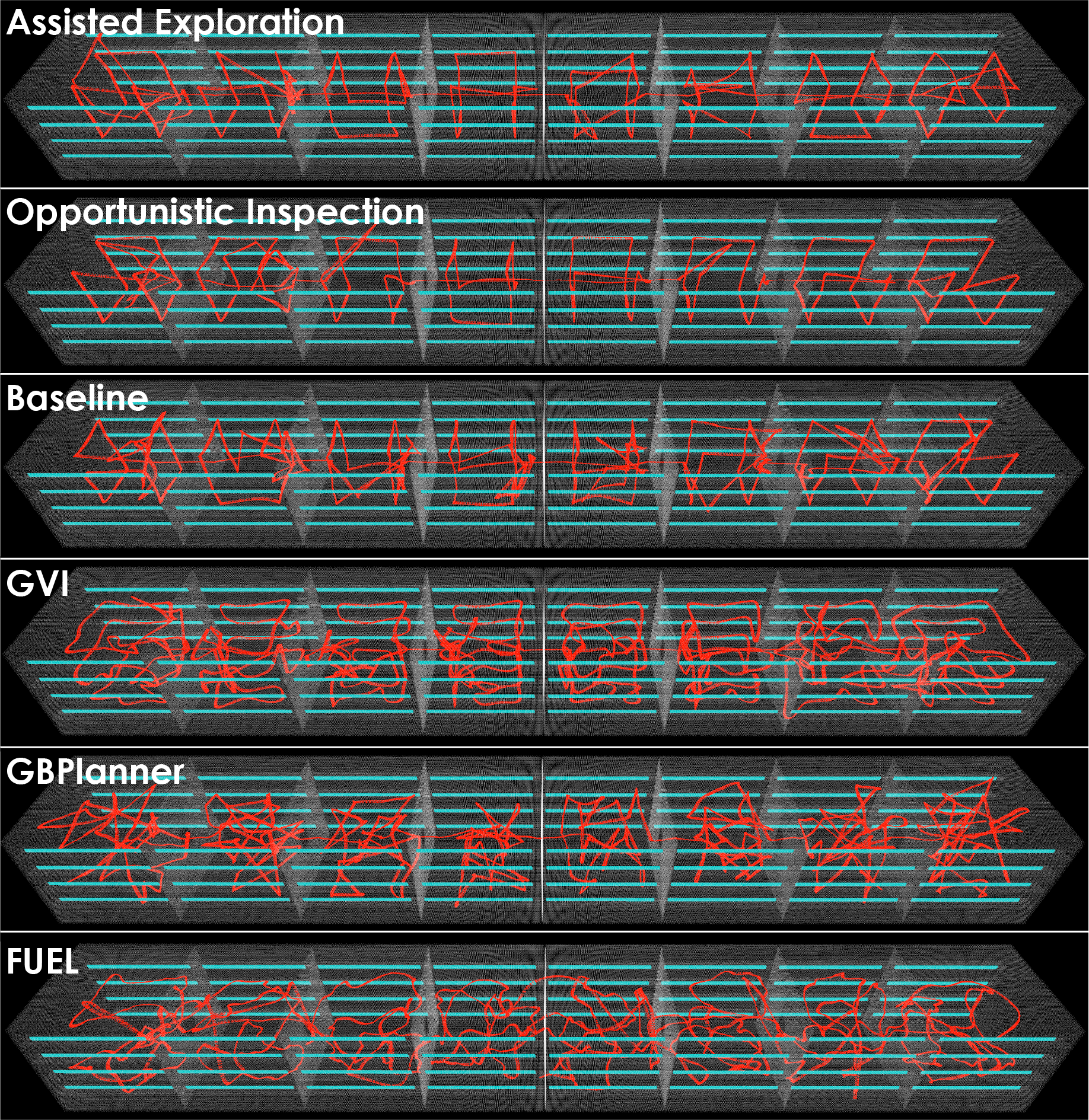}
\centering
\caption{Figure showing the \addition{ballast tank} simulation environment including the longitudinals along with the robot path from one mission of each planner. The proposed predictive planner significantly outperforms the Baseline, GVI, and pure exploration planners.}
\label{fig:sim1_maps}
\vspace{-1ex} 
\end{figure}

\begin{figure*}[h!]
\centering
    \includegraphics[width=0.99\textwidth]{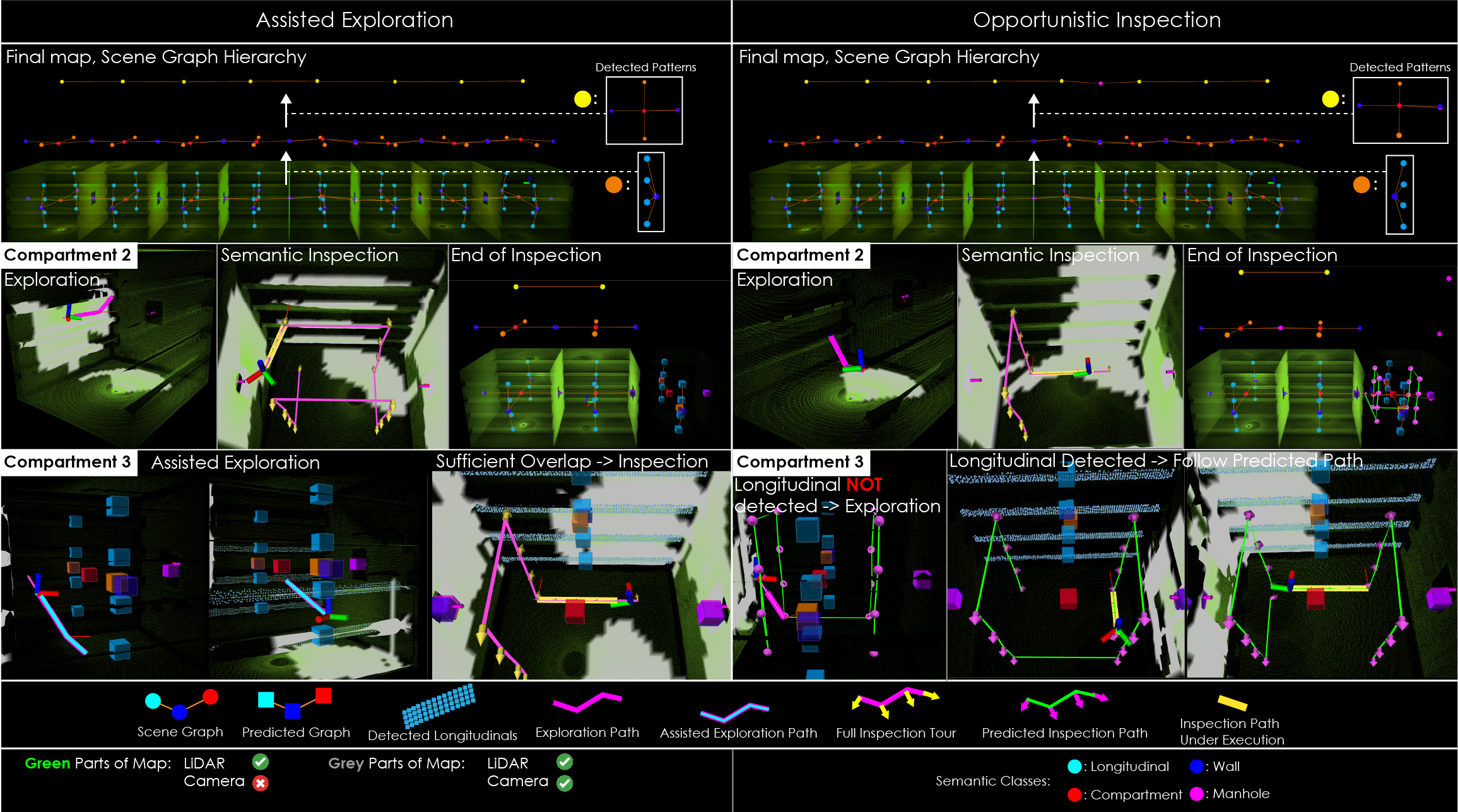}
\centering
\caption{Figure showing indicative steps and final map for the two predictive planning approaches in the \addition{ballast tank} simulation study. The left half of the figure illustrates the \ac{ae} submode showing indicative planning paths, scene graph, detected patterns, and predicted graph. The right half subfigure shows the results for the \ac{oi} submode. The planned paths, detected patterns, predicted graph and paths are presented.\addition{The cyan colored point cloud and \ac{ssg} vertices correspond to the semantics to be inspected.}}
\label{fig:sim1_paths}
\vspace{-1ex} 
\end{figure*}

\begin{figure}[h!]
\centering
    \includegraphics[width=0.99\columnwidth]{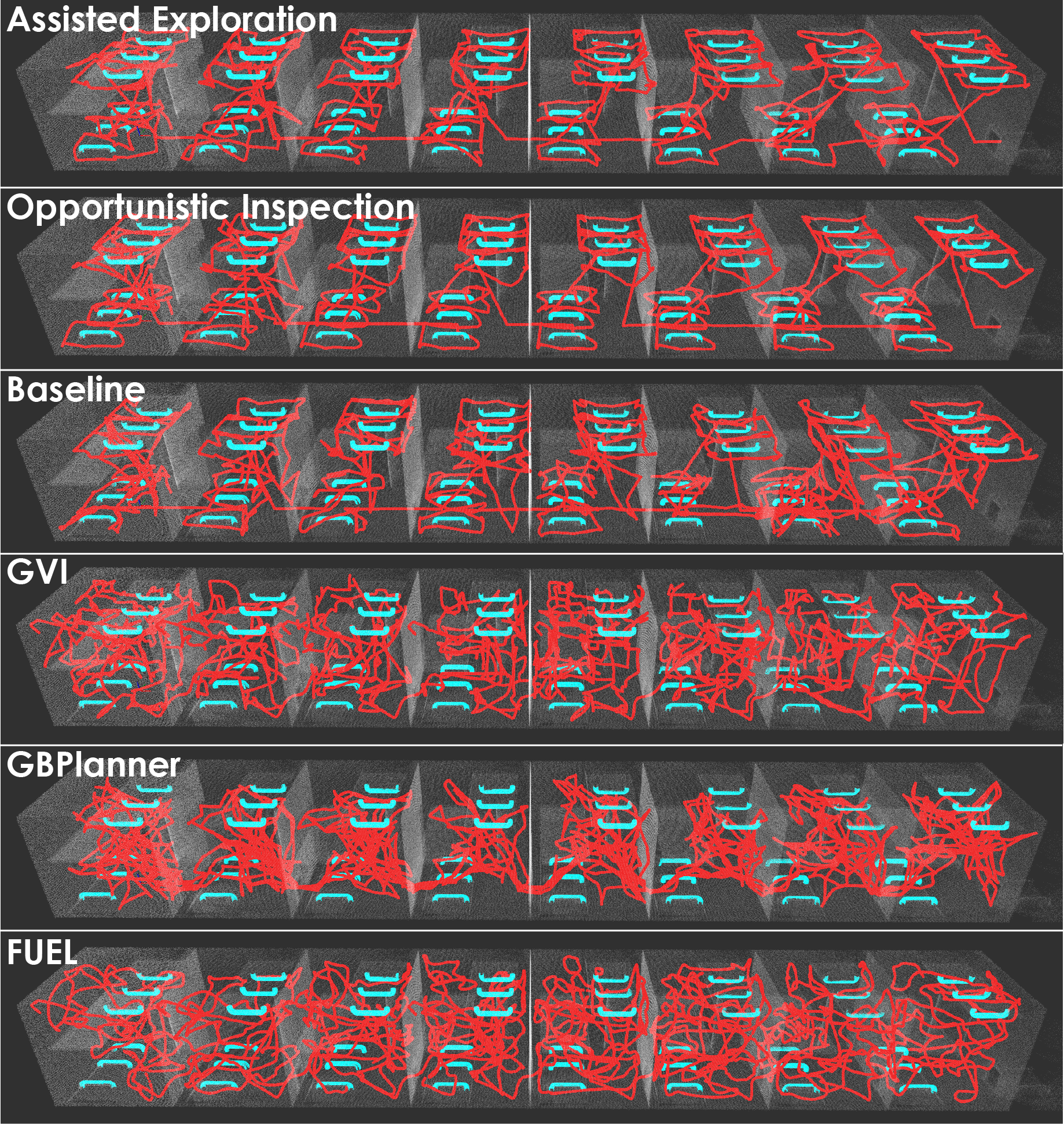}
\centering
\caption{\addition{Figure showing the factory simulation environment including the pipes along with the robot path from one mission of each planner. The proposed predictive planner significantly outperforms the Baseline, GVI, and pure exploration planners.}}
\label{fig:sim2_maps}
\vspace{-1ex} 
\end{figure}

\begin{figure*}[h!]
\centering
    \includegraphics[width=0.99\textwidth]{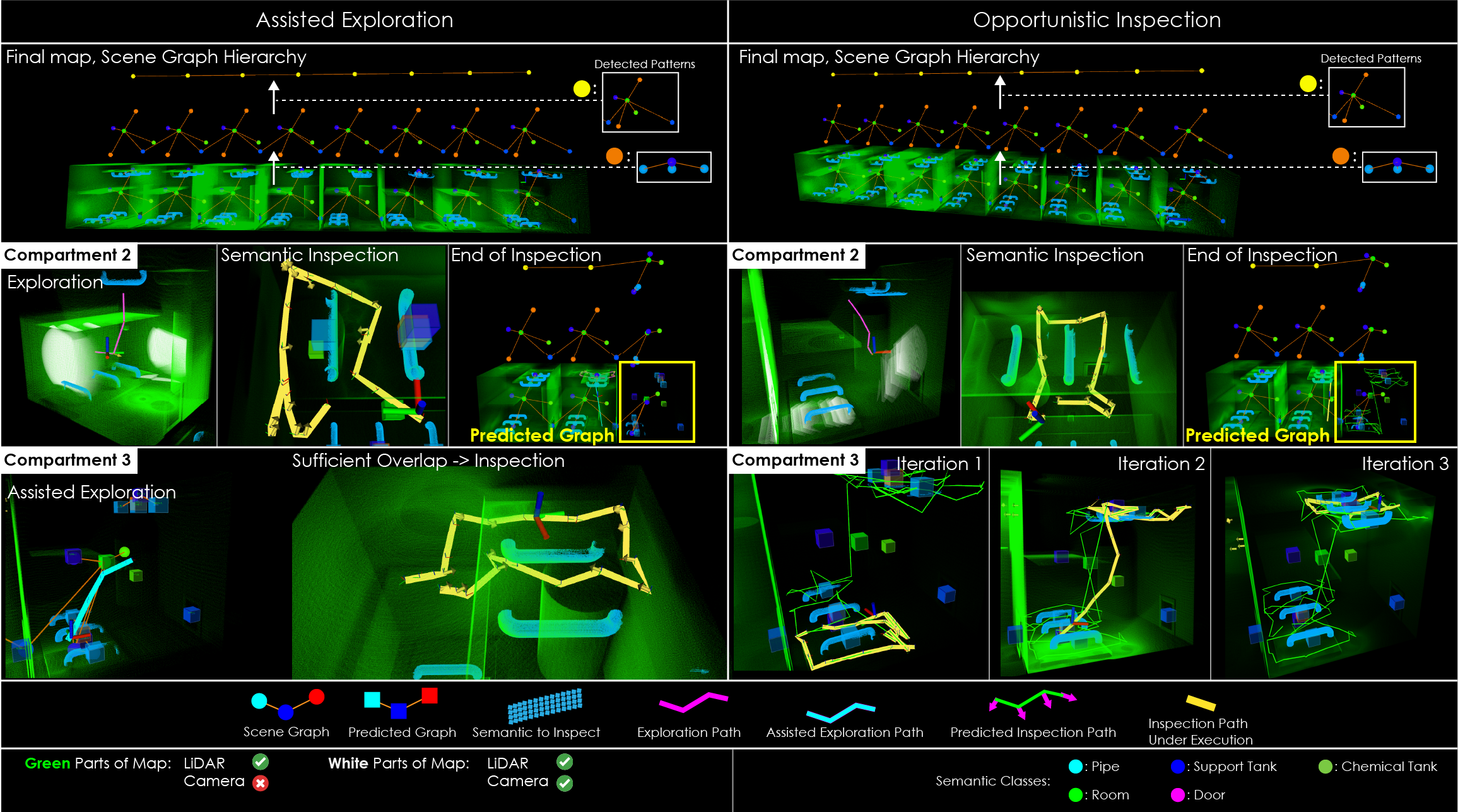}
\centering
\caption{\addition{Figure showing indicative steps and final map for the two predictive planning approaches in the factory simulation study. The left half of the figure illustrates the \ac{ae} submode showing indicative planning paths, scene graph, detected patterns, and predicted graph. The right half subfigure shows the results for the \ac{oi} submode. The planned paths, detected patterns, predicted graph and paths are presented. The cyan colored point cloud and \ac{ssg} vertices correspond to the semantics to be inspected.}}
\label{fig:sim2_paths}
\vspace{-1ex} 
\end{figure*}


\addition{This paper presents two large-scale simulation studies. In the first study, both predictive planning approaches are tested and compared against several exploration and inspection planning methods. We compare the methods in two distinct environments. The first is a large-scale ballast tank consisting of $8$ compartments. Each compartment has dimensions length $\times$ width $\times$ height $=\SI{10}{\meter} \times \SI{10}{\meter} \times \SI{10}{\meter}$ with eight longitudinals to inspect. Longitudinals are metal beams mounted on the walls of ballast tanks to enhance structural integrity. An illustration of these components is provided in Figure~\ref{fig:sim_envs}. The compartments are interconnected by large manholes (one between each pair of compartments) of dimensions height $\times$ width $=\SI{2.0}{\meter} \times \SI{1.5}{\meter}$ which serve as the \ac{evs}. The second environment is an industrial, factory-like setting consisting of several process pipes, which serve as the semantics of interest for inspection. The environment is split into $8$ rooms connected by doors of dimensions height $\times$ width $=\SI{2.0}{\meter} \times \SI{1.5}{\meter}$ that act as the \ac{evs}. Each room contains additional structures, making the environment more cluttered than the ballast tank. The environments and the semantics inside them are shown in Figure~\ref{fig:sim_envs}.
 }

\addition{The second study aims to evaluate the performance of the method in the presence of missing semantics and, therefore, inexact graph patterns. This study is conducted in the ballast tank environment. We procedurally remove one longitudinal from a few compartments in the ballast tank, thus ablating over the number of compartments missing a semantic.}

\addition{The simulations are conducted in the Gazebo simulator with a model of the Resilient Micro Flyer (RMF-Owl)~\cite{rmfowl} collision-tolerant aerial robot. The method described in Section~\ref{subsec:ssg_gen} is used to build the \ac{ssg} for the simulation in the ballast tank, whereas the segmentation camera provided by Gazebo is used for the factory simulation.}

\subsection{Comparative Study}
This study compares the performance of our method against the following exploration and inspection path planning strategies.
\begin{itemize}
    \item \textbf{Baseline:} This approach uses the planning strategy detailed in Section~\ref{subsec:planner_overview} without triggering the \ac{pp} mode and only uses the \ac{ve} and \ac{si} modes. Comparison against this allows us to study the benefit of the predictive planning paradigm against a pure semantics-aware inspection planning approach.
    \item \textbf{GVI:} In this approach, our previously published work on autonomous exploration and General Visual Inspection (GVI) of ballast tanks~\cite{2023expgvi} is utilized. The method performs volumetric exploration along with visual inspection of all mapped surfaces from a given desirable viewing distance. Comparison against this method allows us to study the benefits of using a semantics-aware inspection strategy as opposed to the general inspection of all surfaces.
    \item \textbf{GBPlanner:} We compare against our previous open-sourced graph-based exploration path planner, GBPlanner~\cite{GBPLANNER_JFR_2020,GBPLANNER2COHORT_ICRA_2022}. To ensure that the planner sees all surfaces, the \ac{fov} and the maximum range of the sensor used for exploration are set to be the same as $\Ys_C$. This is to compare against a pure exploration, and thus naturally less efficient, method. 
    \item \textbf{FUEL:} Finally, another exploration path planner, FUEL~\cite{zhou2021fuel}, that employs a fast frontier-based exploration strategy is tested. The sensor parameters are constrained in the same way as GBPlanner for complete coverage. 
\end{itemize}
As the last two methods do not have an explicit way to navigate through narrow manholes, a large manhole was selected in the ballast tank for the two planners to plan through without the need for explicit detection. Such a size of manhole is not representative of real-life ship ballast tanks but without this environment modification, our evaluation runs indicated that neither GBPlanner nor FUEL would be able to complete exploration if a typical manhole size (e.g., $\SI{0.8}{\meter} \times \SI{0.6}{\meter}$) was opted for. However, the proposed predictive planner, Baseline, and GVI use the explicit manhole detection and navigation strategy described in~\cite{manhole2023}.

A total of $5$ missions are conducted for each of the planners \addition{in both simulation environments}. The environment and the robot's path from one mission of each planner are shown in Figures~\ref{fig:sim1_maps}, \addition{ and~\ref{fig:sim2_maps} for the ballast tank and the factory respectively}. Figures~\ref{fig:sim1_paths} and~\ref{fig:sim2_paths} show the final map, the \ac{ssg} at the end of the mission along with the patterns detected and hierarchy built, indicative planning steps, and an indicative graph prediction made during the missions of the predictive planner using the \ac{ae} and \ac{oi} submodes \addition{for the respective simulations}. The parameters used in these missions are listed in Table~\ref{tab:sim_params}. The quantitative results comparing the performance of each method are shown in Table~\ref{tab:sim_stats}. 
\addition{In the factory environment, the viewpoint sampling strategy of the \ac{si} mode is modified to accommodate generic semantics. 
Viewpoints are sampled within a cuboidal bounding box with dimensions $b_1+r_C \times b_2+r_C \times b_3+r_C$, where $\mathbf{b} = [b_1, b_2, b_3]$ is the bounding box of the semantic under inspection.
The remaining procedure to build the graph $\Gbb_{SI}$ and calculate the inspection path remains the same.}
The metrics used for comparison are the following:
\begin{itemize}
    \item \textbf{Inspection time per compartment \addition{ / room} (s):} This is the average time spent by the robot in each compartment \addition{ / room}. For FUEL and GBPlanner, we calculate this time as the total mission time divided by the number of compartments, since these methods do not explicitly split their planning steps per compartment \addition{ / room}. Note that for \ac{ae} and \ac{oi}, we only account for the compartments \addition{ / rooms} in which the respective submode was used, meaning we indicate the effect of exploiting environment prediction. Naturally, the more the compartments \addition{ / rooms} within which prediction can be exploited, the higher the importance of predictive planning. 
    \item \textbf{Semantic Surface Coverage (\%):} This relates to the percentage of semantic surface seen by the camera sensor $\Ys_C$ during the mission at the desired viewing distance $r_C$. \addition{In simulations, we have access to the ground truth mesh of the semantics. For each robot pose in the mission, rays are cast using the camera model $\Ys_C$ from that pose. The mesh faces intersected by the rays are accumulated over the entire mission. The semantic surface coverage is then defined as the percentage of the number of cumulative seen mesh faces to the total number of faces in the mesh.}
    \item \textbf{Path length (m):} The total distance traveled by the robot in a mission.
\end{itemize}

\begin{table}[]
\centering
\begin{tabular}{|l|l|}
\hline
\textbf{Parameter}  & \textbf{Value} \\ \hline
\multicolumn{2}{|l|}{\textbf{SUBDUE Related Parameters}} \\ \hline
$\gamma_b$          & $3$ \\ \hline
$\gamma_l$          & $30$ \\ \hline
$t_{thr}$           & $0.2$ \\ \hline
Vertex/Edge Addition Cost    & $1.0$  \\ \hline
Vertex/Edge Deletion Cost    & $1.0$  \\ \hline
Vertex Label Substitution Cost   & $4.0$  \\ \hline
Edge Label Substitution Cost   & $1.0$  \\ \hline
$\gamma_p$          & $1.0$  \\ \hline
$d_{\min}$          & $\SI{0.5}{\meter}$  \\ \hline
$d_{\max}$          & $\SI{4.0}{\meter}$  \\ \hline
\multicolumn{2}{|l|}{\textbf{Planning Parameters}} \\ \hline
$[F^H_D, F^V_D]$         & $[360,90]^\circ$  \\ \hline
$[F^H_C, F^V_C]$         & $[90,60]^\circ$  \\ \hline
$r_C$               & $\SI{3.0}{\meter}$  \\ \hline
$\mu$               & $0.05$  \\ \hline
$\zeta$             & $0.01$  \\ \hline
$\delta$            & $0.2$  \\ \hline
$\alpha_{thr}$      & $0.9$  \\ \hline
Max speed $v_{\max}$    & $\SI{2.0}{m/s}$  \\ \hline
\end{tabular}
\vspace{2ex}
\caption{Parameters used in the presented Simulation Study}
\label{tab:sim_params}
\end{table}


\begin{table*}[]
\centering
    \begin{tabular}{|l|l|l|l|l|}
    \hline
    \textbf{}          & \textbf{Inspection Time \addition{per}}              & \textbf{Semantic Surface}   & \textbf{Path Length} & \textbf{Computation Time} \\ 
    \textbf{Method}          & \textbf{Compartment \addition{ / room} (s)}              & \textbf{Coverage(\%)}   & \textbf{(m)} & \textbf{(ms)} \\ \hline
    \multicolumn{5}{|l|}{\textbf{\addition{Comparative Study 1 : Ballast Tank}}} \\ \hline
    PP-AE                    & $48.97$                & $98.96$             & $528.47$              & $148.81$ \\ \hline
    PP-OI                    & $40.33$                & $99.93$             & $522.12$              & $31.58$ \\ \hline
    Baseline                 & $65.58$                & $99.87$             & $645.69$              & $277.51$ \\ \hline
    GVI                      & $88.98$                & $99.46$             & $976.59$              & $2046.46$ \\ \hline
    GBPlanner                & $103.06$               & $96.83$             & $1068.35$             & $102.61$ \\ \hline
    FUEL                     & $101.79$               & $98.44$             & $867.99$              & $42.23$ \\ \hline
    \multicolumn{5}{|l|}{\textbf{\addition{Comparative Study 2: Factory}}} \\ \hline
    \addition{PP-AE}          & \addition{$255.48$}      & \addition{$98.26$}    & \addition{$1376.3288$}    & \addition{$125.43$} \\ \hline
    \addition{PP-OI}          & \addition{$214.40$}      & \addition{$98.72$}    & \addition{$1356.0789$}    & \addition{$46.93$} \\ \hline
    \addition{Baseline}       & \addition{$349.35$}      & \addition{$98.65$}    & \addition{$1438.0326$}    & \addition{$165.90$} \\ \hline
    \addition{GVI}            & \addition{$565.89$}      & \addition{$91.25$}    & \addition{$2539.0344$}    & \addition{$2666.89$} \\ \hline
    \addition{GBPlanner}      & \addition{$622.46$}      & \addition{$73.06$}    & \addition{$3187.9088$}    & \addition{$130.29$} \\ \hline
    \addition{FUEL}           & \addition{$629.98$}      & \addition{$79.70$}    & \addition{$2618.4168$}    & \addition{$38.33$} \\ \hline
    \end{tabular}
    \vspace{2ex}
    \caption{Quantitative Results from Comparative Simulation Study. It can be clearly observed that the two predictive planning submodes \ac{ae} and \ac{oi} outperform the other methods due to their ability to exploit pattern prediction. Even if the Baseline approach uses the semantic information for inspection, it significantly underperforms compared to \ac{ae} and \ac{oi}. The average computation times for each method are shown in the last column. For the \ac{ae}, \ac{oi} the computation time is reported only for the planning steps when the planner was operating in the respective submodes.}
    \label{tab:sim_stats}
\end{table*}

\addition{As shown in Table~\ref{tab:sim_stats}, the proposed predictive planning submodes outperform the rest in terms of inspection time per compartment/room while achieving comparable (often higher) semantic surface coverage. Specifically, the \ac{ae} and \ac{oi} submodes outperform the Baseline by $25\%$ and $38\%$, GVI by $45\%$ and $54\%$, and the exploration planners by $52\%$ and $60\%$ in the ballast tank environment. In the factory environment, they outperform the Baseline by $27\%$ and $38\%$, GVI by $55\%$ and $62\%$, and the exploration planners by $59\%$ and $66\%$.}
This highlights the benefit of using the predictive planning approach. \addition{It is noted that the semantics in the factory environment require the robot to access narrow parts between the walls and the pipes. Since the exploration planners only aim to map the given volume, these parts of the semantics are not seen properly. Hence, the two exploration planners have lower semantic surface coverage in the factory environment as compared to the ballast tank.}

\begin{figure*}[h!]
\centering
    \includegraphics[width=0.99\textwidth]{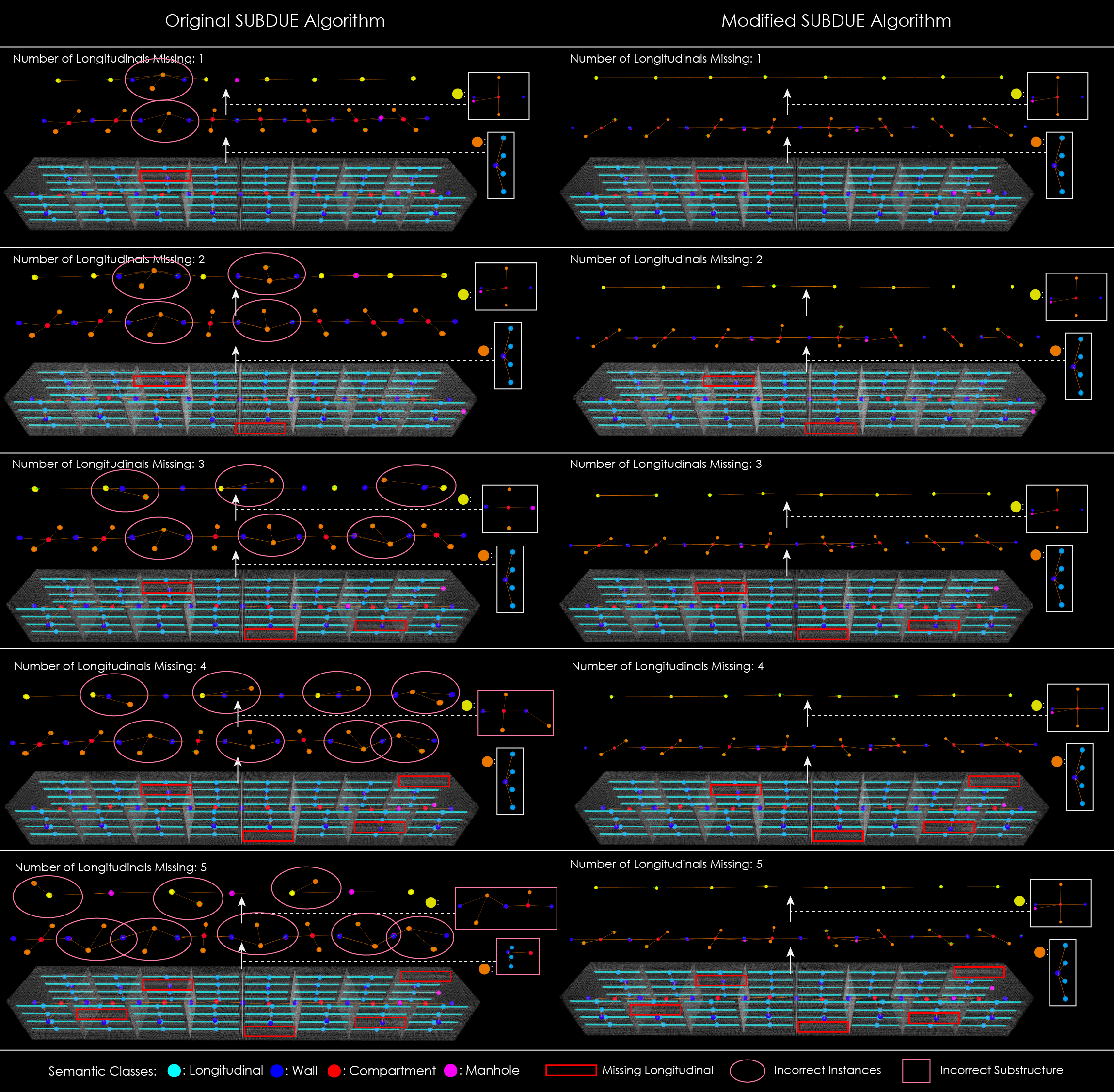}
\centering
\caption{Figure showing the $5$ versions of the simulation environment with inexact patterns along with the detected patterns and hierarchies of compressed graphs. The left side figures show the original SUBDUE algorithm tested in each version of the ballast tank. It can be clearly seen that the modified algorithm was able to identify the correct patterns but the original algorithm did not.}
\label{fig:sim1_defects}
\end{figure*}

\begin{table}[]
    \centering
    \begin{tabular}{|l|l|l|l|}
    \hline
    \textbf{}          & \textbf{Inspection Time /}              & \textbf{Semantic Surface}   & \textbf{Path Length} \\ 
    \textbf{Method}          & \textbf{Compartment(s)}              & \textbf{Coverage(\%)}   & \textbf{(m)} \\ \hline
    \multicolumn{4}{|l|}{Number of compartments missing a longitudinal: $1$ } \\ \hline
    PP-AE                       & $47.5097$                & $99.8947$      & $514.2059$ \\ \hline
    PP-OI                       & $39.4998$       & $99.9659$      & $531.9714$ \\ \hline
    Baseline                 & $64.7500$                & $99.576$       & $618.2521$ \\ \hline
    \multicolumn{4}{|l|}{Number of compartments missing a longitudinal: $2$ } \\ \hline
    PP-AE                       & $45.5292$                & $98.5946$      & $515.5791$ \\ \hline
    PP-OI                       & $42.5706$                & $98.8924$      & $540.3622$ \\ \hline
    Baseline                 & $62.3510$                & $99.5951$      & $613.6175$ \\ \hline
    \multicolumn{4}{|l|}{Number of compartments missing a longitudinal: $3$ } \\ \hline
    PP-AE                       & $47.1471$                & $98.6304$      & $518.9256$ \\ \hline
    PP-OI                       & $43.6990$                & $99.0683$      & $538.8774$ \\ \hline
    Baseline                 & $62.4917$                & $98.6113$      & $635.8960$ \\ \hline
    \multicolumn{4}{|l|}{Number of compartments missing a longitudinal: $4$ } \\ \hline
    PP-AE                       & $46.0173$                & $99.7938$      & $514.2457$ \\ \hline
    PP-OI                       & $45.5803$                & $98.9333$      & $552.1726$ \\ \hline
    Baseline                 & $63.9893$                & $98.9602$      & $610.8366$ \\ \hline
    \multicolumn{4}{|l|}{Number of compartments missing a longitudinal: $5$ } \\ \hline
    PP-AE                       & $44.6311$                & $99.8967$      & $510.0772$ \\ \hline
    PP-OI                       & $46.515$                 & $99.3527$      & $571.7858$ \\ \hline
    Baseline                 & $62.5788$                & $98.59$        & $627.1848$ \\ \hline
    \end{tabular}
    \vspace{2ex}
    \caption{Ablation Study for Missing Semantics. We evaluate the two predictive planning submodes and the Baseline in ballast tanks with varying number of compartments missing one longitudinal each. The results show that the proposed planner is able to handle the imperfect patterns successfully showing significant improvement over the Baseline.}
    \label{tab:sim_missing_sem}
\end{table}

\subsection{Ablation Study for Missing Semantics}\label{subsec:ablation}

This study aims to evaluate the performance of the proposed method under the circumstances of imperfect patterns. 
We create $5$ versions of the ballast tank shown in Figure~\ref{fig:sim1_maps} where in the $k^{th}$ version, one longitudinal is removed from $k$ compartments.
The purpose of this study is twofold. First, to evaluate the ability of the modified SUBDUE algorithm in the presence of imperfect patterns and compare its performance against the original SUBDUE algorithm. Second, to demonstrate the ability of both Predictive Planning submodes, \ac{ae} and \ac{oi}, to handle imperfect patterns. To this end, the \ac{ae} and \ac{oi} submodes, as well as the Baseline are tested in all versions of the ballast tank. The original SUBDUE algorithm is tested on the \ac{ssg} of the full tank of each version, and the results are compared with the modified version. 
Note that the exploration planners and the GVI are not tested as they do not explicitly account for semantics, therefor their performance will not be affected in any meaningful way by the missing semantics.


Figure~\ref{fig:sim1_defects} presents the patterns detected and the hierarchy built for each version of the ballast tank and the quantitative results of this study are shown in Table~\ref{tab:sim_missing_sem}. The left side figures of Figure~\ref{fig:sim1_defects} show the results of the original SUBDUE algorithm and those on the right side show the modified one. It is evident that the modified SUBDUE algorithm was able to identify the correct patterns despite the missing semantics in all tests, while the original algorithm failed to do so. The original algorithm, at instances, incorrectly grouped the `Compartment' vertex in the instances of the pattern with missing longitudinals. 
As the modified algorithm accounts for the pose of the vertices, the incorrect grouping of the `Compartment' vertex is penalized. Furthermore, the penalty for incorrectly mapped high-degree vertices (such as the `Compartment') leads to a higher graph-matching cost for this incorrect mapping.

Despite the missing semantics, both Predictive Planning submodes were able to provide comparable semantic coverage as the missions with perfect patterns. This shows the ability of the planner to handle imperfect patterns. At the same time, they outperformed the Baseline by more than $25\%$ in terms of inspection time, a result that is comparable to the case of perfect patterns.
For the \ac{oi} submode, the average inspection time per compartment increases slightly as the number of compartments with missing longitudinals increases. This can be attributed to the fact that the robot needs to perform additional exploration steps in the compartments with missing longitudinal to search for longitudinals at the predicted locations.
A slight decrease can be observed in the inspection time per compartment for the \ac{ae} and the Baseline solutions as the robot needs to inspect fewer semantics. 
However, in all cases, the changes are small and the Predictive Planning submodes show a significant advantage over the Baseline.


    

%% file: 07_FieldExperiments.tex

To demonstrate the field readiness of the proposed SPP paradigm, we present extensive field deployments in real ballast tanks inside two oil tanker ships. The deployments were conducted in the side section of the ballast tanks (the point cloud map is shown in Figure~\ref{fig:intro}) using a variant of the Resilient Micro Flyer (RMF)-Owl~\cite{rmfowl} collision-tolerant aerial robot. In both deployments, three missions were conducted testing the \ac{ae} and \ac{oi} submodes along with the Baseline approach described in Section~\ref{sec:simulation}. The performance of the methods is compared using the metrics detailed in Section~\ref{sec:simulation}. \addition{Since we do not have access to ground truth meshes for the semantic surface coverage calculation, we create the meshes using the segmented point cloud. As the longitudinals can be approximated by rectangles, simplified rectangle-shaped meshes were created using the dimensions given by the segmented point cloud.} The parameters used by the predictive planner in both field deployments are listed in Table~\ref{tab:exp_params}. The dataset from both deployments will be open-sourced soon extending our previously released ballast tank dataset available at \url{https://github.com/ntnu-arl/ballast_water_tank_dataset}.

\begin{figure*}[h!]
\centering
    \includegraphics[width=0.99\textwidth]{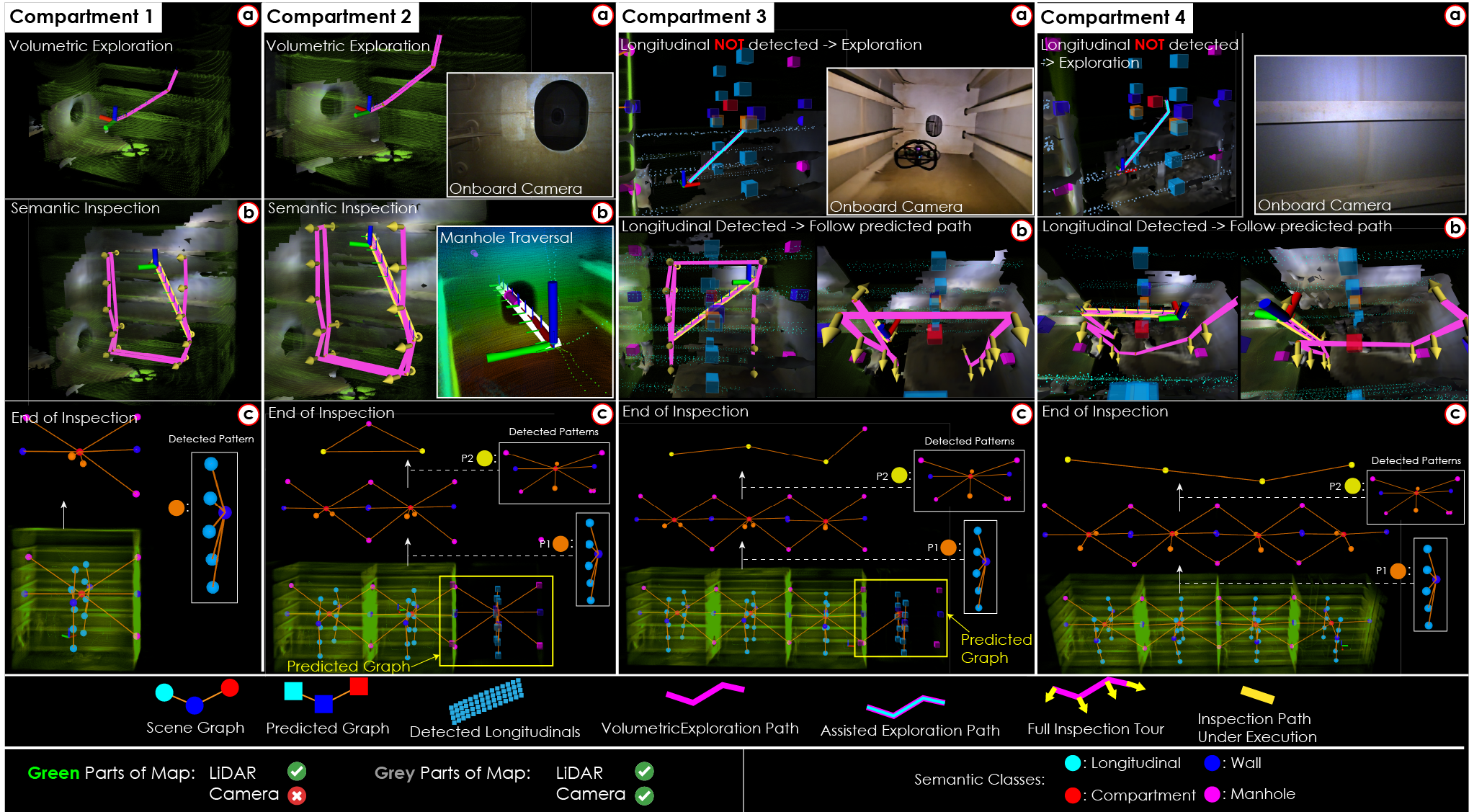}
\centering
\caption{Field Deployment 1: Mission 1. The figure shows the scene graph built, paths planned, and patterns found in each compartment. The robot performed \ac{ve} and \ac{si} steps in the first two compartments after which a successful graph prediction was made. The robot operated in \ac{ae} submode in compartments $3$ and $4$. The planner was able to find correct patterns and utilize the prediction for efficient exploration and inspection.}
\label{fig:as_assist}
\end{figure*}

\begin{figure*}[h!]
\centering
    \includegraphics[width=0.99\textwidth]{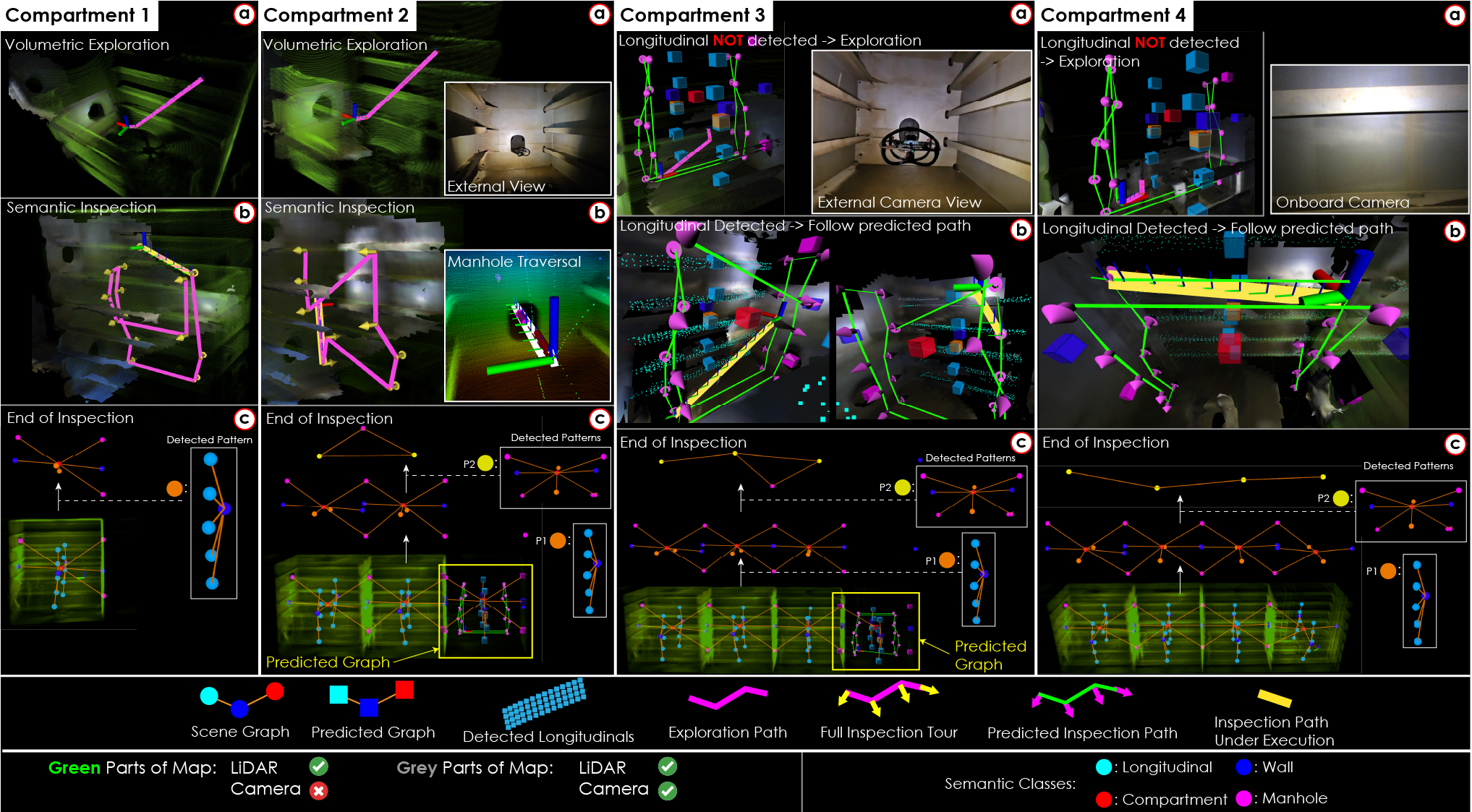}
\centering
\caption{Field Deployment 1: Mission 2. After performing exploration and inspection in the first two compartments without any graph prediction, a successful prediction was made at the end of the inspection of the second compartment. The planner then operated in the \ac{oi} submode in the $3^{\textrm{rd}}$ and $4^{\textrm{th}}$ compartments. The predicted graph and paths, along with indicative planning steps, are shown in the figure.}
\label{fig:as_opp}
\end{figure*}

\begin{figure}[h!]
\centering
    \includegraphics[width=0.9\columnwidth]{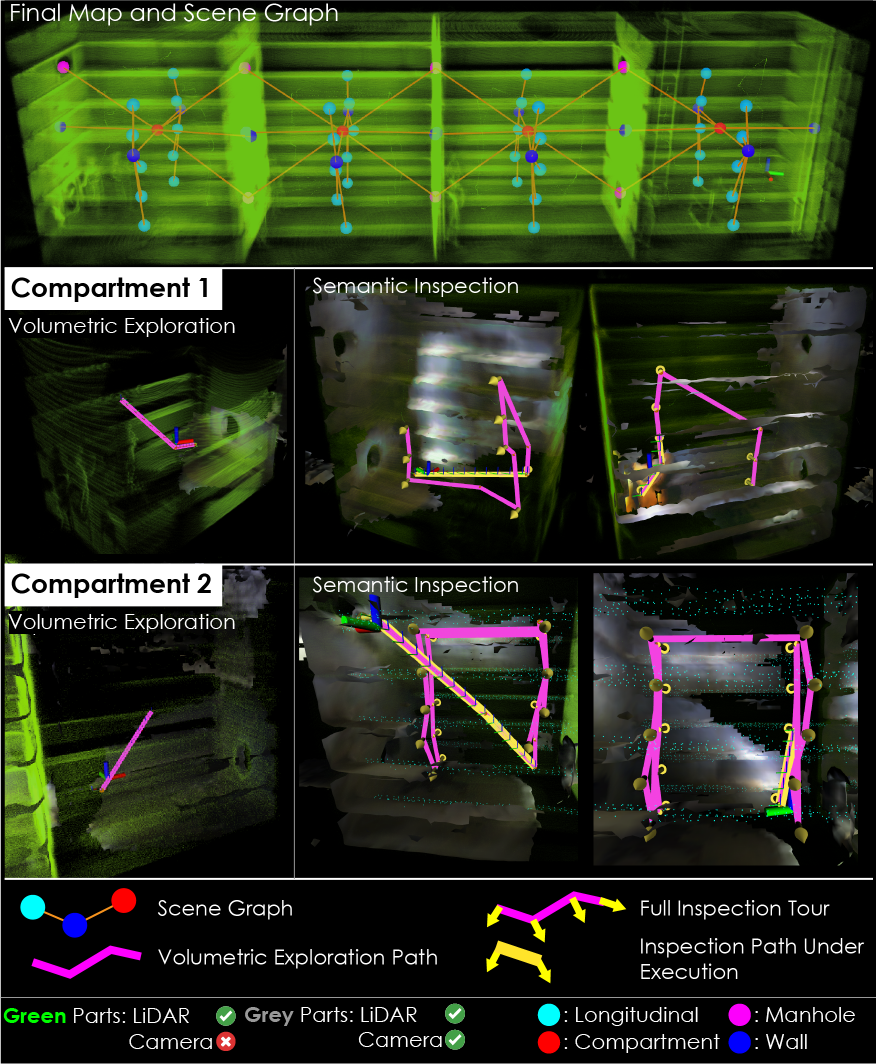}
\centering
\caption{Field Deployment 1: Mission 3. The figure shows indicative planning steps, final map, and scene graph built during this mission. The Baseline approach was used in this mission hence no pattern detection or graph prediction was made.}
\label{fig:as_van}
\end{figure}

\subsection{Collision-tolerant Aerial Robot System Description}
All experiments are conducted using an upgraded version of the collision-tolerant aerial robot RMF-Owl presented in~\cite{rmfowl}. With its collision-tolerant frame, the robot is designed for autonomous navigation in confined environments. The robot presents a small form factor of length $\times$ width $\times$ height $= \SI{0.38}{\meter} \times \SI{0.38}{\meter} \times \SI{0.24}{\meter}$ weighing $\SI{1.45}{\kilogram}$. It utilizes a $4$s $5000$mAh battery providing an endurance of $10$ min.
The robot carries a Khadas VIM4 Single Board Computer (SBC) having $\times 4$ 2.2Ghz Cortex-A73 cores, along with $\times4$ 2.0Ghz Cortex-A53 cores implementing an Amlogic A311D2 big-little architecture. The entire autonomy software stack including SLAM, position control and predictive planning runs onboard this computer. The SBC interfaces with a Pixracer autopilot for low-level control.
The sensing suite of the robot contains an Ouster OS0-64 LiDAR used as $\Ys_D$ (\ac{fov}: $[360,90]^{\circ}$, max range: $\SI{100}{\meter}$), a Blackfly S color camera used as $\Ys_C$ (Resolution: $720 \times 540$, \ac{fov}: $[85,64]^\circ$), and a VectorNav VN100 IMU. This computing and sensing suite allows the robot to maintain a lightweight and small form factor while providing autonomous exploration and inspection capabilities in confined environments. 
The robot uses CompSLAM~\cite{khattak2020complementary}, a multi-modal SLAM framework, for accurate odometry and mapping. Additionally, a Model Predictive Controller~\cite{mpc_rosbookchapter} is used for tracking trajectories given by the proposed Predictive Planner.

\begin{table}[]
\centering
\begin{tabular}{|l|l|}
\hline
\textbf{Parameter}  & \textbf{Value} \\ \hline
\multicolumn{2}{|l|}{\textbf{SUBDUE Related Parameters}} \\ \hline
$\gamma_b$          & $3$ \\ \hline
$\gamma_l$          & $30$ \\ \hline
$t_{thr}$           & $0.2$ \\ \hline
Vertex/Edge Addition Cost    & $1.0$  \\ \hline
Vertex/Edge Deletion Cost    & $1.0$  \\ \hline
Vertex Label Substitution Cost   & $4.0$  \\ \hline
Edge Label Substitution Cost   & $1.0$  \\ \hline
$\gamma_p$          & $1.0$  \\ \hline
$d_{\min}$          & $\SI{0.5}{\meter}$  \\ \hline
$d_{\max}$          & $\SI{4.0}{\meter}$  \\ \hline
\multicolumn{2}{|l|}{\textbf{Planning Parameters}} \\ \hline
$[F^H_D, F^V_D]$         & $[360,90]^\circ$  \\ \hline
$[F^H_C, F^V_C]$          & $[85,64]^\circ$  \\ \hline
$r_C$               & $\SI{1.0}{\meter}$  \\ \hline
$\mu$               & $0.05$  \\ \hline
$\zeta$             & $0.01$  \\ \hline
$\delta$            & $0.2$  \\ \hline
$\alpha_{thr}$      & $0.9$  \\ \hline
Max speed $v_{\max}$    & $\SI{1.5}{m/s}$  \\ \hline
\end{tabular}
\vspace{2ex}
\caption{Parameters used in Field Deployments}
\label{tab:exp_params}
\end{table}

\subsection{Field Deployment 1}

The first evaluation was conducted in an oil tanker, and the robot was deployed inside the side section of one of its ballast tanks. A point cloud map of the tank can be seen in Figure~\ref{fig:as_assist}. Each compartment of the selected section had the dimensions length $\times$ width $\times$ height $=\SI{4.8}{\meter} \times \SI{2.9}{\meter} \times \SI{5.5}{\meter}$. The compartments were each connected by two manholes of dimensions height $\times$ width $=~\SI{0.8}{\meter} \times \SI{0.6}{\meter}$ and $\SI{0.4}{\meter} \times \SI{0.6}{\meter}$ located at a height of $\SI{0.8}{\meter}$ and $\SI{4.3}{\meter}$ respectively from the ground. Each compartment had a total of 12 longitudinals, however, for safety reasons the maximum operating height of the robot was limited to $\SI{4.5}{\meter}$ reducing the scope of inspection to 10 longitudinals per compartment.
Three missions were conducted testing the \ac{oi} and \ac{ae} submodes, as well as the Baseline. In each mission, the robot was tasked to explore and inspect the longitudinals in four compartments. It is highlighted that the Predictive Planner did not have access to any information about the ballast tank --including the compartment dimensions, number of longitudinals, etc-- other than the number of compartments to inspect. Subsequently, we describe each mission in detail. A video detailing Field Deployment $1$ is available at \textbf{\url{https://youtu.be/XhT3nVC9d-I}}.

\subsubsection{\textbf{Mission 1: Assisted Exploration}}
In this mission, the robot started in the first compartment in the \ac{ve} mode and then switched to the \ac{si} mode upon exploring the entire compartment (the definition of compartment here is as per Section~\ref{subsec:ssg_gen}).
Once the compartment was inspected, the pattern detection and prediction step was triggered. The detected pattern consisted of five longitudinal vertices connected to a wall vertex as shown in Figure~\ref{fig:as_assist} (Compartment $1$.c). As the pattern had no $\vartheta^e$, the robot navigated through the nearest untraversed manhole and switched to the \ac{ve} mode for exploration followed by \ac{si} for longitudinal inspection. Upon completion, the pattern detection step was triggered and a two-level hierarchical structure was built as shown in Figure~\ref{fig:as_assist} (Compartment $2$.c). The pattern found in the first level (referred to as `p1' and denoted by the orange circle) consists of five longitudinal vertices connected to a wall vertex. The pattern in the second level (referred to as `p2' and denoted by the yellow circle) consists of a compartment vertex connected to two wall vertices, two vertices representing the substructure `p1', and four manholes. As the substructure `p2' contained $\vartheta^e$ (the manholes), the graph prediction step was triggered and the prediction made by the planner is shown in Figure~\ref{fig:as_assist} (Compartment $2$.c). 

Given this prediction, the robot traversed through the manhole used for the prediction, entered the next compartment, and switched to the \ac{ae} submode. The planner performed \ac{ae} steps until the overlap ratio $\alpha \geq \alpha_{thr}$. At which point the planner switched to the \ac{si} mode to inspect the longitudinals. In this compartment, only $1$ \ac{ae} step was conducted. After finishing the inspection, the pattern detection and prediction step was triggered. A hierarchy of two levels was built with similar substructures found at the end of the inspection of Compartment $2$. The robot traversed through the manhole used for graph prediction, switched to the \ac{ae} submode, and continued the mission. Indicative planning steps in compartments $3$ and $4$ are shown in Figure~\ref{fig:as_assist} (Compartment $3$) and (Compartment $4$) respectively. The total mission time in this case was $\SI{279}{\second}$.

\subsubsection{\textbf{Mission 2: Opportunistic Inspection}}
Similar to Mission 1, the robot started in the first compartment in the \ac{ve} mode. The behavior of the planner in the first two compartments was similar to that in Mission 1. At the end of the inspection of the second compartment, the pattern detection step was triggered and a two-level hierarchy shown in Figure~\ref{fig:as_opp} (Compartment $2$.c) was built. 
As the pattern `p2' contains $\vartheta^e$, a graph prediction was made. The predicted graph extension and the predicted path can be seen in Figure~\ref{fig:as_opp} (Compartment $2$.c). 

The robot navigated through the manhole used for prediction, entered the next compartment, and switched to the \ac{oi} submode. In the first planning iteration, no longitudinal was detected, so the planner performed an exploration step as shown in Figure~\ref{fig:as_opp} (Compartment $3$.a). At this point, the longitudinal corresponding to the first viewpoint in the predicted path was detected and the robot planed a path towards that viewpoint. The robot continued the mission visiting each viewpoint in the predicted path. Two indicative steps are shown in Figure~\ref{fig:as_opp} (Compartment $3$.b). Upon inspecting Compartment $3$, the pattern detection and prediction step was triggered and patterns similar to those at the end of Compartment $2$ were detected. The robot traversed through the relevant manhole and continued the mission. Indicative planning steps in Compartment $4$ are shown in Figure~\ref{fig:as_opp} (Compartment $4$). The total mission time in this mission was $\SI{258}{\second}$.

\subsubsection{\textbf{Mission 3: Baseline}}
The Baseline approach, as described in Section~\ref{sec:simulation}, was used for comparison. The planner does not use the pattern detection and graph prediction steps in this mission. 
In each compartment, the robot started in the \ac{ve} mode, switched to \ac{si} mode when fully explored, and traversed through the closest manhole upon completing the inspection.
The final map and the scene graph built during the mission are shown in Figure~\ref{fig:as_van} along with indicative \ac{ve} and \ac{si} paths. The total mission time was $\SI{318}{\second}$.

The performance of all methods is evaluated using the metrics presented in Section~\ref{sec:simulation}. 
The quantitative results are shown in Table~\ref{tab:exp_stats} (Field Deployment 1). It is clear from the table that the proposed predictive planning submodes outperform the Baseline in terms of inspection time per compartment and path length while maintaining comparable semantic surface coverage. Note that the inspection time for the \ac{ae} and \ac{oi} submodes considers only the last two compartments in which the two submodes were used. This demonstrates that the predictive planning approach is more effective even in real-world scenarios.

\begin{figure*}[h!]
\centering
    \includegraphics[width=0.99\textwidth]{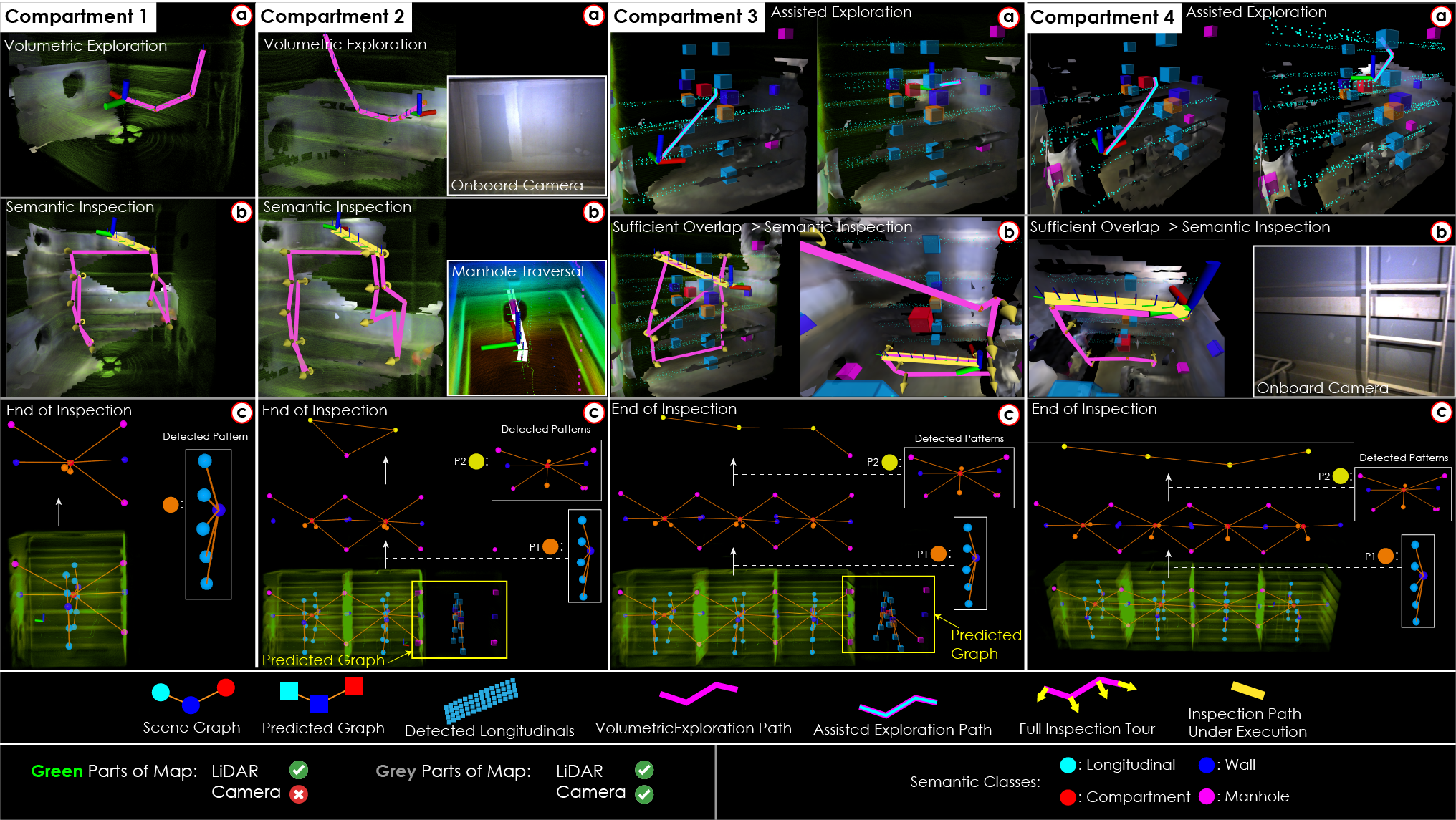}
\centering
\caption{Field Deployment 2: Mission 1. The robot started in the first compartment in the \ac{ve} mode. As no graph prediction was made in the first two compartments, the \ac{pp} mode was not used. After successful prediction at the end of the second compartment, the \ac{ae} submode was used in Compartment $3$. The same behavior was observed in Compartment $4$. The \ac{ae} paths, scene graphs, and maps can be seen in the figure.}
\label{fig:rs_assist}
\end{figure*}

\begin{figure*}[h!]
\centering
    \includegraphics[width=0.99\textwidth]{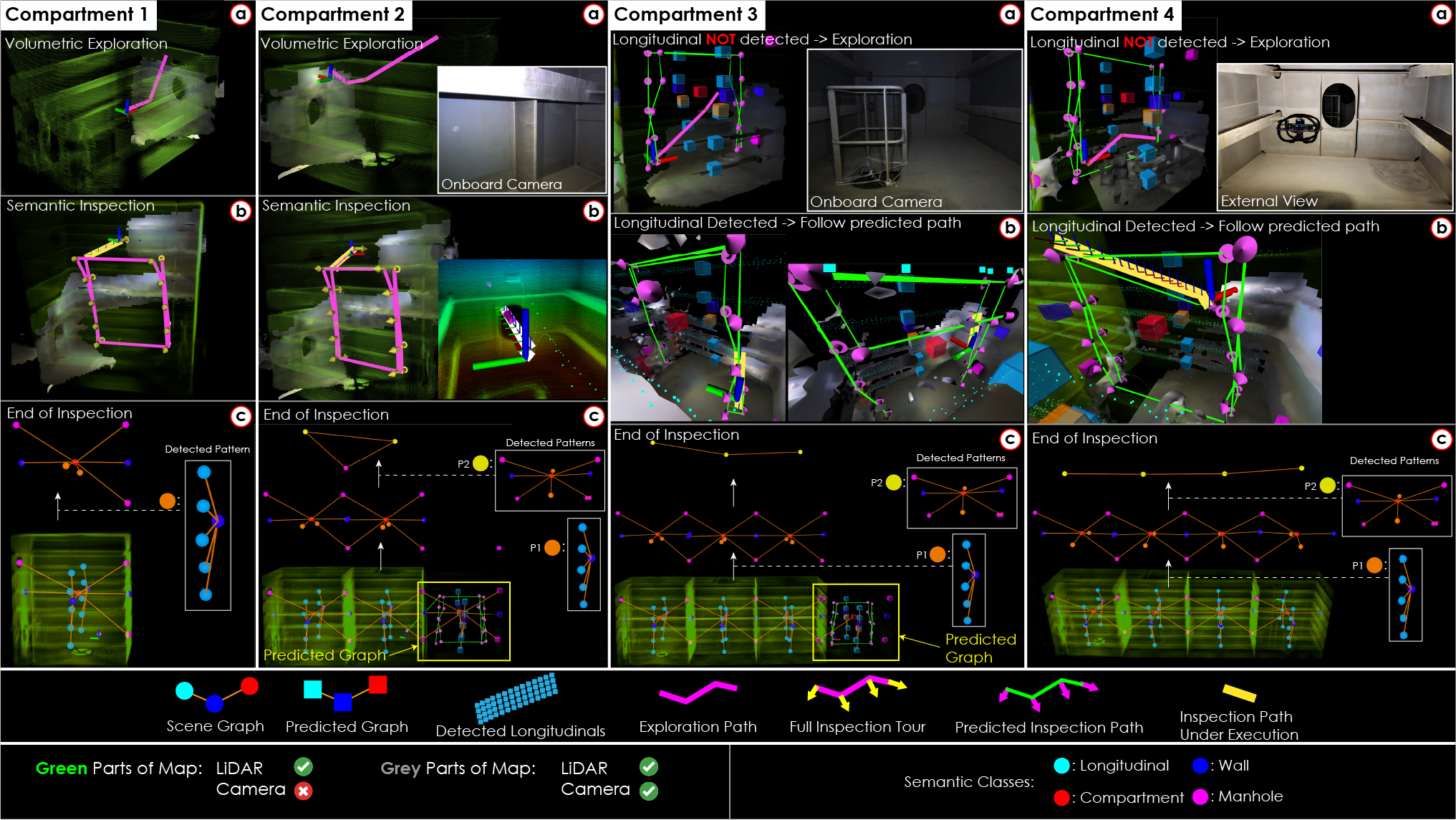}
\centering
\caption{Field Deployment 2: Mission 2. The figure shows indicative planning steps from each compartment. During the four-compartment mission, no graph prediction was made in the first two compartments and only \ac{ve} and \ac{si} modes were used. At the end of the inspection of the second compartment, a successful graph prediction was made and the planner switched to \ac{oi} submode upon entering the third compartment. Similarly the planner operated in the \ac{oi} submode in the fourth compartment as well.}
\label{fig:rs_opp}
\end{figure*}

\begin{figure}[h!]
\centering
    \includegraphics[width=0.9\columnwidth]{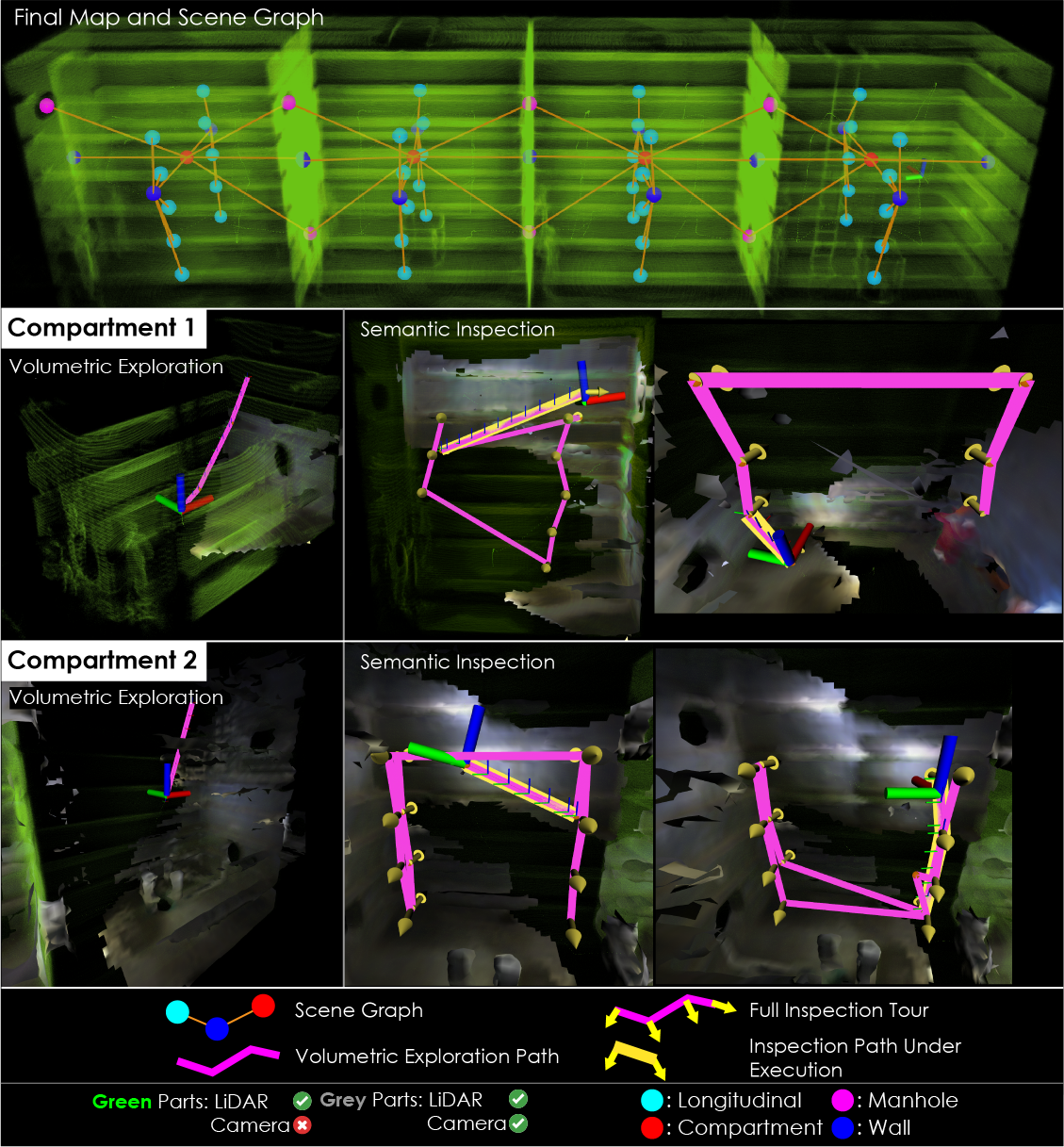}
\centering
\caption{Field Deployment 2: Mission3. The figure shows indicative planning steps for the Baseline solution. The planned paths, scene graph, and final map are shown in the figure. It is highlighted that the Baseline does not do pattern detection and graph prediction thus using only the \ac{ve} and \ac{oi} modes.}
\label{fig:rs_van}
\end{figure}

\subsection{Field Deployment 2}
The second field deployment was conducted in a ballast tank of another oil tanker. The robot was deployed in the side section of the tank. This tank presents a layout similar to that in the first deployment but with a different internal structure. This deployment further shows the repeatability of the method in real-world scenarios. Each level of the ballast tank had compartments of dimensions length $\times$ width $\times$ height $=\SI{4.8}{\meter} \times \SI{2.9}{\meter} \times \SI{5.5}{\meter}$. The compartments were connected by manholes of dimensions height $\times$ width $=~\SI{0.8}{\meter} \times \SI{0.6}{\meter}$ and $\SI{0.4}{\meter} \times \SI{0.6}{\meter}$. 
Similar to Field Deployment $1$, we tested the \ac{ae}, \ac{oi}, and the Baseline solutions. The robot was tasked to inspect $4$ compartments in each mission. The maximum flying height was limited to $\SI{4.5}{\meter}$ allowing the inspection of $10$ longitudinals in each compartment. A video detailing Field Deployment $2$ is available at \textbf{\url{https://youtu.be/7Efu8N33XqU}}.

\subsubsection{\textbf{Mission 1: Assisted Exploration}}
The planner started in the first compartment in the \ac{ve} mode, switched to \ac{si} mode when fully explored and upon completion, triggered the pattern detection. The detected pattern consisted of one wall vertex connected to five longitudinal vertices as shown in Figure~\ref{fig:rs_assist} (Compartment $1$.c). Since the substructure did not contain $\vartheta^e$, the robot traversed to the next compartment through the closest manhole and switched to the \ac{ve} mode. At the end of the inspection, the pattern detection step was triggered and a two-level hierarchical graph structure was built as shown in Figure~\ref{fig:rs_assist} (Compartment $2$.c). The pattern `p1' (orange circle) in the first level consisted of one wall vertex connected to five longitudinal vertices, whereas `p2' in level $2$ (yellow circle) consisted of one compartment vertex connected to two wall vertices, two `p1' vertices, and four manholes. As `p2' contained $\vartheta^e$, the graph prediction step was triggered and the predicted graph is shown in Figure~\ref{fig:rs_assist} (Compartment $2$.c). 

The robot navigated to the next compartment through the manhole used for graph prediction and switched to the \ac{ae} submode. Two \ac{ae} steps were conducted (shown in Figure~\ref{fig:rs_assist} (Compartment $3$.a)) until sufficient overlap between the detected and predicted longitudinals was achieved, at which point the planner switched to the \ac{si} mode. Upon completion, the pattern detection step was triggered. Similar substructures as at the end of Compartment $2$ were detected and the graph prediction was carried out(Figure~\ref{fig:rs_assist} (Compartment $3$.c)). The robot navigated to the next compartment, passing through the relevant manhole, and the process was continued. Figure~\ref{fig:rs_assist} shows indicative planning steps in each of the compartments. The total mission time was $\SI{276}{\second}$.

\subsubsection{\textbf{Mission 2: Opportunistic Inspection}}
During Mission 2, the behavior of the predictive planner in the first two compartments was similar to that in Mission 1. Upon completing the inspection in the second compartment, the pattern detection step was triggered. The hierarchy built was similar to that in Mission 1 as shown in Figure~\ref{fig:rs_opp} (Compartment $2$.c). The graph prediction step was carried out and the predicted graph and predicted inspection path are shown in Figure~\ref{fig:rs_opp} (Compartment $2$.c). The robot navigated through the relevant manhole to enter the next compartment and the \ac{oi} submode was activated. As no longitudinals were detected at the beginning of the first iteration, the planner took an exploration step (Figure~\ref{fig:rs_opp} (Compartment $3$.a). At this point, the longitudinal corresponding to the next viewpoint in the predicted path was detected and the planner planed a path to that viewpoint. This process was repeated until all viewpoints in the predicted path were visited. Next, the pattern detection and graph prediction steps were carried out. After traversing through the relevant manhole, the \ac{oi} submode was triggered as successful graph prediction was made. Figure~\ref{fig:rs_opp} shows indicative planning steps for each compartment. The total mission time in this case was $\SI{267}{\second}$.

\subsubsection{\textbf{Mission 3: Baseline}}
The Baseline approach showed similar behavior as that in Field Deployment 1. In each compartment, the planner performed the \ac{ve} and \ac{si} steps without triggering the pattern detection and graph prediction steps. The final map, scene graph, and indicative planning steps in two compartments are shown in Figure~\ref{fig:rs_van}.  The total mission time was $\SI{314}{\second}$.


The quantitative results comparing the performance of the three missions are shown in Table~\ref{tab:exp_stats} (Field Deployment 2). 
Similar to Field Deployment 1, the predictive planning submodes outperformed the Baseline, with the \ac{oi} having the best performance in terms of inspection time per compartment. 


\begin{table*}[]
\centering
\begin{tabular}{|l|l|l|l|l|}
\hline
\textbf{}          & \textbf{Inspection Time /}              & \textbf{Semantic Surface}   & \textbf{Path Length} & \textbf{Computation Time} \\ 
\textbf{Method}          & \textbf{Compartment (s)}              & \textbf{Coverage (\%)}   & \textbf{(m)} & \textbf{(ms)} \\ \hline
\multicolumn{5}{|l|}{\textbf{Field Deployment 1}} \\ \hline
PP-AE                       & $61.0$                & $97.4603$      & $119.78$  & $270.54$ \\ \hline
PP-OI                       & $57.0$                & $97.3727$      & $115.77$  & $63.91$ \\ \hline
Baseline                    & $73.5$                & $96.3218$      & $129.17$  & $426.62$ \\ \hline
\multicolumn{5}{|l|}{\textbf{Field Deployment 2}} \\ \hline
PP-AE                       & $58.5$                & $95.9184$      & $128.55$  & $245.02$ \\ \hline
PP-OI                       & $55.5$                & $95.3719$      & $125.96$  & $71.29$ \\ \hline
Baseline                    & $72.0$                & $93.9935$      & $131.55$  & $414.75$ \\ \hline
\end{tabular}
\vspace{2ex}
\caption{Quantitative results from both Field Deployments. We deploy the two predictive planning submodes and the Baseline in two ships. The results show that the proposed submodes outperform the Baseline in terms of inspection time per compartment while achieving equal or better semantic coverage in both deployments. The average computation times for each method are shown in the last column. For the \ac{ae}, \ac{oi} the computation time is reported only for the planning steps when the planner was operating in the respective submodes.}
\label{tab:exp_stats}
\end{table*}


\subsection{Summary}
We presented field deployments in the ballast tanks of two ships demonstrating the field readiness of our method. The planner operating in both predictive planning submodes was tested along with the Baseline solution that does not exploit pattern detection and prediction. The \ac{ae} and \ac{oi} submodes show an improvement in the inspection time per compartment of $17.1\%$, $22.6\%$ respectively in Field Deployment 1 and $19.4\%$, $23.6\%$ respectively  over the Baseline. The deployments show that the semantics-aware predictive planning paradigm proposed in this paper can make the inspection planning more efficient while obtaining the same semantic coverage. Among the two predictive planning approaches presented in the paper, the \ac{oi} submode provides marginally better results than \ac{ae} as shown by the quantitative results aligning with the trend seen in the simulation study. 

%% file: 08_Conclusion.tex
This paper presents a new paradigm called ``Semantics-aware Predictive Planning'' that exploits structured spatial semantics patterns in an environment. Specifically, we represent the semantics in the environment as a Semantic Scene Graph. The method exploits and extends a graph structure discovery algorithm called SUBDUE. Key modifications are made to the original algorithm for handling the case of inexact patterns more accurately and for utilization of the pose information of the semantics in the \ac{ssg}. Using these patterns, the work proposes a strategy to identify areas in the \ac{ssg} that can extend to the detected patterns and predicts the evolution of the \ac{ssg}. Finally, we present two inspection path planning strategies that exploit these graph predictions, and tailor them to the application of ship ballast tank inspection. 
Through simulation studies in a model of a ballast tank \addition{and and industrial factory-like environment}, the paper demonstrates that the proposed predictive planning paradigm can perform semantic inspection faster (ranging from $25\%$ to $60\%$ less inspection time) than state-of-the-art exploration and inspection planners as well as a semantics-aware ``Baseline'' solution that used the same volumetric exploration and semantic inspection planners without exploiting the graph predictions, while maintaining equal or higher semantic coverage. Furthermore, we show the benefit of the proposed modifications to SUBDUE through a simulation study in a ballast tank having imperfect patterns. 
The method is deployed in the ballast tanks of two oil tanker ships onboard a collision-tolerant aerial robot. Both the proposed predictive planning strategies along with the Baseline are tested resulting in a total of $6$ experiments. The proposed approaches are able to achieve up to $23\%$ improvement over the Baseline showing the real-world applicability of the method. 

Regarding future work, we have identified four possible avenues. The first relates to a probabilistic formulation of the pattern detection strategy. Currently, the output of the pattern detection algorithm is independent of the previous times it was triggered. Furthermore, it does not account for the detection probabilities of the semantics. Future work can focus on an algorithm that can account for the above and provide a confidence metric over the detected patterns and track them over time. The second future research direction can be to include prior knowledge about patterns in both the pattern detection as well as graph prediction steps. Third, investigation can happen on graph prediction strategies that do not restrict the prediction to the use of specific entry classes. This will enable this paradigm to be used in a larger variety of environments. Finally, research can focus on path planning for actively searching for patterns based on prior knowledge or graph predictions.

%% file: 00_PredictivePlanning.bbl
\begin{thebibliography}{10}
\providecommand{\url}[1]{#1}
\csname url@samestyle\endcsname
\providecommand{\newblock}{\relax}
\providecommand{\bibinfo}[2]{#2}
\providecommand{\BIBentrySTDinterwordspacing}{\spaceskip=0pt\relax}
\providecommand{\BIBentryALTinterwordstretchfactor}{4}
\providecommand{\BIBentryALTinterwordspacing}{\spaceskip=\fontdimen2\font plus
\BIBentryALTinterwordstretchfactor\fontdimen3\font minus \fontdimen4\font\relax}
\providecommand{\BIBforeignlanguage}[2]{{%
\expandafter\ifx\csname l@#1\endcsname\relax
\typeout{** WARNING: IEEEtran.bst: No hyphenation pattern has been}%
\typeout{** loaded for the language `#1'. Using the pattern for}%
\typeout{** the default language instead.}%
\else
\language=\csname l@#1\endcsname
\fi
#2}}
\providecommand{\BIBdecl}{\relax}
\BIBdecl

\bibitem{GBPLANNER_JFR_2020}
T.~Dang, M.~Tranzatto, S.~Khattak, F.~Mascarich, K.~Alexis, and M.~Hutter, ``Graph-based subterranean exploration path planning using aerial and legged robots,'' \emph{Journal of Field Robotics}, vol.~37, no.~8, pp. 1363--1388, 2020.

\bibitem{agha2021nebula}
A.~Agha, K.~Otsu, B.~Morrell, D.~D. Fan, R.~Thakker, A.~Santamaria-Navarro, S.-K. Kim, A.~Bouman, X.~Lei, J.~Edlund \emph{et~al.}, ``Nebula: Quest for robotic autonomy in challenging environments; team costar at the darpa subterranean challenge,'' \emph{arXiv preprint arXiv:2103.11470}, 2021.

\bibitem{BABOOMS_ICRA_15}
\BIBentryALTinterwordspacing
{A. Bircher, K. Alexis, M. Burri, P. Oettershagen, S. Omari, T. Mantel and R. Siegwart}, ``Structural inspection path planning via iterative viewpoint resampling with application to aerial robotics,'' in \emph{IEEE International Conference on Robotics and Automation (ICRA)}, May 2015, pp. 6423--6430. [Online]. Available: \url{https://github.com/ethz-asl/StructuralInspectionPlanner}
\BIBentrySTDinterwordspacing

\bibitem{cao2021tare}
C.~Cao, H.~Zhu, H.~Choset, and J.~Zhang, ``Tare: A hierarchical framework for efficiently exploring complex 3d environments.'' in \emph{Robotics: Science and Systems}, 2021.

\bibitem{shukla2016application}
A.~Shukla and H.~Karki, ``Application of robotics in onshore oil and gas industry—a review part i,'' \emph{Robotics and Autonomous Systems}, vol.~75, pp. 490--507, 2016.

\bibitem{sa2014vertical}
I.~Sa and P.~Corke, ``Vertical infrastructure inspection using a quadcopter and shared autonomy control,'' in \emph{Field and service robotics}.\hskip 1em plus 0.5em minus 0.4em\relax Springer, 2014, pp. 219--232.

\bibitem{gehring2019anymal}
C.~Gehring, P.~Fankhauser, L.~Isler, R.~Diethelm, S.~Bachmann, M.~Potz, L.~Gerstenberg, and M.~Hutter, ``Anymal in the field: Solving industrial inspection of an offshore hvdc platform with a quadrupedal robot,'' in \emph{12th Conference on Field and Service Robotics (FSR 2019)}, 2019.

\bibitem{caprari2012highly}
G.~Caprari, A.~Breitenmoser, W.~Fischer, C.~H{\"u}rzeler, F.~T{\^a}che, R.~Siegwart, O.~Nguyen, R.~Moser, P.~Schoeneich, and F.~Mondada, ``Highly compact robots for inspection of power plants,'' \emph{Journal of Field Robotics}, vol.~29, no.~1, pp. 47--68, 2012.

\bibitem{chan2015towards}
B.~Chan, H.~Guan, J.~Jo, and M.~Blumenstein, ``Towards uav-based bridge inspection systems: A review and an application perspective,'' \emph{Structural Monitoring and Maintenance}, vol.~2, no.~3, pp. 283--300, 2015.

\bibitem{bartolomei2023fast}
L.~Bartolomei, L.~Teixeira, and M.~Chli, ``Fast multi-uav decentralized exploration of forests,'' \emph{IEEE Robotics and Automation Letters}, 2023.

\bibitem{CERBERUS_SCIENCE_2022}
M.~Tranzatto, T.~Miki, M.~Dharmadhikari, L.~Bernreiter, M.~Kulkarni, F.~Mascarich, O.~Andersson, S.~Khattak, M.~Hutter, R.~Siegwart, and K.~Alexis, ``Cerberus in the darpa subterranean challenge,'' \emph{Science Robotics}, vol.~7, no.~66, p. eabp9742, 2022.

\bibitem{CERBERUS_WINS_FR2022submission}
M.~Tranzatto, M.~Dharmadhikari, L.~Bernreiter, M.~Camurri, S.~Khattak, F.~Mascarich, P.~Pfreundschuh, D.~Wisth, S.~Zimmermann, M.~Kulkarni \emph{et~al.}, ``Team cerberus wins the darpa subterranean challenge: Technical overview and lessons learned,'' \emph{arXiv preprint arXiv:2207.04914}, 2022.

\bibitem{GBPLANNER2COHORT_ICRA_2022}
M.~Kulkarni, M.~Dharmadhikari, M.~Tranzatto, S.~Zimmermann, V.~Reijgwart, P.~De~Petris, H.~Nguyen, N.~Khedekar, C.~Papachristos, L.~Ott, R.~Siegwart, M.~Hutter, and K.~Alexis, ``Autonomous teamed exploration of subterranean environments using legged and aerial robots,'' in \emph{2022 International Conference on Robotics and Automation (ICRA)}.\hskip 1em plus 0.5em minus 0.4em\relax IEEE, 2022, pp. 3306--3313.

\bibitem{hudson2021heterogeneous}
N.~Hudson, F.~Talbot, M.~Cox, J.~Williams, T.~Hines, A.~Pitt, B.~Wood, D.~Frousheger, K.~L. Surdo, T.~Molnar \emph{et~al.}, ``Heterogeneous ground and air platforms, homogeneous sensing: Team csiro data61's approach to the darpa subterranean challenge,'' \emph{arXiv preprint arXiv:2104.09053}, 2021.

\bibitem{rouvcek2019darpa}
T.~Rou{\v{c}}ek, M.~Pecka, P.~{\v{C}}{\'\i}{\v{z}}ek, T.~Pet{\v{r}}{\'\i}{\v{c}}ek, J.~Bayer, V.~{\v{S}}alansk{\`y}, D.~He{\v{r}}t, M.~Petrl{\'\i}k, T.~B{\'a}{\v{c}}a, V.~Spurn{\`y} \emph{et~al.}, ``Darpa subterranean challenge: Multi-robotic exploration of underground environments,'' in \emph{International Conference on Modelling and Simulation for Autonomous Systems}.\hskip 1em plus 0.5em minus 0.4em\relax Springer, Cham, 2019, pp. 274--290.

\bibitem{explorer_phase_i_ii}
S.~Scherer, V.~Agrawal, G.~Best, C.~Cao, K.~Cujic, R.~Darnley, R.~DeBortoli, E.~Dexheimer, B.~Drozd, R.~Garg, I.~Higgins, J.~Keller, D.~Kohanbash, L.~Nogueira, R.~Pradhan, M.~Tatum, V.~K.~Viswanathan, S.~Willits, S.~Zhao, H.~Zhu, D.~Abad, T.~Angert, G.~Armstrong, R.~Boirum, A.~Dongare, M.~Dworman, S.~Hu, J.~Jaekel, R.~Ji, A.~Lai, Y.~Hsuan~Lee, A.~Luong, J.~Mangelson, J.~Maier, J.~Picard, K.~Pluckter, A.~Saba, M.~Saroya, E.~Scheide, N.~Shoemaker-Trejo, J.~Spisak, J.~Teza, F.~Yang, A.~Wilson, H.~Zhang, H.~Choset, M.~Kaess, A.~Rowe, S.~Singh, J.~Zhang, G.~A.~Hollinger, and M.~Travers, ``Resilient and modular subterranean exploration with a team of roving and flying robots,'' \emph{Field Robotics}, 2021.

\bibitem{hughes2022hydra}
N.~Hughes, Y.~Chang, and L.~Carlone, ``Hydra: A real-time spatial perception system for {3D} scene graph construction and optimization,'' 2022.

\bibitem{hughes2024foundations}
\BIBentryALTinterwordspacing
N.~Hughes, Y.~Chang, S.~Hu, R.~Talak, R.~Abdulhai, J.~Strader, and L.~Carlone, ``Foundations of spatial perception for robotics: Hierarchical representations and real-time systems,'' \emph{The International Journal of Robotics Research}, 2024. [Online]. Available: \url{https://doi.org/10.1177/02783649241229725}
\BIBentrySTDinterwordspacing

\bibitem{Wu_2021_CVPR}
S.-C. Wu, J.~Wald, K.~Tateno, N.~Navab, and F.~Tombari, ``Scenegraphfusion: Incremental 3d scene graph prediction from rgb-d sequences,'' in \emph{Proceedings of the IEEE/CVF Conference on Computer Vision and Pattern Recognition (CVPR)}, June 2021, pp. 7515--7525.

\bibitem{schmid2022panoptic}
L.~Schmid, J.~Delmerico, J.~Sch{\"o}nberger, J.~Nieto, M.~Pollefeys, R.~Siegwart, and C.~Cadena, ``Panoptic multi-tsdfs: a flexible representation for online multi-resolution volumetric mapping and long-term dynamic scene consistency,'' in \emph{2022 IEEE International Conference on Robotics and Automation (ICRA)}, 2022, pp. 8018--8024.

\bibitem{Wang2021semantic_info_planning}
C.~Wang, J.~Cheng, W.~Chi, T.~Yan, and M.~Q.-H. Meng, ``Semantic-aware informative path planning for efficient object search using mobile robot,'' \emph{IEEE Transactions on Systems, Man, and Cybernetics: Systems}, vol.~51, no.~8, pp. 5230--5243, 2021.

\bibitem{Papatheodorou_ICRA2023}
S.~Papatheodorou, N.~Funk, D.~Tzoumanikas, C.~Choi, B.~Xu, and S.~Leutenegger, ``Finding things in the unknown: Semantic object-centric exploration with an {MAV},'' in \emph{IEEE International Conference on Robotics and Automation}, London, United Kingdom, May 2023.

\bibitem{fredriksson2024topometric}
S.~Fredriksson, A.~Saradagi, and G.~Nikolakopoulos, ``Robotic exploration through semantic topometric mapping,'' in \emph{2024 IEEE International Conference on Robotics and Automation (ICRA)}, 2024, pp. 9404--9410.

\bibitem{ginting2024semanticbeliefbehaviorgraph}
\BIBentryALTinterwordspacing
M.~F. Ginting, D.~D. Fan, S.-K. Kim, M.~J. Kochenderfer, and A.~akbar Agha-mohammadi, ``Semantic belief behavior graph: Enabling autonomous robot inspection in unknown environments,'' 2024. [Online]. Available: \url{https://arxiv.org/abs/2401.17191}
\BIBentrySTDinterwordspacing

\bibitem{roth2023viplanner}
P.~Roth, J.~Nubert, F.~Yang, M.~Mittal, and M.~Hutter, ``Viplanner: Visual semantic imperative learning for local navigation,'' \emph{2024 IEEE International Conference on Robotics and Automation (ICRA)}, May 2023.

\bibitem{rosinol2020kimera}
A.~Rosinol, M.~Abate, Y.~Chang, and L.~Carlone, ``Kimera: an open-source library for real-time metric-semantic localization and mapping,'' in \emph{2020 IEEE International Conference on Robotics and Automation (ICRA)}.\hskip 1em plus 0.5em minus 0.4em\relax IEEE, 2020, pp. 1689--1696.

\bibitem{maggio2024clio}
D.~Maggio, Y.~Chang, N.~Hughes, M.~Trang, D.~Griffith, C.~Dougherty, E.~Cristofalo, L.~Schmid, and L.~Carlone, ``Clio: Real-time task-driven open-set 3d scene graphs,'' \emph{arXiv preprint arXiv:2404.13696}, 2024.

\bibitem{bavle2022situational}
H.~Bavle, J.~L. Sanchez-Lopez, M.~Shaheer, J.~Civera, and H.~Voos, ``Situational graphs for robot navigation in structured indoor environments,'' \emph{IEEE Robotics and Automation Letters}, vol.~7, no.~4, pp. 9107--9114, 2022.

\bibitem{2023swap}
M.~Dharmadhikari and K.~Alexis, ``Semantics-aware exploration and inspection path planning,'' in \emph{2023 IEEE International Conference on Robotics and Automation (ICRA)}, 2023, pp. 3360--3367.

\bibitem{ginting2024seek}
M.~F. Ginting, S.-K. Kim, D.~D. Fan, M.~Palieri, M.~J. Kochenderfer, and A.~akbar Agha-mohammadi, ``Seek: Semantic reasoning for object goal navigation in real world inspection tasks,'' in \emph{Robotics: Science and Systems}, 2024.

\bibitem{voxblox}
H.~Oleynikova, Z.~Taylor, M.~Fehr, R.~Siegwart, and J.~Nieto, ``Voxblox: Incremental 3d euclidean signed distance fields for on-board mav planning,'' in \emph{IEEE/RSJ International Conference on Intelligent Robots and Systems (IROS)}, 2017.

\bibitem{armeni_iccv19}
I.~Armeni, Z.-Y. He, J.~Gwak, A.~R. Zamir, M.~Fischer, J.~Malik, and S.~Savarese, ``3d scene graph: A structure for unified semantics, 3d space, and camera,'' in \emph{Proceedings of the IEEE International Conference on Computer Vision}, 2019.

\bibitem{kim2020ssg}
U.-H. Kim, J.-M. Park, T.-j. Song, and J.-H. Kim, ``3-d scene graph: A sparse and semantic representation of physical environments for intelligent agents,'' \emph{IEEE Transactions on Cybernetics}, vol.~50, no.~12, pp. 4921--4933, 2020.

\bibitem{rosinol2020dssg}
A.~Rosinol, A.~Gupta, M.~Abate, J.~Shi, and L.~Carlone, ``3d dynamic scene graphs: Actionable spatial perception with places, objects, and humans,'' \emph{Robotics: Science and Systems}, 2020.

\bibitem{rmfowl}
P.~D. Petris, H.~Nguyen, M.~Dharmadhikari, M.~Kulkarni, N.~Khedekar, F.~Mascarich, and K.~Alexis, ``Rmf-owl: A collision-tolerant flying robot for autonomous subterranean exploration,'' in \emph{2022 International Conference on Unmanned Aircraft Systems (ICUAS)}, 2022, pp. 536--543.

\bibitem{2022domain_adapt_seg}
R.~Zurbrügg, H.~Blum, C.~Cadena, R.~Siegwart, and L.~Schmid, ``Embodied active domain adaptation for semantic segmentation via informative path planning,'' \emph{IEEE Robotics and Automation Letters}, vol.~7, no.~4, pp. 8691--8698, 2022.

\bibitem{yin2023dformer}
B.~Yin, X.~Zhang, Z.~Li, L.~Liu, M.-M. Cheng, and Q.~Hou, ``Dformer: Rethinking rgbd representation learning for semantic segmentation,'' \emph{arXiv preprint arXiv:2309.09668}, 2023.

\bibitem{jia2024geminifusion}
D.~Jia, J.~Guo, K.~Han, H.~Wu, C.~Zhang, C.~Xu, and X.~Chen, ``Geminifusion: Efficient pixel-wise multimodal fusion for vision transformer,'' 2024.

\bibitem{zhang2023cmx}
J.~Zhang, H.~Liu, K.~Yang, X.~Hu, R.~Liu, and R.~Stiefelhagen, ``Cmx: Cross-modal fusion for rgb-x semantic segmentation with transformers,'' \emph{IEEE Transactions on Intelligent Transportation Systems}, 2023.

\bibitem{alama2025rayfrontsopensetsemanticray}
\BIBentryALTinterwordspacing
O.~Alama, A.~Bhattacharya, H.~He, S.~Kim, Y.~Qiu, W.~Wang, C.~Ho, N.~Keetha, and S.~Scherer, ``Rayfronts: Open-set semantic ray frontiers for online scene understanding and exploration,'' 2025. [Online]. Available: \url{https://arxiv.org/abs/2504.06994}
\BIBentrySTDinterwordspacing

\bibitem{viswanathan2024actionablehierarchicalscenerepresentation}
\BIBentryALTinterwordspacing
V.~K. Viswanathan, M.~A.~V. Saucedo, S.~G. Satpute, C.~Kanellakis, and G.~Nikolakopoulos, ``An actionable hierarchical scene representation enhancing autonomous inspection missions in unknown environments,'' 2024. [Online]. Available: \url{https://arxiv.org/abs/2412.19582}
\BIBentrySTDinterwordspacing

\bibitem{zaenker2020hypermap}
T.~Zaenker, F.~Verdoja, and V.~Kyrki, ``Hypermap mapping framework and its application to autonomous semantic exploration,'' in \emph{2020 IEEE International Conference on Multisensor Fusion and Integration for Intelligent Systems (MFI)}, 2020, pp. 133--139.

\bibitem{kay2021semanticnbvreconstruction}
S.~A. Kay, S.~Julier, and V.~M. Pawar, ``Semantically informed next best view planning for autonomous aerial 3d reconstruction,'' in \emph{2021 IEEE/RSJ International Conference on Intelligent Robots and Systems (IROS)}, 2021, pp. 3125--3130.

\bibitem{simons2025seguesemanticguidedexploration}
\BIBentryALTinterwordspacing
C.~Simons, A.~Samanta, A.~K. Roy-Chowdhury, and K.~Karydis, ``Segue: Semantic guided exploration for mobile robots,'' 2025. [Online]. Available: \url{https://arxiv.org/abs/2504.03629}
\BIBentrySTDinterwordspacing

\bibitem{rs12050891}
\BIBentryALTinterwordspacing
R.~Ashour, T.~Taha, J.~M.~M. Dias, L.~Seneviratne, and N.~Almoosa, ``Exploration for object mapping guided by environmental semantics using uavs,'' \emph{Remote Sensing}, vol.~12, no.~5, 2020. [Online]. Available: \url{https://www.mdpi.com/2072-4292/12/5/891}
\BIBentrySTDinterwordspacing

\bibitem{yu2024semanticawarenextbestviewmultidofsmobile}
\BIBentryALTinterwordspacing
X.~Yu and C.-W. Chen, ``Semantic-aware next-best-view for multi-dofs mobile system in search-and-acquisition based visual perception,'' 2024. [Online]. Available: \url{https://arxiv.org/abs/2404.16507}
\BIBentrySTDinterwordspacing

\bibitem{milas2023asep}
A.~Milas, A.~Ivanovic, and T.~Petrovic, ``Asep: An autonomous semantic exploration planner with object labeling,'' \emph{IEEE Access}, vol.~11, pp. 107\,169--107\,183, 2023.

\bibitem{lu2024semRecedingHorizon}
L.~Lu, Y.~Zhang, P.~Zhou, J.~Qi, Y.~Pan, C.~Fu, and J.~Pan, ``Semantics-aware receding horizon planner for object-centric active mapping,'' \emph{IEEE Robotics and Automation Letters}, vol.~9, no.~4, pp. 3838--3845, 2024.

\bibitem{STACHE2023104288}
\BIBentryALTinterwordspacing
F.~Stache, J.~Westheider, F.~Magistri, C.~Stachniss, and M.~Popović, ``Adaptive path planning for uavs for multi-resolution semantic segmentation,'' \emph{Robotics and Autonomous Systems}, vol. 159, p. 104288, 2023. [Online]. Available: \url{https://www.sciencedirect.com/science/article/pii/S0921889022001774}
\BIBentrySTDinterwordspacing

\bibitem{rukin2023activelearning}
J.~Rückin, F.~Magistri, C.~Stachniss, and M.~Popović, ``An informative path planning framework for active learning in uav-based semantic mapping,'' \emph{IEEE Transactions on Robotics}, vol.~39, no.~6, pp. 4279--4296, 2023.

\bibitem{liu2023multiaerialsemanticmapping}
X.~Liu, A.~Prabhu, F.~Cladera, I.~D. Miller, L.~Zhou, C.~J. Taylor, and V.~Kumar, ``Active metric-semantic mapping by multiple aerial robots,'' in \emph{2023 IEEE International Conference on Robotics and Automation (ICRA)}, 2023, pp. 3282--3288.

\bibitem{miller2024spomp}
I.~D. Miller, F.~Cladera, T.~Smith, C.~J. Taylor, and V.~Kumar, ``Air-ground collaboration with spomp: Semantic panoramic online mapping and planning,'' \emph{IEEE Transactions on Field Robotics}, pp. 1--1, 2024.

\bibitem{cladera2024challengesopportunitieslargescaleexploration}
\BIBentryALTinterwordspacing
F.~Cladera, I.~D. Miller, Z.~Ravichandran, V.~Murali, J.~Hughes, M.~A. Hsieh, C.~J. Taylor, and V.~Kumar, ``Challenges and opportunities for large-scale exploration with air-ground teams using semantics,'' 2024. [Online]. Available: \url{https://arxiv.org/abs/2405.07169}
\BIBentrySTDinterwordspacing

\bibitem{bartolomei2021activesemanticperception}
L.~Bartolomei, L.~Teixeira, and M.~Chli, ``Semantic-aware active perception for uavs using deep reinforcement learning,'' in \emph{2021 IEEE/RSJ International Conference on Intelligent Robots and Systems (IROS)}, 2021, pp. 3101--3108.

\bibitem{HU2025105949_indoorSemanticNav}
\BIBentryALTinterwordspacing
D.~Hu and V.~J. Gan, ``Semantic navigation for automated robotic inspection and indoor environment quality monitoring,'' \emph{Automation in Construction}, vol. 170, p. 105949, 2025. [Online]. Available: \url{https://www.sciencedirect.com/science/article/pii/S092658052400685X}
\BIBentrySTDinterwordspacing

\bibitem{chen2023rspmp}
D.~Chen, M.~Zhuang, X.~Zhong, W.~Wu, and Q.~Liu, ``Rspmp: real-time semantic perception and motion planning for autonomous navigation of unmanned ground vehicle in off-road environments.'' \emph{Applied Intelligence}, 2023.

\bibitem{sun2024rsmpnet}
J.~Sun, J.~Wu, Z.~Ji, and Y.-K. Lai, ``Rsmpnet: Relationship guided semantic map prediction,'' in \emph{2024 IEEE/CVF Winter Conference on Applications of Computer Vision (WACV)}, 2024, pp. 302--311.

\bibitem{du2020learningrelationgraph}
H.~Du, X.~Yu, and L.~Zheng, ``Learning object relation graph and tentative policy for visual navigation,'' in \emph{Computer Vision -- ECCV 2020}.\hskip 1em plus 0.5em minus 0.4em\relax Cham: Springer International Publishing, 2020, pp. 19--34.

\bibitem{liang2021sscnav}
Y.~Liang, B.~Chen, and S.~Song, ``Sscnav: Confidence-aware semantic scene completion for visual semantic navigation,'' in \emph{2021 IEEE International Conference in Robotics and Automation (ICRA)}, 2021.

\bibitem{Yang2019_113270}
W.~Yang, X.~Wang, A.~Farhadi, A.~Gupta, and R.~Mottaghi, ``Visual semantic navigation using scene priors,'' in \emph{Proceedings of (ICLR) International Conference on Learning Representations}, May 2019.

\bibitem{qiu2020learning}
Y.~{Qiu}, A.~{Pal}, and H.~I. {Christensen}, ``Learning hierarchical relationships for object-goal navigation,'' in \emph{2020 Conference on Robot Learning (CoRL)}, 2020.

\bibitem{ravichandran2025spineonlinesemanticplanning}
\BIBentryALTinterwordspacing
Z.~Ravichandran, V.~Murali, M.~Tzes, G.~J. Pappas, and V.~Kumar, ``Spine: Online semantic planning for missions with incomplete natural language specifications in unstructured environments,'' 2025. [Online]. Available: \url{https://arxiv.org/abs/2410.03035}
\BIBentrySTDinterwordspacing

\bibitem{shrestha2019learnedmapprediction}
R.~Shrestha, F.-P. Tian, W.~Feng, P.~Tan, and R.~Vaughan, ``Learned map prediction for enhanced mobile robot exploration,'' in \emph{2019 International Conference on Robotics and Automation (ICRA)}, 2019, pp. 1197--1204.

\bibitem{tao2023seer}
Y.~Tao, Y.~Wu, B.~Li, F.~Cladera, A.~Zhou, D.~Thakur, and V.~Kumar, ``Seer: Safe efficient exploration for aerial robots using learning to predict information gain,'' in \emph{2023 IEEE International Conference on Robotics and Automation (ICRA)}, 2023, pp. 1235--1241.

\bibitem{strom2015predictfromprev}
D.~P. Ström, F.~Nenci, and C.~Stachniss, ``Predictive exploration considering previously mapped environments,'' in \emph{2015 IEEE International Conference on Robotics and Automation (ICRA)}, 2015, pp. 2761--2766.

\bibitem{scenecompletionsurvey}
L.~Roldão, R.~de~Charette, and A.~Verroust-Blondet, ``3d semantic scene completion: A survey,'' \emph{International Journal of Computer Vision}, 2022.

\bibitem{Holder2002}
\BIBentryALTinterwordspacing
L.~Holder, D.~Cook, J.~Gonzalez, and I.~Jonyer, \emph{Structural Pattern Recognition in Graphs}.\hskip 1em plus 0.5em minus 0.4em\relax Boston, MA: Springer US, 2002, pp. 255--279. [Online]. Available: \url{https://doi.org/10.1007/978-1-4613-0231-5_10}
\BIBentrySTDinterwordspacing

\bibitem{manhole2023}
M.~Dharmadhikari, P.~De~Petris, H.~Nguyen, M.~Kulkarni, N.~Khedekar, and K.~Alexis, ``Manhole detection and traversal for exploration of ballast water tanks using micro aerial vehicles,'' in \emph{2023 International Conference on Unmanned Aircraft Systems (ICUAS)}, 2023, pp. 103--109.

\bibitem{helsgaun2000effective}
K.~Helsgaun, ``An effective implementation of the lin--kernighan traveling salesman heuristic,'' \emph{European journal of operational research}, vol. 126, no.~1, pp. 106--130, 2000.

\bibitem{karaman2010rrtstar}
S.~Karaman and E.~Frazzoli, ``Incremental sampling-based algorithms for optimal motion planning,'' 2010.

\bibitem{2023expgvi}
M.~Dharmadhikari, P.~De~Petris, M.~Kulkarni, N.~Khedekar, H.~Nguyen, A.~E. Stene, E.~Sjøvold, K.~Solheim, B.~Gussiaas, and K.~Alexis, ``Autonomous exploration and general visual inspection of ship ballast water tanks using aerial robots,'' in \emph{2023 21st International Conference on Advanced Robotics (ICAR)}, 2023, pp. 409--416.

\bibitem{zhou2021fuel}
B.~Zhou, Y.~Zhang, X.~Chen, and S.~Shen, ``Fuel: Fast uav exploration using incremental frontier structure and hierarchical planning,'' \emph{IEEE Robotics and Automation Letters}, vol.~6, no.~2, pp. 779--786, 2021.

\bibitem{khattak2020complementary}
S.~Khattak, H.~Nguyen, F.~Mascarich, T.~Dang, and K.~Alexis, ``Complementary multi--modal sensor fusion for resilient robot pose estimation in subterranean environments,'' in \emph{2020 International Conference on Unmanned Aircraft Systems (ICUAS)}.\hskip 1em plus 0.5em minus 0.4em\relax IEEE, 2020, pp. 1024--1029.

\bibitem{mpc_rosbookchapter}
{M. Kamel, T. Stastny, K. Alexis, and R. Siegwart}, ``Model predictive control for trajectory tracking of unmanned aerial vehicles using ros,'' \emph{Springer Book on Robot Operating System (ROS)}.

\end{thebibliography}
